\documentclass[preprint,3p,times]{elsarticle}

\usepackage{amssymb}
\usepackage{amsthm}

\usepackage{amsmath}
\usepackage{url}
\usepackage{syntax}
\usepackage{multirow}
\usepackage{listings}
\usepackage{algorithm}
\usepackage[noend]{algpseudocode}
\usepackage{subcaption}
\usepackage{booktabs}
\usepackage{bbm}
\usepackage{xcolor}

\theoremstyle{definition}
\newtheorem{definition}{Definition}
\newtheorem{example}{Example}
\newtheorem{assumption}{Assumption}
\theoremstyle{plain}
\newtheorem{lemma}{Lemma}
\newtheorem{sublemma}{Sublemma}
\newtheorem{theorem}{Theorem}
\newtheorem{corollary}{Corollary}

\DeclareMathAlphabet{\mathbbm}{U}{bbold}{m}{n}
\newcommand{\Description}[1]{}

\journal{Artificial Intelligence}

\begin{document}

\begin{frontmatter}



    \title{Domain-Independent Dynamic Programming}


    \author[nii,sokendai]{Ryo Kuroiwa}
    \ead{kuroiwa@nii.ac.jp}
    \affiliation[nii]{organization={National Institute of Informatics},
        addressline={2-1-2 Hitotsubashi, Chiyoda-ku},
        city={Tokyo},
        postcode={101-8430},
        country={Japan}}
    \affiliation[sokendai]{organization={The Graduate University for Advanced Studies, SOKENDAI},
        addressline={Shonan Village, Hayama},
        city={Kanagawa},
        postcode={240-0193},
        country={Japan}}

    \author[uoft]{J. Christopher Beck}
    \ead{jcb@mie.utoronto.ca}
    \affiliation[uoft]{organization={Department of Mechanical and Industrial Enginnering, University of Toronto},
        addressline={5 King's College Road},
        city={Toronto},
        postcode={M5S 3G8},
        state={Ontario},
        country={Canada}}

    \begin{abstract}
        For combinatorial optimization problems, model-based paradigms such as mixed-integer programming (MIP) and constraint programming (CP) aim to decouple modeling and solving a problem: the `holy grail' of declarative problem solving.
We propose domain-independent dynamic programming (DIDP), a novel model-based paradigm based on dynamic programming (DP).
While DP is not new, it has typically been implemented as a problem-specific method.
We introduce Dynamic Programming Description Language (DyPDL), a formalism to define DP models based on a state transition system, inspired by artificial intelligence (AI) planning.
We show that heuristic search algorithms can be used to solve DyPDL models and propose seven DIDP solvers.
We experimentally compare our DIDP solvers with commercial MIP and CP solvers (solving MIP and CP models, respectively) on common benchmark instances of eleven combinatorial optimization problem classes.
We show that DIDP outperforms MIP in nine problem classes, CP also in nine problem classes, and both MIP and CP in seven.
DIDP also achieves superior performance to existing state-based solvers including domain-independent AI planners.

    \end{abstract}


    \begin{highlights}
        \item We propose domain-independent dynamic programming (DIDP), a novel model-based paradigm for combinatorial optimization.
        \item The modeling language for DIDP is designed so that a user can investigate efficient models by incorporating redundant information.
        \item We implement DIDP solvers using heuristic search in an open-source software framework.
        \item The DIDP solvers outperform commercial mixed-integer programming and constraint programming solvers in a number of problem classes.
    \end{highlights}


\end{frontmatter}






\section{Introduction} \label{sec:introduction}

Combinatorial optimization is a class of problems requiring a set of discrete decisions to be made to optimize an objective function.
It has wide real-world application fields including transportation \cite{VRP2014}, scheduling \cite{Cheng1993,Pinedo2009}, and manufacturing \cite{Boysen2021} and thus has been an active research topic in artificial intelligence (AI) \cite{Russel2020book,Beck1998,Bengio2021} and operations research (OR) \cite{Korte2018}.
Among methodologies to solve combinatorial optimization problems, model-based paradigms such as mixed-integer programming (MIP) and constraint programming (CP) are particularly important as they represent steps toward the `holy grail' of declarative problem-solving \cite{Freuder1997}:
in model-based paradigms, a user formulates a problem as a mathematical model and then solves it using a general-purpose solver, i.e., a user just needs to define a problem to solve it.
Benefitting from this declarative nature, MIP and CP have been applied to a wide range of combinatorial optimization problems \cite{Hungerlander2018,Booth2016,Gadegaard2021,Saadaoui2019,Letchford2016,Ritt2018,Bukchin2018,Keha2009,Martin2021,Morin2018,Laborie2018}.

Dynamic programming (DP) \cite{Bellman1957} is a commonly used computational method to solve diverse decision-making problems, but little work has considered DP as a model-based paradigm for combinatorial optimization.
In DP, a problem is described by a Bellman equation, a recursive equation where the optimal objective value of the original problem is defined by the optimal objective values of subproblems (states).
Although a Bellman equation can be viewed as a declarative mathematical model, it has typically been solved by a problem-specific algorithm in previous work on DP for combinatorial optimization problems \cite{Held1962,Dumas1995,Gromicho2012,Righini2008,Righini2009,Lawler1964,DeLaBanda2007,DeLaBanda2011}.

We propose domain-independent dynamic programming (DIDP), a novel model-based paradigm for combinatorial optimization based on DP.
Our modeling language is inspired by domain-independent AI planning \cite{Ghalib2004}, where a problem is described by a common modeling language based on a state transition system.
At the same time, DIDP follows the approach of OR, which investigates efficient optimization models.
For example, in MIP, different constraints can result in different strengths of linear relaxations while sharing the same integer feasible region.
In CP, global constraints are specifically designed for common substructures of combinatorial optimization problems so that a solver can exploit them to achieve high performance.
Moreover, problem-specific DP methods for combinatorial optimization sometimes exploit problem-specific information to reduce the effort to solve the Bellman equations \cite{Dumas1995,Righini2008,Righini2009,DeLaBanda2007,DeLaBanda2011}.
Following these approaches, DIDP allows (but does not require) a user to explicitly include information that is implied by the problem definition and can be potentially exploited by a solver.
This design opens up the possibility of investigating the development of better models as in MIP and CP.
To solve the models, we develop general-purpose DIDP solvers using state space search, particularly heuristic search algorithms \cite{Russel2020,Pearl1982,Edelkamp2012}, which are also used in AI planners \cite{BonetG2001,HoffmannN01,Helmert06}.

\subsection{Related Work} \label{sec:related-work}

Some prior studies used DP as a problem-independent framework.
There are four main directions explored in the literature: theoretical formalisms for DP, automated DP in logic programming languages, DP languages or software for applications other than combinatorial optimization, and decision diagram-based solvers.
However, compared to DIDP, they are insufficient to be model-based paradigms for combinatorial optimization;
they either are purely theoretical, are not explicitly designed for combinatorial optimization, or require additional information specific to the solving algorithm.

Some problem-independent formalisms for DP have been developed, but they were not actual modeling languages.
In their seminal work, \citeauthor{Karp1967}~\cite{Karp1967} introduced a sequential decision process (sdp), a problem-independent formalism for DP based on a finite state automaton.
While they used sdp to develop DP algorithms problem-independently, they noted that ``In many applications, however, the state-transition representation is not the most natural form of problem statement'' and thus introduced a discrete decision process, a formalism to describe a combinatorial optimization problem, from which sdp is derived.
This line of research was further investigated by subsequent work \cite{Ibaraki1972,Ibaraki1973a,Ibaraki1973b,Ibaraki1974,Martelli1975b,Helman1982,Kumar1988}.
In particular, \citeauthor{Kumar1988}~\cite{Kumar1988} proposed a theoretical formalism based on context-free grammar.

In logic programming languages, a recursive function can be defined with memoization \cite{Michie1968} or tabling \cite{Bird1980,Tamaki1986} techniques that store the result of a function evaluation in memory and reuse it given the same arguments.
To efficiently perform memoization, \citeauthor{Puchinger2008}~\cite{Puchinger2008} proposed allowing a user to define a bound on the objective value to prune states, motivated by applications to combinatorial optimization.
Picat \cite{Picat} hybridizes logic programming with other paradigms including MIP, CP, and AI planning, and the AI planning module uses tabling with state pruning based on a user-defined objective bound.
Unlike DIDP, the above approaches cannot model dominance between different states since memoization or tabling can avoid the evaluation of a state only when the state is exactly the same as a previously evaluated one.

DP languages and software have been developed for multiple application fields.
Algebraic dynamic programming (ADP) \cite{Giegerich2002} is a software framework to formulate a DP model using context-free grammar.
It was designed for bioinformatics and was originally limited to DP models where a state is described by a string.
While ADP has been extended to describe a state using data structures such as a set, it is still focused on bioinformatics applications \cite{zuSiederdissen2015}.
Dyna is a declarative programming language for DP built on logic programming, designed for natural language processing and machine learning applications \cite{Eisner2005,Vieira2017}.
For optimal control, DP solvers have been developed in MATLAB \cite{Sundstorm2009,Miretti2021}.

\citeauthor{Hooker2013}~\cite{Hooker2013} pointed out that a decision diagram (DD), a data structure based on directed graphs, can be used to represent a DP model, and a solution can be extracted as a path in the DD.
While constructing such a DD may require large amount of computational space and time, \citeauthor{Bergman2016}~\cite{Bergman2016} proposed DD-based branch-and-bound, an algorithm to solve a DP model by repeatedly constructing relaxed and restricted DDs, which are approximations of the exact DD;
they are computationally cheaper to construct and give bounds on the optimal objective value.
To construct a relaxed DD, DD-based branch-and-bound requires a merging operator, a function to map two states to a single state.
The currently developed general-purpose solvers using DD-based branch-and-bound, ddo \cite{Bergman2016,Gillard2020} and CODD \cite{Michel2024}, require a user to provide a merging operator in addition to a DP model.
Therefore, compared to DIDP, they are DD solvers rather than model-based paradigms based on DP.

Several approaches do not fit in the above directions.
DP2PNSolver \cite{Lew2006} takes a DP model coded in a Java style language, gDPS, as input and compiles it to program code, e.g., Java code.
It is not designed to incorporate information more than a Bellman equation, and the output program code solves the Bellman equation by simple state enumeration.
DP was also used as a unified modeling format to combine CP with reinforcement learning \cite{Felix2021,Cappart2021}, but this approach is still in the scope of CP;
in their framework, a CP model based on a DP model is given to a CP solver, and a reinforcement learning agent trained in a Markov decision process based on the same DP model is used for a value selection heuristic of the CP solver.

Our modeling language is inspired by Planning Domain Definition Language (PDDL) \cite{Ghallab1998,FoxL03}, a standard modeling language for AI planning, which is based on a state transition system.
While PDDL and related languages such as PDDL+ \cite{Fox2006} and Relational Dynamic Influence Diagram Language (RDDL) \cite{Sanner2010} are dominant in AI planning, other modeling languages for state transition systems have been proposed in AI.
\citeauthor{Hernadvolgyi2000}~\cite{Hernadvolgyi2000} proposed PSVN, where a state is described by a fixed length vector.
PSVN is designed to allow the automatic generation of a heuristic function for heuristic search algorithms.
In the CP community, HADDOCK \cite{Gentzel2020}, a modeling language for DDs based on a state transition system, was proposed and used with a CP solver for constraint propagation.

\citeauthor{Martelli1975}~\cite{Martelli1975} and subsequent work \cite{Alfonso1979,Stefania1981} showed that a generalized version of heuristic search algorithms can be used for DP.
Unified frameworks for DP, heuristic search, and branch-and-bound were proposed by \citeauthor{Ibaraki1978}~\cite{Ibaraki1978} and \citeauthor{Kumar1988}~\cite{Kumar1988}.
\citeauthor{Holte2015}~\cite{Holte2015} discussed the similarity between abstraction heuristics for heuristic search and state space relaxation for DP.

\subsection{Contributions}

We summarize the contributions of this paper as follows:
\begin{itemize}
    \item We propose DIDP, a novel model-based paradigm for combinatorial optimization based on DP.
          While existing model-based paradigms such as MIP and CP use constraint-based representations of problems, DIDP uses a state-based representation.
    \item We develop a state-based modeling language for DIDP.
          Although our language is inspired by PDDL, an existing language for AI planning, it is specifically designed for combinatorial optimization and has features that PDDL does not have:
          a user can investigate efficient optimization models by incorporating redundant information such as dominance relation and bounds.
    \item We formally show that a class of state space search algorithms can be used as DIDP solvers under reasonable theoretical conditions.
          We implement such solvers in a software framework and demonstrate that they empirically outperform MIP, CP, and existing state-based solvers including AI planners in a number of combinatorial optimization problems.
          Since our framework is published as open-source software, AI researchers can use it as a platform to develop and apply state space search algorithms to combinatorial optimization problems.
\end{itemize}

\subsection{Overview}

In Section~\ref{sec:dp}, we introduce DP with an example.
In Section~\ref{sec:dypdl}, we define Dynamic Programming Description Language (DyPDL), a modeling formalism for DIDP.
We also present the theoretical properties of DyPDL including its computational complexity.
In Section~\ref{sec:yaml-dypdl}, we present YAML-DyPDL, a practical modeling language for DyPDL.
In Section~\ref{sec:heuristic-search}, we propose DIDP solvers.
First, we show that state space search can be used to solve DyPDL models under reasonable theoretical conditions.
In particular, we prove the completeness and optimality of state space search for DyPDL.
Then, we develop seven DIDP solvers using existing heuristic search algorithms.
In Section~\ref{sec:models}, we formulate DyPDL models for existing combinatorial optimization problem classes.
In Section~\ref{sec:experiment}, we experimentally compare the DIDP solvers with commercial MIP and CP solvers using the common benchmark instances of eleven problem classes.
We show that the best-performing DIDP solver outperforms both MIP and CP on seven problem classes while the worst performer does so on six of eleven.
We also demonstrate that DIDP achieves superior or competitive performance to existing state-based frameworks, domain-independent AI planners, Picat, and ddo.
Section~\ref{sec:conclusion} concludes the paper.

This paper is an extended version of two conference papers \cite{Kuroiwa2023CAASDy,Kuroiwa2023Anytime}, which lacked formal and theoretical aspects due to space limitations.
This paper has the following new contributions:
\begin{itemize}
    \item We formally define DyPDL and theoretically analyze it.
    \item We formally define heuristic search for DyPDL and show its completeness and optimality.
    \item We introduce DyPDL models for two additional problems, the orienteering problem with time windows and the multi-dimensional knapsack problem, and use them in the experimental evaluation.
          These two problems are maximization problems in contrast to the nine minimization problems used in the conference papers.
    \item We show the optimality gap achieved by DIDP solvers in the experimental evaluation.
          Our dual bound computation method for beam search is improved from our previous work \cite{Kuroiwa2024Parallel}.
    \item We empirically investigate the importance of heuristic functions in a DIDP solver.
    \item We empirically compare the DIDP solvers with domain-independent AI planners, Picat, and ddo.
\end{itemize}

\section{Dynamic Programming} \label{sec:dp}

Dynamic Programming (DP) is a computational problem-solving method, where a problem is recursively decomposed to subproblems, and each problem is represented by a \emph{state} \cite{Bellman1957}.
In this paper, we focus on DP to solve combinatorial optimization problems, where we make a finite number of discrete decisions to optimize an objective function.
In particular, we assume that a state is transformed into another state by making a decision, and a solution is a finite sequence of decisions.\footnote{
    We acknowledge that DP has more general applications beyond our assumptions;
    for example, when DP is applied to Markov decision processes, the outcome of a decision is represented by a probability distribution over multiple states, and a solution is a policy that maps a state to a probability distribution over decisions \cite{Bellman1957,Howard1960,Sutton2018}.
    However, this is not the topic of this work as we restrict ourselves to solutions that can be represented by a path of an alternating sequence of states and transitions.
}

One formal definition of such DP is the \emph{deterministic finite-state problem} \cite{Bertsekas1995}, represented by the following equation:
\begin{equation*}
    x_{k+1} = f_k(x_k, u_k), \quad k = 0, 1, ..., m-1
\end{equation*}
where $x_k$ is the state of the system at stage $k$, $u_k$ is the decision variable at stage $k$, $f_k$ is a function that describes the state transition, and $m$ is the horizon.
The objective is to make a sequence of decisions, corresponding to value assignments to $u_0, u_1, ..., u_{m-1}$, minimizing the total cost:
\begin{equation*}
    g_m(x_m) + \sum_{k=0}^{m-1} g_k(x_k, u_k)
\end{equation*}
where $g_k$ is the cost function at stage $k$, and $g_m$ is the terminal cost function.
We assume that the domains of $x_k$ and $u_k$ are finite for each $k$.
The optimal value of the cost function can also be described by a \emph{Bellman equation} \cite{Bellman1957}, a recursive definition of the value function $V$ that maps a state $x_k$ to the optimal cost to reach a terminal state:
\begin{align*}
    V(x_k) = \begin{cases}
                 g_m(x_m)                                    & \text{if } k = m \\
                 \min_{u_k} g_k(x_k, u_k) + V(f_k(x_k, u_k)) & \text{if } k < m
             \end{cases}
\end{align*}

As a running example, we use the traveling salesperson problem with time windows (TSPTW) \cite{Svelsbergh1985}.
In this problem, a set of customers $N = \{ 0, ..., n - 1 \}$ is given.
A solution is a tour starting from the depot (index $0$), visiting each customer exactly once, and returning to the depot.
Visiting customer $j$ from $i$ requires the travel time $c_{ij} \geq 0$.
In the beginning, $t=0$.
The visit to customer $i$ must be within a time window $[a_i, b_i]$.
Upon earlier arrival, waiting until $a_i$ is required.
The objective is to minimize the total travel time, excluding the waiting time.
Finding a valid tour for TSPTW is NP-complete \cite{Svelsbergh1985}.

\citeauthor{Dumas1995}~\cite{Dumas1995} applied DP to solve TSPTW.
In their approach, a state is a tuple of variables $(U, i, t)$, where $U$ is the set of unvisited customers, $i$ is the current location, and $t$ is the current time.
At each step, we consider visiting one of the unvisited customers as a decision.
The cost of a decision is the travel time to the next customer, and the state is updated accordingly.
In the deterministic finite-state problem, given $x_k = (U, i, t)$ for $k = 0, ..., n-2$, we have
\begin{align*}
     & u_k = j \in U \text{ s.t. } t + c_{ij} \leq b_{j}                       \\
     & f_k((U, i, t), j) = (U \setminus \{ j \}, j, \max\{ t + c_{ij}, a_j \}) \\
     & g_k((U, i, t), j) = c_{ij}
\end{align*}
with the horizon $m = n - 1$, the terminal state cost $g_{n-1}((U, i, t)) = c_{i0}$, and $x_0 = (N \setminus \{ 0 \}, 0, 0)$.
The Bellman equation is defined as follows:
\begin{align}
    \begin{split}
         & V(U, i, t) =           \begin{cases}
                                      c_{i0}                                                                                                    & \text{if } U = \emptyset     \\
                                      \min\limits_{j \in U : t + c_{ij} \leq b_j} c_{ij} + V(U \setminus \{ j \}, j, \max\{ t + c_{ij}, a_j \}) & \text{if } U \neq \emptyset. \\
                                  \end{cases}
    \end{split} \label{eqn:tsptw:transitions}
\end{align}
We assume the second line of Equation~\eqref{eqn:tsptw:transitions} to be $\infty$ if there is no $j \in U$ with $t + c_{ij} \leq b_j$, meaning that no feasible solution exists from state $(U, i, t)$.
The first line defines the terminal cost function based on the fact that the stage of state $(U, i, t)$ is $n - 1 - |U|$.

The state-based problem representation in DP is fundamentally different from existing model-based paradigms such as mixed-integer programming (MIP) and constraint programming (CP), which are constraint-based:
the problem is defined by a set of decision variables, constraints on the variables, and an objective function.
For example, a MIP model for TSPTW \cite{Hungerlander2018} uses a binary variable $x_{ij}$ to represent if customer $j$ is visited from $i$ and a continuous variable $t_i$ to represent the time to visit $i$.
The model includes constraints ensuring that each customer is visited exactly once ($\sum_{j \in N \setminus \{ i \}} x_{ij} = \sum_{j \in N \setminus \{ i \}} x_{ji} = 1$), and the time windows are satisfied ($a_i \leq t_i \leq b_i$).
The objective function is represented as $\sum_{i \in N} \sum_{j \in N \setminus \{ i \}} c_{ij} x_{ij}$.
Similarly, a CP model uses a variable to represent the time to visit each customer and constraints to ensure that the time window is satisfied \cite{Booth2016}.

\section{Dynamic Programming Description Language (DyPDL)} \label{sec:dypdl}

To shift from problem-specific DP methods to a declarative model-based paradigm based on DP, we formally define a class of models that we focus on in this paper.
We introduce a solver-independent theoretical formalism, Dynamic Programming Description Language (DyPDL).
Since DP is based on a state-based representation, we design DyPDL inspired by AI planning formalisms such as STRIPS \cite{FikesN1971} and SAS+ \cite{BackstromN95}, where a problem is described as a state transition system.
In DyPDL, a problem is described by states and transitions between states, and a solution corresponds to a sequence of transitions satisfying particular conditions.
DyPDL is also similar to the deterministic finite-state problem \cite{Bertsekas1995}, but it is different in the following aspects:
the number of decisions is not necessarily fixed in advance and the cost is not necessarily additive.

In DyPDL, a state is a complete assignment to state variables.

\begin{definition} \label{def:state-variables}
    A \emph{state variable} is either an \emph{element, set, or numeric variable}.
    An element variable $v$ has domain $D_v = \mathbb{Z}^+_0$ (nonnegative integers).
    A set variable $v$ has domain $D_v = 2^{\mathbb{Z}^+_0}$ (sets of nonnegative integers).
    A numeric variable $v$ has domain $D_v = \mathbb{Q}$ (rational numbers).
\end{definition}

\begin{definition} \label{def:state}
    Given a set of state variables $\mathcal{V} = \{ v_1, ..., v_n \}$, a \emph{state} is a tuple of values $S = (d_1, ..., d_n)$ where $d_i \in D_{v_i}$ for $i = 1, ..., n$.
    We denote the value $d_i$ of variable $v_i$ in state $S$ by $S[v_i]$ and the set of all states by $\mathcal{S} = D_{v_1} \times ... \times D_{v_n}$.
\end{definition}

\begin{example} \label{example:state-variables}
    In our TSPTW example in Equation~\eqref{eqn:tsptw:transitions}, a state is represented by three variables: a set variable $U$, an element variable $i$, and a numeric variable $t$.
\end{example}

A state can be transformed into another state by changing the values of the state variables.
To describe such changes, we define \emph{expressions}: functions returning a value given a state.

\begin{definition}
    An \emph{element expression} $e : \mathcal{S} \to \mathbb{Z}^+_0$ is a function that maps a state to a nonnegative value.
    A \emph{set expression} $e : \mathcal{S} \to 2^{\mathbb{Z}^+_0}$ is a function that maps a state to a set of nonnegative values.
    A \emph{numeric expression} $e : \mathcal{S} \to \mathbb{Q}$ is a function that maps a state to a numeric value.
    A \emph{condition} $c : \mathcal{S} \to \{ \bot, \top \}$ is a function that maps a state to a Boolean value.
    Given a state $S$, we denote $c(S) = \top$ by $S \models c$ and $c(S) = \bot$ by $S \not\models c$.
    For a set of conditions $C$, we denote $\forall c \in C, S \models c$ by $S \models C$ and $\exists c \in C, S \not\models c$ by $S \not\models C$.
\end{definition}

With the above expressions, we formally define transitions, which transform one state into another.
A transition has an effect to update state variables, preconditions defining when it is applicable, and a cost expression to update the objective value.

\begin{definition}
    A \emph{transition} $\tau$ is a 3-tuple $\langle \mathsf{eff}_\tau, \mathsf{cost}_\tau, \mathsf{pre}_\tau \rangle$ where $\mathsf{eff}_\tau $ is the \emph{effect}, $\mathsf{cost}_\tau$ is the \emph{cost expression}, and $\mathsf{pre}_\tau$ is the set of \emph{preconditions}.
    \begin{itemize}
        \item The effect $\mathsf{eff}_\tau$ is a tuple of expressions $(e_1, ..., e_n)$ where $e_i : \mathcal{S} \to D_{v_i}$.
              We denote the expression $e_i$ corresponding to a state variable $v_i$ by $\mathsf{eff}_\tau[v_i]$.
        \item The cost expression $\mathsf{cost}_\tau : \mathbb{Q} \times \mathcal{S} \to \mathbb{Q}$ is a function that maps a numeric value $x$ and a state $S$ to a numeric value $\mathsf{cost}_\tau(x, S)$.
        \item $\mathsf{pre}_\tau$ is a set of conditions, i.e., $c : \mathcal{S} \to \{ \bot, \top \}$ for each $c \in \mathsf{pre}_\tau$, and each of such a condition is called a precondition of $\tau$.
    \end{itemize}
    Given a set of transitions $\mathcal{T}$, the set of \emph{applicable} transitions $\mathcal{T}(S)$ in a state $S$ is defined as $\mathcal{T}(S) = \{ \tau \in \mathcal{T} \mid S \models \mathsf{pre}_\tau \}$.
    Given a state $S$ and a transition $\tau \in \mathcal{T}(S)$, the \emph{successor state} $S[\![\tau]\!]$, which results from applying $\tau$ in $S$, is defined as $S[\![\tau]\!][v] = \mathsf{eff}_\tau[v](S)$ for each variable $v$.
    For a sequence of transitions $\sigma = \langle \sigma_1, ..., \sigma_m \rangle$, the state $S[\![\sigma]\!]$, which results from applying $\sigma$ in $S$, is defined as $S[\![\sigma]\!] = S[\![\sigma_1]\!][\![\sigma_2]\!]...[\![\sigma_m]\!]$.
    If $\sigma$ is an empty sequence, i.e., $\sigma = \langle \rangle$, $S[\![\sigma]\!] = S$.
\end{definition}

In terms of the deterministic finite-state problem, a transition represents a value assignment to a decision variable as well as the state transition function and the cost function given the assignment.
In other words, to represent the deterministic finite-state problem in DyPDL, we define a transition $\tau$ for each possible value $\hat{u}_k$ of $u_k$.
Then, $S[\![\tau]\!] = f_k(S, \hat{u}_k)$, $\textsf{cost}_\tau(x, S) = x + g_k(S, \hat{u}_k)$, and $S \models \mathsf{pre}_\tau$ if and only if $\hat{u}_k$ is a valid value of $u_k$ in state $S$.

\begin{example} \label{example:transitions}
    In our TSPTW example in Equation~\eqref{eqn:tsptw:transitions}, we consider a transition $\tau_j$ representing the decision to visit customer $j$.
    The effect of $\tau_j$ is defined as $\mathsf{eff}_{\tau_j}[U](S) = S[U] \setminus \{ j \}$, $\mathsf{eff}_{\tau_j}[i](S) = j$, and $\mathsf{eff}_{\tau_j}[t](S) = \max\{ S[t] + c_{S[i],j}, a_j \}$.
    The preconditions are $S \mapsto j \in S[U]$ and $S \mapsto S[t] + c_{S[i],j} \leq b_j$.
    The cost expression is defined as $\mathsf{cost}_{\tau_j}(x, S) = c_{S[i],j} + x$.
\end{example}

We define a base case, where further transitions are not considered, and the cost is defined by a function of the state.
We use the term `base case' to represent the conditions where the recursion of a Bellman equation stops.

\begin{definition}[Base case] \label{def:base-case}
    A \emph{base case} $B$ is a tuple $\langle C_B, \mathsf{base\_cost}_B \rangle$ where $C_B$ is a set of conditions and $\mathsf{base\_cost}_B : \mathcal{S} \to \mathbb{Q}$ is a numeric expression.
    A state $S$ with $S \models C_B$ is called a \emph{base state}, and its \emph{base cost} is $\mathsf{base\_cost}_B(S)$.
\end{definition}

In terms of the deterministic finite-state problem, the base cost corresponds to the terminal cost function, i.e., $\mathsf{base\_cost}_B(S) = g_m(S)$.
However, while the base state is defined based on conditions in DyPDL, the terminal state is defined based on the horizon in the deterministic finite-state problem.
If we explicitly include the current stage $k$ as a state variable, we can define a base case $B$ with $C_B = \{ S \mapsto S[k] = m \}$.

\begin{example} \label{example:base-case}
    In our TSPTW example in Equation~\eqref{eqn:tsptw:transitions}, a base case $B$ is defined by $C_B = \{ S \mapsto S[U] = \emptyset \}$ and $\mathsf{base\_cost}_B(S) = c_{S[i],0}$.
\end{example}

Now, we define the state transition system based on the above definitions.
We additionally introduce the target state, which is the start of the state transition system, and the state constraints, which must be satisfied by all states.
The name `target state' is inspired by a Bellman equation; the target state is the target of the Bellman equation, corresponding to the original problem, to which we want to compute the optimal objective value.
The cost of a solution is computed by repeatedly applying the cost expression of a transition backward from the base cost;
this is also inspired by recursion in a Bellman equation, where the objective value of the current state is computed from the objective values of the successor states.

\begin{definition} \label{def:dypdl}
    A \emph{DyPDL model} is a tuple $\langle \mathcal{V}, S^0, \mathcal{T}, \mathcal{B}, \mathcal{C} \rangle$ with a specification of \emph{minimization} or \emph{maximization}, where $\mathcal{V}$ is the set of state variables, $S^0$ is a state called the \emph{target state}, $\mathcal{T}$ is the set of transitions, $\mathcal{B}$ is the set of base cases, and $\mathcal{C}$ is a set of conditions called \emph{state constraints}.

    A \emph{solution} for a DyPDL model is a sequence of transitions $\sigma = \langle \sigma_1, ..., \sigma_m \rangle$ such that
    \begin{itemize}
        \item All transitions are applicable, i.e., $\sigma_i \in \mathcal{T}(S^{i-1})$ where $S^i = S^{i-1}[\![\sigma_i]\!]$ for $i = 1, ..., m$.
        \item The target state and all intermediate states do not satisfy a base case, i.e., $\forall B \in \mathcal{B}, S^i \not\models C_B$ for $i = 0, ..., m-1$.
        \item The target state, all intermediate states, and the final state satisfy the state constraints, i.e., $S^i \models \mathcal{C}$ for $i = 0, ..., m$.
        \item The final state is a base state, i.e., $\exists B \in \mathcal{B}, S^m \models C_B$.
    \end{itemize}
    If the target state is a base state and satisfies the state constraints, an empty sequence $\langle \rangle$ is a solution.
    Given a state $S$, an \emph{$S$-solution} is a sequence of transitions satisfying the above conditions except that it starts from a state $S$ instead of the target state.

    For minimization, the cost of an $S$-solution $\sigma = \langle \sigma_1, ..., \sigma_m \rangle$ is recursively defined as follows:
    \begin{equation*}
        \mathsf{solution\_cost}(\sigma, S) = \begin{cases}
            \mathsf{cost}_{\sigma_1}(\mathsf{solution\_cost}(\langle \sigma_2, ..., \sigma_m \rangle, S[\![\sigma_1]\!]), S) & \text{ if } |\sigma| \geq 1 \\
            \min_{B \in \mathcal{B} : S \models C_B} \mathsf{base\_cost}_B(S)                                                & \text{ if } |\sigma| = 0    \\
        \end{cases}
    \end{equation*}
    where $|\sigma|$ is the number of transitions in $\sigma$, i.e., $|\sigma| = m$, and $\langle \sigma_2, ..., \sigma_m \rangle = \langle \rangle$ if $m = 1$.
    For maximization, we replace $\min$ with $\max$.
    For a solution for the DyPDL model ($S^0$-solution) $\sigma$, we denote its cost by $\mathsf{solution\_cost}(\sigma)$, omitting $S^0$.
    An \emph{optimal $S$-solution} for minimization (maximization) is an $S$-solution with its cost less (greater) than or equal to the cost of any $S$-solution.
    An \emph{optimal solution} for a DyPDL model is an optimal $S^0$-solution.
\end{definition}

In terms of the deterministic finite-state problem, because each transition $\sigma_i$ is defined with respect to a decision value of $u_i$, a solution for a DyPDL model $\langle \sigma_1, ..., \sigma_m \rangle$ corresponds to value assignments to the decision variables $u_0, ..., u_{m-1}$.

\begin{example} \label{example:dypdl}
    In our TSPTW example in Equation~\eqref{eqn:tsptw:transitions}, the state variables are $U$, $i$, and $t$ as in Example~\ref{example:state-variables}, the target state is $S^0 = (N \setminus \{ 0 \}, 0, 0)$, a transition to visit customer $j$ is defined for each $j \in N \setminus \{ 0 \}$ as in Example~\ref{example:transitions}, and one base case is defined as in Example~\ref{example:base-case}.
    Since each transition $\tau_j$ represents the decision to visit customer $j$, given a solution for the DyPDL model, which is a sequence of transitions, we can construct a tour for TSPTW.
    Given an $S$-solution $\langle \sigma_1, ..., \sigma_m \rangle$ with $m \geq 1$, $S = (U, i, t)$, and $\sigma_1 = \tau_j$ , its cost is defined as $\textsf{solution\_cost}(\langle \sigma_1, ..., \sigma_m \rangle, (U, i, t)) = c_{ij} + \textsf{solution\_cost}(\langle \sigma_2, ..., \sigma_m \rangle, (U \setminus \{ j \}, j, \max\{ t + c_{ij}, a_{j} \}))$.
    The optimization direction is minimization.
    We do not use state constraints in this example at this point and we show a use case later in Section~\ref{sec:redundant}.
\end{example}

\subsection{Complexity} \label{sec:complexity}

We have defined expressions as functions of states and have not specified further details.
Therefore, the complexity of an optimization problem with a DyPDL model depends on the complexity of evaluating expressions.
In addition, for example, if we have an infinite number of preconditions, evaluating the applicability of a single transition may not terminate in finite time.
Given these facts, we consider the complexity of a model whose definition is finite.

\begin{definition} \label{def:finitely-defined}
    A DyPDL model is \emph{finitely defined} if the following conditions are satisfied:
    \begin{itemize}
        \item The numbers of the state variables, transitions, base cases, and state constraints are finite.
        \item Each transition has a finite number of preconditions.
        \item Each base case has a finite number of conditions.
        \item All the effects, the cost expression, the preconditions of the transitions, the conditions and the costs of the base cases, and the state constraints can be evaluated in finite time.
    \end{itemize}
\end{definition}

Even with this restriction, finding a solution for a DyPDL model is an undecidable problem.
We show this by reducing one of the AI planning formalisms, which is undecidable, into a DyPDL model.
We define a numeric planning task and its solution in Definitions~\ref{def:numeric-planning} and \ref{def:numeric-planning-solution} following \citeauthor{Helmert02}~\cite{Helmert02}.

\begin{definition} \label{def:numeric-planning}
    A \emph{numeric planning task} is a tuple $\langle V_P, V_N, \textit{Init}, \textit{Goal}, \textit{Ops} \rangle$ where $V_P$ is a finite set of \emph{propositional variables}, $V_N$ is a finite set of \emph{numeric variables}, $\textit{Init}$ is a state called the \emph{initial state}, $\textit{Goal}$ is a finite set of \emph{propositional and numeric conditions}, and $\textit{Ops}$ is a finite set of \emph{operators}.
    \begin{itemize}
        \item A \emph{state} is defined by a pair of functions $(\alpha, \beta)$, where $\alpha: V_P \to \{ \bot, \top \}$ and $\beta: V_N \to \mathbb{Q}$.
        \item A propositional condition is written as $v = \top$ where $v \in V_P$.
              A state $(\alpha, \beta)$ satisfies it if $\alpha(v) = \top$.
        \item A numeric condition is written as $f(v_1, ..., v_n) \textbf{ relop } 0$ where $v_1, ..., v_n \in V_N$, $f : \mathbb{Q}^n \to \mathbb{Q}$ maps $n$ numeric variables to a rational number, and $\textbf{relop} \in \{ =, \neq, <, \leq, \geq, > \}$.
              A state $(\alpha, \beta)$ satisfies it if $f(\beta(v_1), ..., \beta(v_n)) \textbf{ relop } 0$.
    \end{itemize}

    An operator in $\textit{Ops}$ is a pair $\langle \textit{Pre}, \textit{Eff} \rangle$, where $\textit{Pre}$ is a finite set of conditions (\emph{preconditions}), and $\textit{Eff}$ is a finite set of \emph{propositional and numeric effects}.
    \begin{itemize}
        \item A propositional effect is written as $v \leftarrow t$ where $v \in V_P$ and $t \in \{ \bot, \top \}$.
        \item A numeric effect is written as $v \leftarrow f(v_1, ..., v_n)$ where $v, v_1, ..., v_n \in V_N$ and $f : \mathbb{Q}^n \to \mathbb{Q}$ maps $n$ numeric variables to a rational number.
    \end{itemize}

    All functions that appear in numeric conditions and numeric effects are restricted to functions represented by arithmetic operators $\{ +, -, \cdot, / \}$, but the divisor must be a non-zero constant.
\end{definition}

\begin{definition} \label{def:numeric-planning-solution}
    Given a numeric planning task $\langle V_P, V_N, \textit{Init}, \textit{Goal}, \textit{Ops} \rangle$, the \emph{state transition graph} is a directed graph where nodes are states and there is an edge $((\alpha, \beta), (\alpha', \beta'))$ if there exists an operator $\langle \textit{Pre}, \textit{Eff} \rangle \in \textit{Ops}$ satisfying the following conditions.
    \begin{itemize}
        \item $(\alpha, \beta)$ satisfies all conditions in $\textit{Pre}$.
        \item $\alpha'(v) = t$ if $v \leftarrow t \in \textit{Eff}$ and $\alpha'(v) = \alpha(v)$ otherwise.
        \item $\beta'(v) = f(\beta(v_1), ..., \beta(v_n))$ if $v \leftarrow f(v_1, ..., v_n) \in \textit{Eff}$ and $\beta'(v) = \beta(v)$ otherwise.
    \end{itemize}
    A \emph{solution} for the numeric planning task is a path from the initial state to a state that satisfies all goal conditions in $\textit{Goal}$ in the state transition graph.
\end{definition}

\begin{example} \label{example:numeric-planning}
    We represent TSPTW as a numeric planning task similar to the DyPDL model in Example~\ref{example:dypdl}.
    Unlike the DyPDL model, we ignore the objective value since the definition of a numeric planning task does not consider it.
    In our numeric planning task, a proposition $u_i = \top$ represents that customer $i$ is not visited, $v_i = \top$ represents that customer $i$ is visited, and $l_i = \top$ represents that the current location is customer $i$.
    We have only one numeric variable $t$ representing the current time.
    Therefore,
    \begin{itemize}
        \item $V_P = \{ u_1, ..., u_{n-1}, v_1, ..., v_{n-1}, l_0, ..., l_{n-1} \}$.
        \item $V_n = \{ t \}$.
        \item $\textit{Init} = (\alpha^0, \beta^0)$ such that $\forall i \in N \setminus \{ 0 \}, \alpha^0(u_i) = \top, \alpha^0(v_i) = \bot, \alpha^0(l_i) = \bot$, $\alpha^0(l_0) = \top$, and $\beta^0(t) = 0$.
        \item $\textit{Goal} = \{ v_1 = \top, ..., v_n = \top, l_0 = \top \}$.
    \end{itemize}
    To visit customer $j \in N \setminus \{ 0 \}$ from $i \in N$, we define two operators, one of which represents earlier arrival.
    In addition, we define an operator to return to the depot $0$ from customer $i$ after visiting all customers.
    We have the following three types of operators.
    \begin{itemize}
        \item $\textit{Pre} = \{ l_i = \top, u_j = \top, t + c_{ij} - a_j < 0 \}$ and $\textit{Eff} = \{ l_i \leftarrow \bot, l_j \leftarrow \top, u_j \leftarrow \bot, v_j \leftarrow \top, t \leftarrow a_j \}$.
        \item $\textit{Pre} = \{ l_i = \top, u_j = \top, t + c_{ij} - a_j \geq 0, t + c_{ij} - b_j \leq 0 \}$ and $\textit{Eff} = \{ l_i \leftarrow \bot, l_j \leftarrow \top, u_j \leftarrow \bot, v_j \leftarrow \top, t \leftarrow t + c_{ij} \}$.
        \item $\textit{Pre} = \{ l_i = \top, v_1 = \top, ..., v_{n-1} = \top \ \}$ and $\textit{Eff} = \{ l_i \leftarrow \bot, l_0 \leftarrow \top, t \leftarrow t + c_{i0} \}$.
    \end{itemize}
\end{example}

\citeauthor{Helmert02}~\cite{Helmert02} showed that finding a solution for the above-defined numeric planning task is undecidable.
To show the undecidability of DyPDL, we reduce a numeric planning task into a DyPDL model by replacing all propositional variables with a single set variable.

\begin{theorem} \label{thm:undecidable}
    Finding a solution for a finitely defined DyPDL model is undecidable.
\end{theorem}

\begin{proof}
    Let $\langle V_P, V_N, \textit{Init}, \textit{Goal}, \textit{Ops} \rangle$ be a numeric planning task.
    We compose a DyPDL model as follows:
    \begin{itemize}
        \item If $V_P \neq \emptyset$, we introduce a set variable $P'$ in the DyPDL model.
              For each numeric variable $v \in V_N$ in the numeric planning task, we introduce a numeric variable $v'$ in the DyPDL model.
        \item Let $(\alpha^0, \beta^0) = \textit{Init}$.
              We index propositional variables in $V_P$ using $i = 0, ..., |V_P|-1$ and denote the $i$-th variable by $u_i$.
              In the target state $S^0$ of the DyPDL model, $S^0[P'] = \{ i \in \{ 0, ..., |V_P|-1 \} \mid \alpha^0(u_i) = \top \}$ and $S^0[v'] = \beta^0(v)$ for each numeric variable $v \in V_N$.
        \item We introduce a base case $B = \langle C_B, 0 \rangle$ in the DyPDL model.
              For each propositional condition $u_i = \top$ in $\textit{Goal}$, we introduce a condition $S \mapsto i \in S[P']$ in $C_B$.
              For each numeric condition $f(v_1, ..., v_n) \textbf{ relop } 0$ in $\textit{Goal}$, we introduce a condition $S \mapsto f(S[v_1], ..., S[v_n]) \textbf{ relop } 0$ in $C_B$.
        \item For each operator $o = \langle \textit{Pre}, \textit{Eff} \rangle$, we introduce a transition $o' = \langle \mathsf{eff}_{o'}, \mathsf{cost}_{o'}, \mathsf{pre}_{o'} \rangle$ with $\mathsf{cost}_{o'}(x, S) = x + 1$.
        \item For each propositional condition $u_i = \top$ in $\textit{Pre}$, we introduce $S \mapsto i \in S[P
                          ']$ in $\mathsf{pre}_{o'}$.
              For each numeric condition $f(v_1, ..., v_n) \textbf{ relop }$ 0 in $\textit{Pre}$, we introduce $S \mapsto f(S[v'_1], ..., S[v'_n]) \textbf{ relop } 0$ in $\mathsf{pre}_{o'}$.
        \item Let $\textit{Add} = \{ i \in \{ 0, ..., |V_P|-1 \} \mid u_i \leftarrow \top \in \textit{Eff} \}$ and $\textit{Del} = \{ i \in \{ 0, ..., |V_P|-1 \} \mid u_i \leftarrow \bot \in \textit{Eff} \} $.
              We have $\mathsf{eff}_{o'}[P'](S) = (S[P'] \setminus \textit{Del}) \cup \textit{Add}$.
              We have $\mathsf{eff}_{o'}[v'](S) = f(S[v'_1], ..., S[v'_n])$ if $v \leftarrow f(v_1, ..., v_n) \in \textit{Eff}$ and $\mathsf{eff}_{o'}[v'](S) = S[v']$ otherwise.
        \item The set of state constraints is empty.
    \end{itemize}
    The construction of the DyPDL model is done in finite time.
    The numbers of propositional variables, numeric variables, goal conditions, transitions, preconditions, and effects are finite.
    Therefore, the DyPDL model is finitely defined.

    Let $\sigma = \langle o'_1, ..., o'_m \rangle$ be a solution for the DyPDL model.
    Let $S^j = S^{j-1}[\![o'_j]\!]$ for $j = 1, ..., m$.
    Let $(\alpha^j, \beta^j)$ be a numeric planning state such that $\alpha^j(u_i) = \top$ if $i \in S^j[P']$, $\alpha^j(u_i) = \bot$ if $i \notin S^j[P']$, and $\beta^j(v) = S^j[v']$.
    Note that $(\alpha^0, \beta^0) = \textit{Init}$ satisfies the above condition by construction.
    We prove that the state transition graph for the numeric planning task has edge $((\alpha^j, \beta^j), (\alpha^{j+1}, \beta^{j+1}))$ for $j = 0, ..., m-1$, and $(\alpha^m, \beta^m)$ satisfies all conditions in $\textit{Goal}$.

    Let $o_j = (\textit{Pre}_j, \textit{Eff}_j)$.
    Since $o'_j$ is applicable in $S^j$, for each propositional condition $u_i = \top$ in $\textit{Pre}_j$, the set variable satisfies $i \in S^j[P']$.
    For each numeric condition $f(v_1, ..., v_n) \textbf{ relop } 0$ in $\textit{Pre}_j$, the numeric variables satisfy $f(S^j[v'_1], ..., S^j[v'_n]) \textbf{ relop } 0$.
    By construction, $\alpha^j(u_i) = \top$ for $i \in S^j[P']$ and $f(\beta^j(v_1), ..., \beta^j(v_n)) \textbf{ relop } 0$.
    Therefore, $(\alpha^j, \beta^j)$ satisfies all conditions in $\textit{Pre}_j$.
    Similarly, $(\alpha^m, \beta^m)$ satisfies all conditions in $\textit{Goal}$ since $S^m$ satisfies all conditions in the base case.

    Let $\textit{Add}_j = \{ i \in \{ 0, ..., |V_P| - 1 \} \mid u_i \leftarrow \top \in \textit{Eff}_j \}$ and $\textit{Del}_j = \{ i \in \{ 0, ..., |V_P| - 1 \} \mid u_i \leftarrow \bot \in \textit{Eff}_j \} $.
    By construction, $S^{j+1}[P'] = (S^j[P'] \setminus \textit{Del}_j) \cup \textit{Add}_j$.
    Therefore, for $i$ with $u_i \leftarrow \bot \in \textit{Eff}_j$, we have $i \notin S^{j+1}[P']$, which implies $\alpha^{j+1}(u_i) = \bot$.
    For $i$ with $u_i \leftarrow \top \in \textit{Eff}_j$, we have $i \in S^{j+1}[P']$, which implies $\alpha^{j+1}(u_i) = \top$.
    For other $i$, $i \in S^{j+1}[P']$ if $i \in S^j[P']$ and $i \notin S^{j+1}[P']$ if $i \notin S^j[P']$, which imply $\alpha^{j+1}(u_i) = \alpha^j(u_i)$.
    For $v$ with $v \leftarrow f(v_1, ..., v_n) \in \textit{Eff}_j$, we have $S^{j+1}[v'] = f(S^j[v'_1], ..., S^j[v'_n]) = f(\beta^j(v_1), ..., \beta^j(v_n))$, which implies $\beta^{j+1}(v) = f(\beta^j(v_1), ..., \beta^j(v_n))$.
    For other $v$, we have $S^{j+1}[v'] = S^j[v'] = \beta^j(v)$, which implies $\beta^{j+1}(v) = \beta^j(v)$.
    Therefore, edge $((\alpha^j, \beta^j), (\alpha^{j+1}, \beta^{j+1}))$ exists in the state transition graph.

    Thus, by solving the DyPDL model, we can find a solution for the numeric planning task.
    Since the numeric planning task is undecidable, finding a solution for a DyPDL model is also undecidable.
\end{proof}

Note that applying this reduction to the numeric planning task in Example~\ref{example:numeric-planning} does not result in the DyPDL model in Example~\ref{example:dypdl}.
For example, Example~\ref{example:numeric-planning} defines two operators for each pair of customers and one operator for each customer to return to the depot, and thus DyPDL model constructed from this numeric planning task has $2n(n-1) + n - 1$ transitions in total.
In contrast, Example~\ref{example:dypdl} defines one transition for each customer, resulting in $n-1$ transitions in total.
This difference is due to the expressiveness of DyPDL, i.e., referring the value of a state variable in preconditions and effects, enables us to model the same problem with fewer transitions.

With the above proof, DyPDL can be considered a superset of the numeric planning formalism in Definition~\ref{def:numeric-planning}, which corresponds to a subset of Planning Domain Definition Language (PDDL) 2.1 \cite{FoxL03} called level 2 of PDDL 2.1 \cite{Helmert02}.
As we will present later, our practical modeling language for DyPDL enables a user to include dominance between states and bounds on the solution cost, which cannot be modeled in PDDL.
In contrast, full PDDL 2.1 has additional features such as durative actions, which cannot be modeled in DyPDL.

\citeauthor{Helmert02}~\cite{Helmert02} and subsequent work \cite{GnadHJS2023} proved undecidability for more restricted numeric planning formalisms than Definition~\ref{def:numeric-planning} such as subclasses of a restricted numeric planning task, where functions in numeric conditions are linear and numeric effects increase or decrease a numeric variable only by a constant.
We expect that these results can be easily applied to DyPDL since our reduction is straightforward.
Previous work also investigated conditions with which a numeric planning task becomes more tractable, e.g., decidable or PSPACE-complete \cite{Helmert02,ShleyfmanGJ2023,GiganteS2023}.
We also expect that we can generalize such conditions to DyPDL.
However, in this paper, we consider typical properties of DP models for combinatorial optimization problems.

In our TSPTW example, we have a finite number of states;
each state is the result of a sequence of visits to a finite number of customers.
In addition, such a sequence does not encounter the same state more than once since the cardinality of the set of unvisited customers, $U$, is strictly decreasing.
We formalize these properties as finiteness and acyclicity.
In practice, all DP models presented and evaluated in this paper are finite and acyclic.

First, we introduce reachability.
A state $S'$ is reachable from another state $S$ if there exists a sequence of transitions that transforms $S$ into $S'$.
The definition is similar to that of an $S$-solution.

\begin{definition}[Reachability] \label{def:reachability}
    Let $S$ be a state satisfying the state constraints.
    A state $S'$ is \emph{reachable} from a state $S$ with a non-empty sequence of transitions $\sigma = \langle \sigma_1, ..., \sigma_m \rangle$ if the following conditions are satisfied:
    \begin{itemize}
        \item All transitions are applicable, i.e., $\sigma_1 \in \mathcal{T}(S)$ and $\sigma_{i+1} \in \mathcal{T}(S^i)$ where $S^1 = S[\![\sigma_1]\!]$ and $S^{i+1} = S^i[\![\sigma_{i+1}]\!]$ for $i = 1, ..., m-1$.
        \item All intermediate states do not satisfy a base case, i.e., $\forall B \in \mathcal{B}, S^i \not\models C_B$ for $i = 1, ..., m-1$.
        \item All intermediate states satisfy the state constraints, i.e., $S^i \models \mathcal{C}$ for $i = 1, ..., m-1$.
        \item The final state is $S'$, i.e., $S' = S^m$.
    \end{itemize}
    We say that $S'$ is reachable from $S$ if there exists a non-empty sequence of transitions with the above conditions.
    We say that $S$ is a reachable state if it is the target state or reachable from the target state.
\end{definition}

\begin{definition} \label{def:finite}
    A DyPDL model is \emph{finite} if it is finitely defined, and the set of reachable states is finite.
\end{definition}

\begin{definition} \label{def:acyclic}
    A DyPDL model is \emph{acyclic} if any reachable state is not reachable from itself.
\end{definition}

If a model is finite, we can enumerate all reachable states and check if there is a base state in finite time.
If there is a reachable base state, and the model is acyclic, then there are a finite number of solutions, and each solution has a finite number of transitions.
Therefore, by enumerating all sequences with which a state is reachable, identifying solutions, and computing their costs, we can find an optimal solution in finite time.

\begin{theorem} \label{thm:acyclic-decidable}
    A finite and acyclic DyPDL model has an optimal solution, or the model is infeasible.
    A problem to decide if a solution whose cost is less (greater) than a given rational number exists for minimization (maximization) is decidable.
\end{theorem}

\subsection{The Bellman Equation for DyPDL} \label{sec:bellman}

A Bellman equation is useful to succinctly represent a DP model, illustrated by our example DP model for TSPTW (Equation~\eqref{eqn:tsptw:transitions}).
Here, we make an explicit connection between DyPDL and a Bellman equation.
Focusing on finite and acyclic DyPDL models, which are typical properties in combinatorial optimization problems, we present the Bellman equation representing the optimal solution cost.\footnote{While a Bellman equation can be used for a cyclic model by considering a fixed point, we do not consider it in this paper as such a model is not used in this paper, and it complicates our discussion.}
A Bellman equation requires a special cost structure, the Principle of Optimality \cite{Bellman1957}.
We formally define it in the context of DyPDL.
Intuitively, an optimal $S$-solution must be constructed from an optimal $S[\![\tau]\!]$-solution, where $\tau$ is a transition applicable in $S$.

\begin{definition} \label{def:principle}
    Given any reachable state $S$ and an applicable transition $\tau \in \mathcal{T}(S)$, a DyPDL model satisfies the \emph{Principle of Optimality} if for any $x, y \in \mathbb{Q}$, $x \leq y \rightarrow \mathsf{cost}_\tau(x, S) \leq \mathsf{cost}_\tau(y, S)$.
\end{definition}

With this property, we give the Bellman equation for a DyPDL model, defining the value function $V$ that maps a state to the optimal $S$-solution cost or $\infty$ ($-\infty$) if an $S$-solution does not exist in minimization (maximization).

\begin{theorem} \label{thm:bellman}
    Consider a finite, acyclic, and minimization DyPDL model $\langle \mathcal{V}, S^0, \mathcal{T}, \mathcal{B}, \mathcal{C} \rangle$ satisfying the Principle of Optimality.
    For each reachable state $S$, there exists an optimal $S$-solution with a finite number of transitions, or there does not exist an $S$-solution.
    Let $\mathcal{S}'$ be a set of reachable states, and $V : \mathcal{S}' \to \mathbb{Q} \cup \{ \infty \}$ be a function of a state that returns $\infty$ if there does not exist an $S$-solution or the cost of an optimal $S$-solution otherwise.
    Then, $V$ satisfies the following equation:
    \begin{equation}
        V(S) = \begin{cases} \label{eqn:min-bellman}
            \infty                                                                          & \text{if } S \not\models \mathcal{C}                                       \\
            \min\limits_{B \in \mathcal{B} : S \models C_B} \mathsf{base\_cost}_B(S)        & \text{else if } \exists B \in \mathcal{B}, S \models C_B                   \\
            \min\limits_{\tau \in \mathcal{T}(S)} \mathsf{cost}_{\tau}(V(S[\![\tau]\!]), S) & \text{else if } \exists \tau \in \mathcal{T}(S), V(S[\![\tau]\!]) < \infty \\
            \infty                                                                          & \text{else.}
        \end{cases}
    \end{equation}
    For maximization, we replace $\infty$ with $-\infty$, $\min$ with $\max$, and $<$ with $>$.
\end{theorem}

\begin{proof}
    Since the model is acyclic, we can define a partial order over reachable states where $S$ precedes its successor state $S[\![\tau]\!]$ if $\tau \in \mathcal{T}(S)$.
    We can sort reachable states topologically according to this order.
    Since the set of reachable states is finite, there exists a state that does not precede any reachable state.
    Let $S$ be such a state.
    Then, one of the following holds: $S \not\models \mathcal{C}$, $S$ is a base state, or $\mathcal{T}(S) = \emptyset$ by Definition~\ref{def:reachability}.
    If $S \not\models \mathcal{C}$, there does not exist an $S$-solution and $V(S) = \infty$, which is consistent with the first line of Equation~\eqref{eqn:min-bellman}.
    If $S$ satisfies a base case, since the only $S$-solution is an empty sequence by Definition~\ref{def:dypdl}, $V(S) = \min_{B \in \mathcal{B} : S \models C_B} \mathsf{base\_cost}_B(S)$, which is consistent with the second line of Equation~\eqref{eqn:min-bellman}.
    If $S$ is not a base state and $\mathcal{T}(S) = \emptyset$, then there does not exist an $S$-solution and $V(S) = \infty$, which is consistent with the fourth line of Equation~\eqref{eqn:min-bellman}.

    Assume that for each reachable state $S[\![\tau]\!]$ preceded by a reachable state $S$ in the topological order, one of the following conditions holds:
    \begin{enumerate}
        \item There does not exist an $S[\![\tau]\!]$-solution, i.e., $V(S[\![\tau]\!]) = \infty$.
        \item There exists an optimal $S[\![\tau]\!]$-solution with a finite number of transitions with cost $V(S[\![\tau]\!]) < \infty$.
    \end{enumerate}
    If the first case holds for each $\tau \in \mathcal{T}(S)$, there does not exist an $S$-solution, and $V(S) = \infty$.
    Since $V(S[\![\tau]\!]) = \infty$ for each $\tau$, $V(S) = \infty$ is consistent with the fourth line of Equation~\eqref{eqn:min-bellman}.
    The first case of the assumption also holds for $S$.
    If the second case holds for some $\tau$, there exists an optimal $S[\![\tau]\!]$-solution with a finite number of transitions and the cost $V(S[\![\tau]\!]) < \infty$.
    By concatenating $\tau$ and the optimal $S[\![\tau]\!]$-solution, we can construct an $S$-solution with a finite number of transitions and the cost $\mathsf{cost}_{\tau}(V(S[\![\tau]\!]), S)$.
    This $S$-solution is better or equal to any other $S$-solution $\sigma = \langle \sigma_1, ..., \sigma_m \rangle$ starting with $\sigma_1 = \tau$ since
    \begin{equation*}
        \mathsf{solution\_cost}(\sigma, S) = \mathsf{cost}_{\tau}(\mathsf{solution\_cost}(\langle \sigma_2, ..., \sigma_m \rangle, S[\![\tau]\!]), S) \geq \mathsf{cost}_{\tau}(V(S[\![\tau]\!]), S)
    \end{equation*}
    by the Principle of Optimality (Definition~\ref{def:principle}).
    By considering all possible $\tau$,
    \begin{equation*}
        V(S) = \min\limits_{\tau \in \mathcal{T}(S)} \mathsf{cost}_{\tau}(V(S[\![\tau]\!]), S),
    \end{equation*}
    which is consistent with the third line of Equation~\eqref{eqn:min-bellman}.
    The second case of the assumption also holds for $S$.
    We can prove the theorem by mathematical induction.
    The proof for maximization is similar.
\end{proof}

\subsection{Redundant Information} \label{sec:redundant}

In AI planning, a model typically includes only information necessary to define a problem \cite{Ghallab1998,McDermott2000}.
In contrast, in operations research (OR), an optimization model often includes redundant information implied by the other parts of the model, e.g., valid inequalities in MIP.
While such information is unnecessary to define a model, it sometimes improves the performance of solvers.
Redundant information has also been exploited by some problem-specific DP methods in previous work \cite{Dumas1995,Puchinger2008,DeLaBanda2011}.
For example, \citeauthor{Dumas1995}~\cite{Dumas1995} used redundant information implied by the Bellman equation for TSPTW (Equation~\eqref{eqn:tsptw:transitions}).
Given a state $(U, i, t)$, if a customer $j \in U$ cannot be visited by the deadline $b_j$ even if we use the shortest path with travel time $c^*_{ij}$, then the state does not lead to a solution, so it is ignored.
While this technique was algorithmically used in previous work, we can represent it declaratively using the value function and formulate it as a state constraint.
\begin{equation}
    V(U, i, t) = \infty \text{ if } \exists j \in U, t + c^*_{ij} > b_j. \label{eqn:tsptw:infeasibility}
\end{equation}

\begin{example}
    In our TSPTW Example, Inequality~\eqref{eqn:tsptw:infeasibility} can be represented by a set of state constraints $\{ S \mapsto j \not\in U \lor S[t] + c^*_{S[i],j} \leq b_j \mid j \in N \setminus \{ 0 \} \}$.
\end{example}

While we have represented Equation~\eqref{eqn:tsptw:infeasibility} using state constraints, we can also consider other types of redundant information.
We introduce three new concepts, state dominance, dual bound functions, and forced transitions, to declaratively represent such information.

\subsubsection{State Dominance}

Another technique used by \citeauthor{Dumas1995}~\cite{Dumas1995} is state dominance.
If we have two states $(U, i, t)$ and $(U, i, t')$ with $t \leq t'$, then $(U, i, t)$ leads to at least as good a solution as $(U, i, t')$, so the latter is ignored.
In terms of the value function:
\begin{equation}
    V(U, i, t) \leq V(U, i, t') \text{ if } t \leq t'. \label{eqn:tsptw:dominance}
\end{equation}

We define state dominance for DyPDL.
Intuitively, one state $S$ dominates another state $S'$ if $S$ always leads to an as good or a better solution, and thus $S'$ can be ignored when we have $S$.
In addition to this intuition, we require that $S$ does not leads to a longer solution.
This requirement is to ensure that we preserve at least one state leading to a base state after pruning states by dominance.
For example, suppose a maximization problem represented by $V(S) = \max_{\tau \in \mathcal{T}(S)} p_\tau + V(S[\![\tau]\!])$ with $p_\tau > 0$ and $V(S) = 0$ for a base state $S$.
In this problem, since $V(S) > V(S[\![\tau]\!])$ for any $\tau \in \mathcal{T}(S)$, $V(S)$ is maximized by the target state $S^0$, and thus all states would be dominated by $S^0$.
However, if we ignore all states except for $S^0$, we cannot find any solution.
Therefore, we introduce a condition that prevents pruning states closer to a base state.

\begin{definition}[State Dominance] \label{def:dominance}
    For a minimization DyPDL model, a state $S$ \emph{dominates} another state $S'$, denoted by $S' \preceq S$, iff, for any $S'$-solution $\sigma'$, there exists an $S$-solution $\sigma$ such that
    $\mathsf{solution\_cost}(\sigma, S) \leq \mathsf{solution\_cost}(\sigma', S')$ and $\sigma$ has no more transitions than $\sigma'$.
    For maximization, we replace $\leq$ with $\geq$.
\end{definition}

Our definition of dominance is inspired by simulation-based dominance in AI planning \cite{Torralba015}.
In that paradigm, $S$ dominates $S'$ only if for each applicable transition $\tau'$ in $S'$, there exists an applicable transition $\tau$ in $S$ such that $S[\![\tau]\!]$ dominates $S'[\![\tau']\!]$.
Also, if $S$ dominates a base state $S'$ (a goal state in planning terminology), $S$ is also a base state.
In addition to the original transitions, simulation-based dominance adds a NOOP transition, which stays in the current state.
In simulation-based dominance, if we have an $S'$-solution, we also have $S$-solution with an equal number of transitions (or possibly fewer transitions if we exclude NOOP).
Therefore, intuitively, Definition~\ref{def:dominance} is a generalization of simulation-based dominance;
a formal discussion is out of the scope of this paper.

In practice, it may be difficult to always detect if one state dominates another or not, and thus an algorithm may focus on dominance that can be easily detected based on a sufficient condition, e.g., $t \leq t'$ for $(U, i, t)$ and $(U, i, t')$.
We define an approximate dominance relation to represent such a strategy.
First, we clarify the condition that should be satisfied by an approximate dominance relation.

\begin{theorem} \label{thm:dominance-preorder}
    For a DyPDL model, the dominance relation is a preorder, i.e., the following conditions hold.
    \begin{itemize}
        \item $S \preceq S$ for a state $S$ (reflexivity).
        \item $S'' \preceq S' \land S' \preceq S \rightarrow S'' \preceq S$ for states $S$, $S'$, and $S''$ (transitivity).
    \end{itemize}
\end{theorem}

\begin{proof}
    The first condition holds by Definition~\ref{def:dominance}.
    For the second condition, for any $S''$-solution $\sigma''$, there exists an equal or better $S'$-solution $\sigma'$ having no more transitions than $\sigma''$.
    There exists an equal or better $S$-solution $\sigma$ for $\sigma'$ having no more transitions than $\sigma'$.
    Therefore, the cost of $\sigma$ is equal to or better than $\sigma''$, and $\sigma$ has no more transitions than $\sigma''$.
\end{proof}

\begin{definition} \label{def:approx-dominance}
    For a DyPDL model, an \emph{approximate dominance relation} $\preceq_a$ is a preorder over two states such that $S' \preceq_a S \rightarrow S' \preceq S$ for reachable states $S$ and $S'$.
\end{definition}

\begin{example}
    For our TSPTW example, Inequality~\eqref{eqn:tsptw:dominance} is represented by an approximate dominance relation such that $S' \preceq_a S$ iff $S[U] = S'[U]$, $S[i] = S'[i]$, and $S[t] \leq S'[t]$.
    \begin{itemize}
        \item The reflexivity holds since $S[U] = S[U]$, $S[i] = S[i]$, and $S[t] \leq S[t]$.
        \item The transitivity holds since $S[U] = S'[U]$ and $S'[U] = S''[U]$ imply $S[U] = S''[U]$, $S[i] = S'[i]$ and $S'[i] = S''[i]$ imply $S[i] = S''[i]$, and $S[t] \leq S'[t]$ and $S'[t] \leq S''[t]$ imply $S[t] \leq S''[t]$.
        \item All $S$- and $S'$-solutions have the same number of transitions since each solution has $|S[U]| = |S'[U]|$ transitions to visit all unvisited customers.
    \end{itemize}
\end{example}

An approximate dominance relation is sound but not complete:
it always detects the dominance if two states are the same and otherwise may produce a false negative but never a false positive.

Similarly to Inequality~\eqref{eqn:tsptw:dominance}, we can represent approximate dominance relation in DyPDL using the value function.
In what follows, we assume that $-\infty < x < \infty$ for any $x \in \mathbb{Q}$.

\begin{theorem}
    Given a finite and acyclic DyPDL model satisfying the Principle of Optimality, let $V$ be the value function of the Bellman equation for minimization.
    Given an approximate dominance relation $\preceq_a$, for reachable states $S$ and $S'$,
    \begin{equation} \label{eqn:dominance}
        V(S) \leq V(S') \text{ if } S' \preceq_a S.
    \end{equation}
    For maximization, we replace $\leq$ with $\geq$.
\end{theorem}

\begin{proof}
    For reachable states $S$ and $S'$ with $S' \preceq_a S$, assume that there exist $S$- and $S'$-solutions.
    Then, $V(S)$ ($V(S')$) is the cost of an optimal $S$-solution ($S'$-solution).
    By Definitions~\ref{def:dominance} and \ref{def:approx-dominance}, for minimization, an optimal $S$-solution has an equal or smaller cost than any $S'$-solution, so $V(S) \leq V(S')$.
    If there does not exist an $S$-solution, by Definition~\ref{def:dominance}, there does not exist an $S'$-solution, so $V(S) = V(S') = \infty$, thus $V(S) \leq V(S')$.
    The proof for the maximization is similar.
\end{proof}

\subsubsection{Dual Bound Function}

While not used by \citeauthor{Dumas1995}~\cite{Dumas1995}, bounds on the value function have been used by previous problem-specific DP methods \cite{Puchinger2008,DeLaBanda2011}.
When we know an upper bound on the optimal solution cost for minimization, and we can prove that a state cannot lead to a solution with its cost less than the upper bound, we can ignore the state.
For our TSPTW example, we define a lower bound function based on the one used for a sequential ordering problem by previous work \cite{Libralesso2020}.
Since the minimum travel time to visit customer $j$ is $c^{\text{in}}_j = \min_{k \in N \setminus \{ j \}} c_{kj}$, the cost of visiting all customers in $U$ and returning to the depot is at least the sum of $c^{\text{in}}_j$ for each $j \in U \cup \{ 0 \}$.
Similarly, we also use the minimum travel time from $j$, $c^{\text{out}}_j = \min_{k \in N \setminus \{ j \}} c_{jk}$, to underestimate the total travel time.
Using the value function,
\begin{equation}
    V(U, i, t) \geq \max\left\{ \sum_{j \in U \cup \{ 0 \} } c^{\text{in}}_j, \sum_{j \in U \cup \{ i \}} c^{\text{out}}_j \right\}. \label{eqn:tsptw:bound}
\end{equation}

We formalize this technique as a dual bound function that underestimates (overestimates) the cost of a solution in minimization (maximization).
We use the name `dual' since it is a lower bound for minimization and an upper bound for maximization, similar to dual bounds in mathematical optimization.

\begin{definition} \label{def:dual-bound}
    For a DyPDL model, a function $\eta : \mathcal{S} \mapsto \mathbb{Q} \cup \{ \infty, -\infty \}$ is a \emph{dual bound function} iff, for any reachable state $S$ and any $S$-solution $\sigma$, $\eta(S)$ is a \emph{dual bound} on $\mathsf{solution\_cost}(\sigma, S)$, i.e., $\eta(S) \leq \mathsf{solution\_cost}(\sigma, S)$ for minimization.
    For maximization, we replace $\leq$ with $\geq$.
\end{definition}

\begin{example}
    For our TSPTW example, Inequality~\eqref{eqn:tsptw:bound} is represented as $\eta(S) = \max\left\{ \sum\limits_{j \in S[U] \cup \{ 0 \}} c^{\text{in}}_j, \sum\limits_{j \in S[U] \cup \{ S[i] \}} c^{\text{out}}_j  \right\}$.
\end{example}

A function that always returns $-\infty$ ($\infty$) for minimization (maximization) is trivially a dual bound function.
If there exists an $S$-solution $\sigma$ for minimization, $\eta(S) \leq \mathsf{solution\_cost}(\sigma, S) < \infty$.
Otherwise, $\eta(S)$ can be any value, including $\infty$.
Thus, if a dual bound function can detect that an $S$-solution does not exist, the function should return $\infty$ to tell a solver that there is no $S$-solution.
For maximization, a dual bound function should return $-\infty$ in such a case.

\begin{theorem}
    Given a finite and acyclic DyPDL model satisfying the Principle of Optimality, let $V$ be the value function of the Bellman equation for minimization.
    Given a dual bound function $\eta$, for a reachable state $S$,
    \begin{equation} \label{eqn:dual-bound}
        V(S) \geq \eta(S).
    \end{equation}
    For maximization, we replace $\geq$ with $\leq$.
\end{theorem}

\begin{proof}
    For a reachable state $S$, if there exists an $S$-solution, the cost of an optimal $S$-solution is $V(S)$.
    By Definition~\ref{def:dual-bound}, $\eta(S)$ is a lower bound of the cost of any $S$-solution, so $\eta(S) \leq V(S)$.
    Otherwise, $\eta(S) \leq V(S) = \infty$.
    The proof for maximization is similar.
\end{proof}

\subsubsection{Forced Transitions} \label{sec:forced}

We introduce yet another type of redundant information, forced transitions.
Since forced transitions are not identified in our TSPTW example, we first present a motivating example, the talent scheduling problem \cite{Cheng1993}.
In this problem, a set of actors $A$ and a set of scenes $N$ are given, and the objective is to find a sequence of scenes to shoot to minimize the total cost paid for the actors.
In a scene $s \in N$, a set of actors $A_s \subseteq A$ plays for $d_s$ days.
We incur cost $c_a$ of actor $a$ for each day they are on location.
If an actor plays on days $i$ and $j$, they are on location on days $i, i + 1, ..., j$ even if they do not play on day $i + 1$ to $j - 1$.
The objective is to find a sequence of scenes such that the total cost is minimized.

\citeauthor{DeLaBanda2011}~\cite{DeLaBanda2011} proposed a DP method for talent scheduling, where a state is represented by a set of unscheduled scenes $Q$.
The actors who played in scenes $N \setminus Q$ have already arrived at the location and are staying there if they play in a scene in $Q$.
Therefore, the set of actors remaining on location is
\begin{equation*}
    L(Q) = \left( \bigcup_{s \in Q} A_s \cap \bigcup_{s \in N \setminus Q } A_s \right).
\end{equation*}
At each step, we shoot a scene $s$ from $Q$ and pay the cost $d_s \sum_{a \in A_s \cup L(Q)} c_a$.
The Bellman equation is as follows:
\begin{equation}
    V(Q) = \begin{cases}
        0                                                                                 & \text{if } Q = \emptyset     \\
        \min\limits_{s \in Q} d_s \sum_{a \in A_s \cup L(Q)} c_a + V(Q \setminus \{ s \}) & \text{if } Q \neq \emptyset.
    \end{cases}
\end{equation}
In this model, we can shoot a scene $s$ without paying an extra cost if $A_s = L(Q)$.
Therefore, if such an scene exists, we should shoot it next ignoring other scenes.
Using $b_s = d_s \sum_{a \in A_s} c_s$, which can be precomputed,
\begin{equation} \label{eqn:talent-forced}
    V(Q) = b_s + V(Q \setminus \{ s \}) \text{ if } \exists s \in Q, A_s = L(Q).
\end{equation}
We can exploit more redundant information in the preconditions of the transitions and a dual bound function, following \citeauthor{DeLaBanda2011}~\cite{DeLaBanda2011}.
We present the complete model with such information in \ref{sec:cp-models}.

We formalize Equation~\eqref{eqn:talent-forced} as forced transitions in DyPDL.
Similar to state dominance, we also require that a forced transition does not lead to a longer solution.

\begin{definition} \label{def:forced}
    Given a set of applicable transitions $\mathcal{T}(S)$ in a state $S$, a transition $\tau \in \mathcal{T}(S)$ is a \emph{forced transition} if an $S$-solution does not exists, or for each $S$-solution $\sigma'$, there exists an $S$-solution $\sigma$ whose first transition is $\tau$ and satisfies $\mathsf{solution\_cost}(\sigma, S) \leq \mathsf{solution\_cost}(\sigma', S)$ for minimization ($\leq$ is replaced with $\geq$ for maximization) and $|\sigma| \leq |\sigma'|$.
\end{definition}

\begin{theorem}
    Given a finite and acyclic DyPDL model satisfying the Principle of Optimality, let $V$ be the value function of the Bellman equation for minimization.
    Let $S$ be a reachable state with $V(S) < \infty$ for minimization or $V(S) > -\infty$ for maximization and $\tau$ be a forced transition applicable in a state $S$.
    Then,
    \begin{equation} \label{eqn:forced}
        V(S) = \mathsf{cost}_\tau(V(S[\![\tau]\!]), S).
    \end{equation}
\end{theorem}

\begin{proof}
    We assume minimization, and the proof for maximization is similar.
    Since $V(S) < \infty$, there exists an $S$-solution.
    Since the model is finite and acyclic, there are a finite number of $S$-solutions, and we can find an optimal $S$-solution.
    By Definition~\ref{def:forced}, there exists an optimal $S$-solution $\langle \tau, \sigma_1, ..., \sigma_m \rangle$.
    Since $V(S)$ is the cost of an optimal $S$-solution,
    \begin{equation*}
        V(S) = \mathsf{solution\_cost}(\langle \tau, \sigma_1, ..., \sigma_m \rangle, S) = \mathsf{cost}_\tau(\mathsf{solution\_cost}(\langle \sigma_1, ..., \sigma_m \rangle, S[\![\tau]\!]), S).
    \end{equation*}
    Since an $S[\![\tau]\!]$-solution $\langle \sigma_1, ..., \sigma_m \rangle$ exists, there exists an optimal $S[\![\tau]\!]$-solution $\langle \sigma'_1, ..., \sigma'_{m'} \rangle$ with cost $V(S[\![\tau]\!]) \leq \mathsf{solution\_cost}(\langle \sigma_1, ..., \sigma_m \rangle, S[\![\tau]\!])$.
    By the Principle of Optimality (Definition~\ref{def:principle}),
    \begin{equation*}
        V(S) \geq \mathsf{cost}_\tau(V(S[\![\tau]\!]), S).
    \end{equation*}
    Since $\langle \tau, \sigma'_1, ..., \sigma'_{m'} \rangle$ is also an $S$-solution,
    \begin{equation*}
        V(S) \leq \mathsf{solution\_cost}(\langle \tau, \sigma'_1, ..., \sigma'_{m'} \rangle, S) = \mathsf{cost}_\tau(\mathsf{solution\_cost}(\langle \sigma'_1, ..., \sigma'_{m'} \rangle, S[\![\tau]\!]), S) = \mathsf{cost}_\tau(V(S[\![\tau]\!]), S).
    \end{equation*}
    Therefore, $V(S) = \mathsf{cost}_\tau(V(S[\![\tau]\!]), S)$.
\end{proof}

\section{YAML-DyPDL: A Practical Modeling Language} \label{sec:yaml-dypdl}

As a practical modeling language for DyPDL, we propose YAML-DyPDL on top of a data serialization language, YAML 1.2.\footnote{\url{https://yaml.org/}}
YAML-DyPDL is inspired by PDDL in AI planning \cite{Ghallab1998}.
However, in PDDL, a model typically contains only information necessary to define a problem, while YAML-DyPDL allows a user to explicitly model redundant information, i.e., implications of the definition.
Such is the standard convention in OR and is commonly exploited in problem-specific DP algorithms for combinatorial optimization (e.g., \citeauthor{Dumas1995}~\cite{Dumas1995}).
In particular, in YAML-DyPDL, a user can explicitly define an approximate dominance relation (Definition~\ref{def:approx-dominance}), dual bound functions (Definition~\ref{def:dual-bound}), and forced transitions (Definition~\ref{def:forced}).
In PDDL, while forced transitions may be realized with preconditions of actions, approximate dominance relation and dual bound functions cannot be modeled.

In the DyPDL formalism, expressions and conditions are defined as functions.
In a practical implementation, the kinds of functions that can be used as expressions are defined by the syntax of a modeling language.
In YAML-DyPDL, for example, arithmetic operations (e.g., addition, subtraction, multiplication, and division) and set operations (e.g., adding an element, removing an element, union, intersection, and difference) using state variables can be used.
We give an example of YAML-DyPDL here.
A detailed description of the syntax is given as software documentation in our repository.\footnote{\url{https://github.com/domain-independent-dp/didp-rs/blob/main/didp-yaml/docs/dypdl-guide.md}}

\subsection{Example}

\begin{figure}[h!]
    \centering
    \scriptsize
    \begin{minipage}[t]{0.4\linewidth}
\begin{lstlisting}[lastline=41,frame=single]
cost_type: integer
reduce: min
objects:
  - customer
state_variables:
  - name: U
    type: set
    object: customer
  - name: i
    type: element
    object: customer
  - name: t
    type: integer
    preference: less
tables:
  - name: a
    type: integer
    args:
      - customer
  - name: b
    type: integer
    args:
      - customer
  - name: c
    type: integer
    args:
      - customer
      - customer
  - name: cstar
    type: integer
    args:
      - customer
      - customer
  - name: cin
    type: integer
    args:
      - customer
  - name: cout
    type: integer
    args:
      - customer
\end{lstlisting}
    \end{minipage}
    \hfil
    \begin{minipage}[t]{0.4\linewidth}
\begin{lstlisting}[firstline=42,firstnumber=42,frame=single]
transitions:
  - name: visit
    parameters:
      - name: j
        object: U
    effect:
      U: (remove j U)
      i: j
      t: (max (+ t (c i j)) (a j))
    cost: (+ (c i j) cost)
    preconditions:
      - (<= (+ t (c i j)) (b j))
constraints:
  - condition: (<= (+ t (cstar i j)) (b j))
    forall:
      - name: j
        object: U
base_cases:
  - conditions:
      - (is_empty U)
    cost: (c i 0)
dual_bounds:
  - (+ (sum cin U) (cin 0))
  - (+ (sum cout U) (cout i))
\end{lstlisting}
    \end{minipage}
    \caption{YAML-DyPDL domain file for TSPTW.}
    \Description{YAML-DyPDL domain file for TSPTW.}
    \label{lst:domain}
\end{figure}

We present how the DyPDL model in Example~\ref{example:dypdl} is described by YAML-DyPDL.
Following PDDL, we require two files, a domain file and a problem file, to define a DyPDL model.
A domain file describes a class of problems by declaring state variables and constants and defining transitions, base cases, and dual bound functions using expressions.
In contrast, a problem file describes one problem instance by defining information specific to that instance, e.g., the target state and the values of constants.

Figure~\ref{lst:domain} shows the domain file for the DyPDL model of TSPTW.
The domain file is a map in YAML, which associates keys with values.
In YAML, a key and a value are split by \textsf{:}.
Keys and values can be maps, lists of values, strings, integers, and floating-point numbers.
A list is described by multiple lines starting with \textsf{-}, and each value after \textsf{-} is an element of the list.
In YAML, we can also use a JSON-like syntax,\footnote{\url{https://www.json.org/json-en.html}} where a map is described as \textsf{\{ key\_1: value\_1, ..., key\_n: value\_n \}}, and a list is described as \textsf{[value\_1, ..., value\_n]}.

\subsubsection{Cost Type}

The first line defines key \textsf{cost\_type} and its value \textsf{integer}, meaning that the cost of the DyPDL model is computed in integers.
While the DyPDL formalism considers numeric expressions that return a rational number, in a software implementation, it is beneficial to differentiate integer and continuous values.
In YAML-DyPDL, we explicitly divide numeric expressions into integer and continuous expressions.
The value of the key \textsf{reduce} is \textsf{min}, which means that we want to minimize the cost.
\subsubsection{Object Types}

The key \textsf{objects}, whose value is a list of strings, defines \emph{object types}.
In the example, the list only contains one value, \textsf{customer}.
An object type is associated with a set of $n$ nonnegative integers $\{ 0, ..., n-1 \}$, where $n$ is defined in a problem file.
The \textsf{customer} object type represents a set of customers $N = \{ 0, ..., n-1 \}$ in TSPTW.
The object type is used later to define a set variable and constants.

\subsubsection{State Variables}

The key \textsf{state\_variables} defines state variables.
The value is a list of maps describing a state variable.
For each state variable, we have key \textsf{name} defining the name and key \textsf{type} defining the type, which is either \textsf{element}, \textsf{set}, \textsf{integer}, or \textsf{continuous}.

The variable \textsf{U} is the set variable $U$ representing the set of unvisited customers.
YAML-DyPDL requires associating a set variable with an object type.
The variable $U$ is associated with the object type, \textsf{customer}, by \textsf{object: customer}.
Then, the domain of $U$ is restricted to $2^N$.
This requirement arises from practical implementations of set variables;
we want to know the maximum cardinality of a set variable to efficiently represent it in a computer program (e.g., using a fixed length bit vector).

The variable \textsf{i} is the element variable $i$ representing the current location.
YAML-DyPDL also requires associating an element variable with an object type for readability;
by associating an element variable with an object type, it is easier to understand the meaning of the variable.
However, the domain of the element variable is not restricted by the number of objects, $n$; while objects are indexed from $0$ to $n-1$, a user may want to use $n$ to represent none of them.

The variable \textsf{t} is the numeric variable $t$ representing the current time.
For this variable, the preference is defined by \textsf{preference: less}, which means that a state having smaller $t$ dominates another state if $U$ and $i$ are the same.
Such a variable is called a \emph{resource variable}.
Resource variables define approximate dominance relation in Definition~\ref{def:approx-dominance}:
given two states $S$ and $S'$, if a state  $S[v] \geq S'[v]$ for each resource variable $v$ where greater is preferred (\textsf{preference: greater}), $S[v] \leq S'[v]$ for each resource variable where less is preferred (\textsf{preference: less}), and $S[v] = S'[v]$ for each non-resource variable $v$, then $S$ dominates $S'$.
This relation trivially satisfies reflexivity and transitivity, and thus it is a preorder.

\subsubsection{Tables}

The value of the key \textsf{tables} is a list of maps declaring tables of constants.
A table maps a tuple of objects to a constant.
The table \textsf{a} represents the beginning of the time window $a_j$ at customer $j$, so the values in the table are integers (\textsf{type: integer}).
The concrete values are given in a problem file.
The key \textsf{args} defines the object types associated with a table using a list.
For \textsf{a}, one customer $j$ is associated with the value $a_j$, so the list contains only one string \textsf{customer}.
The tables \textsf{b}, \textsf{cin}, and \textsf{cout} are defined for the deadline $b_j$, the minimum travel time to a customer $c^{\text{in}}_j$, and the minimum travel time from a customer $c^{\text{out}}_j$, respectively.
The table \textsf{c} is for $c_{kj}$, the travel time from customer $k$ to $j$.
This table maps a pair of customers to an integer value, so the value of \textsf{args} is a list equivalent to \textsf{[customer, customer]}.
Similarly, the shortest travel time $c^*_{kj}$ is represented by the table \textsf{cstar}.

\subsubsection{Transitions} \label{sec:yaml-transitions}

The value of the key \textsf{transitions} is a list of maps defining transitions.
Using \textsf{parameters}, we can define multiple transitions in the same scheme but associated with different objects.
The key \textsf{name} defines the name of the parameter, \textsf{j}, and \textsf{object} defines the object type.
Basically, the value of the key \textsf{object} should be the name of the object type, e.g., \textsf{customer}.
However, we can also use the name of a set variable.
In the example, by using \textsf{object: U}, we state that the transition is defined for each object $j \in N$ with a precondition $j \in U$.

The key \textsf{preconditions} defines preconditions by using a list of conditions.
In YAML-DyPDL, conditions and expressions are described by arithmetic operations in a LISP-like syntax.
In the precondition of the transition in our example, \textsf{(c i j)} corresponds to $c_{ij}$, so \textsf{($<=$ (+ t (c i j)) (b j))} corresponds to $t + c_{ij} \leq b_j$.
The key \textsf{effect} defines the effect by using a map, whose keys are names of the state variables.
For set variable \textsf{U}, the value is a set expression \textsf{(remove j U)}, corresponding to $U \setminus \{ j \}$.
For element variable \textsf{i}, the value is an element expression \textsf{j}, corresponding to $j$.
For integer variable \textsf{t}, the value is an integer expression \textsf{(max (+ t (c i j)) (a j))}, corresponding to $\max\{ t + c_{ij}, a_j \}$.
The key \textsf{cost} defines the cost expression \textsf{(+ (c i j) cost)}, corresponding to $c_{ij} + x$.
In the example, the cost expression must be an integer expression since the \textsf{cost\_type} is \textsf{integer}.
In the cost expression, we can use \textsf{cost} to represent the cost of the successor state ($x$).
We can also have a key \textsf{forced}, whose value is Boolean, indicating that transition is known to be a forced transition when it is applicable.
We do not have it in the example, which means the transition is not known to be forced.

\subsubsection{State Constraints}

The value of the key \textsf{constraints} is a list of state constraints.
In the DyPDL model, we have $\forall j \in U, t + c^*_{ij} \leq b_j$.
Similarly to the definition of transitions, we can define multiple state constraints with the same scheme associated with different objects using \textsf{forall}.
The value of the key \textsf{forall} is a map defining the name of the parameter and the associated object type or set variable.
The value of the key \textsf{condition} is a string describing the condition, \textsf{($<=$ (+ t (cstar i j)) (b j))}, which uses the parameter \textsf{j}.

\subsubsection{Base Cases}

The value of the key \textsf{base\_cases} is a list of maps defining base cases.
Each map has two keys, \textsf{conditions} and \textsf{cost}.
The value of the key \textsf{conditions} is a list of conditions, and the value of the key \textsf{cost} is a numeric expression (must be an integer expression in the example since \textsf{cost\_type} is \textsf{integer}).
The condition \textsf{(is\_empty U)} corresponds to $U = \emptyset$, and the cost \textsf{(c i 0)} corresponds to $c_{i0}$.

\subsubsection{Dual Bound Functions}

The value of the key \textsf{dual\_bounds} is a list of numeric expressions describing dual bound functions.
In the example, we use \textsf{(+ (sum cin U) (cin 0))} and \textsf{(+ (sum cout U) (cout i))} corresponding to $\sum_{j \in U} c^{\text{in}}_j + c^{\text{in}}_0 = \sum_{j \in U \cup \{ 0 \}} c^{\text{in}}_j$ and $\sum_{j \in U} c^{\text{out}}_j + c^{\text{out}}_i = \sum_{j \in U \cup \{ i \}} c^{\text{out}}_j$, respectively.
Since \textsf{cost\_type} is \textsf{integer}, they are integer expressions.

\subsubsection{Problem File}

\begin{figure}[t]
    \centering
    \scriptsize
    \fbox{\begin{minipage}[t]{0.55\linewidth}
\begin{lstlisting}[lastline=40]
object_numbers:
  customer: 4
target:
  U: [1, 2, 3]
  i: 0
  t: 0
table_values:
  a: { 1: 5, 2: 0, 3: 8 }
  b: { 1: 16, 2: 10, 3: 14 }
  c:
    {
      [0, 1]: 3, [0, 2]: 4, [0, 3]: 5,
      [1, 0]: 3, [1, 2]: 5, [1, 3]: 4,
      [2, 0]: 4, [2, 1]: 5, [2, 3]: 3,
      [3, 0]: 5, [3, 1]: 4, [3, 2]: 3,
    }
  cstar:
    {
      [0, 1]: 3, [0, 2]: 4, [0, 3]: 5,
      [1, 0]: 3, [1, 2]: 5, [1, 3]: 4,
      [2, 0]: 4, [2, 1]: 5, [2, 3]: 3,
      [3, 0]: 5, [3, 1]: 4, [3, 2]: 3,
    }
  cin: { 0: 3, 1: 3, 2: 3, 3: 3 }
  cout: { 0: 3, 1: 3, 2: 3, 3: 3 }
\end{lstlisting}
        \end{minipage}}
    \caption{YAML-DyPDL problem file for TSPTW.}
    \Description{YAML-DyPDL problem file for TSPTW.}
    \label{lst:problem}
\end{figure}

Turning to the problem file (Figure~\ref{lst:problem}), the value of \textsf{object\_numbers} is a map defining the number of objects for each object type.
The value of \textsf{target} is a map defining the values of the state variables in the target state.
For the set variable \textsf{U}, a list of nonnegative integers is used to define a set of elements in the set.
The value of \textsf{table\_values} is a map defining the values of the constants in the tables.
For \textsf{a}, \textsf{b}, \textsf{cin}, and \textsf{cout}, a key is the index of an object, and a value is an integer.
For \textsf{c} and \textsf{cstar}, a key is a list of the indices of objects.

\subsection{Complexity}

In Section~\ref{sec:dypdl}, we showed that finding a solution for a DyPDL model is undecidable in general by reducing a numeric planning task to a DyPDL model.
YAML-DyPDL has several restrictions compared to Definition~\ref{def:dypdl}.
A set variable is associated with an object type, restricting its domain to a subset of a given finite set.
In addition, expressions are limited by the syntax.
However, these restrictions do not prevent the reduction.

\begin{theorem}
    Finding a solution for a finitely defined DyPDL model is undecidable even with the following restrictions.
    \begin{itemize}
        \item The domain of each set variable $v$ is restricted to $2^{N_v}$ where $N_v = \{ 0, ..., n_v - 1 \}$, and $n_v$ is a positive integer.
        \item Numeric expressions and element expressions are functions represented by arithmetic operations $\{ +, -, \cdot, / \}$.
        \item Set expressions are functions constructed by a set of constants, set variables, and the intersection, union, and difference of two set expressions.
        \item A condition compares two numeric expressions, compares two element expressions, or checks if a set expression is a subset of another set expression.
    \end{itemize}
\end{theorem}

\begin{proof}
    We can follow the proof of Theorem~\ref{thm:undecidable} even with the restrictions.
    Since the number of propositional variables in the set $V_P$ in a numeric planning task is finite, we can use $n_{P'} = |V_P|$ for the set variable $P'$ representing propositional variables.
    Arithmetic operations $\{ +, -, \cdot, / \}$ are sufficient for numeric expressions by Definition~\ref{def:numeric-planning}.
    Similarly, if we consider a condition $i \in S[P']$, which checks if $i$ is included in a set variable $P'$, as $\{ i \} \subseteq S[P']$, the last two restrictions do not prevent the compilation of the numeric planning task to the DyPDL model.
\end{proof}

With the above reduction, we can use a system to solve a YAML-DyPDL model for the numeric planning formalism in Theorem~\ref{thm:undecidable}.

\section{State Space Search for DyPDL} \label{sec:heuristic-search}

We use state space search \cite{Russel2020}, which finds a path in an implicitly defined graph, to solve a DyPDL model.
In particular, we focus on heuristic search algorithms \cite{Pearl1984,Edelkamp2012}, which estimate path costs using a heuristic function.
Once we interpret the state transition system defined by a DyPDL model as a graph, it is intuitive that we can use state space search to solve the model.
However, in practice, state space search is not always applicable; a DyPDL model needs to satisfy particular conditions.
Therefore, in what follows, we formally present state space search algorithms and the conditions with which they can be used to solve DyPDL models.

\begin{definition} \label{def:stg}
    Given a DyPDL model, \emph{the state transition graph} is a directed graph where nodes are reachable states and there is an edge from $S$ to $S'$ labeled with $\tau$, $(S, S', \tau)$, iff $\tau \in \mathcal{T}(S)$ and  $S' = S[\![\tau]\!]$.
\end{definition}

We use the term \emph{path} to refer to both a sequence of edges in the state transition graph and a sequence of transitions as they are equivalent.
A state $S'$ is reachable from $S$ iff there exists a path from $S$ to $S'$ in the state transition graph, and an $S$-solution corresponds to a path from $S$ to a base state.
Trivially, if a model is acyclic, the state transition graph is also acyclic.

\subsection{Cost Algebras} \label{sec:cost-algebra}

For a DyPDL model, we want to find a solution that minimizes or maximizes the cost.
Shortest path algorithms such as Dijkstra's algorithm \cite{Dijkstra1959} and A* \cite{Hart1968} find the path minimizing the sum of the weights associated with the edges.
In DyPDL, the cost of a solution can be more general, defined by cost expressions of the transitions.
\citeauthor{Edelkamp2005}~\cite{Edelkamp2005} extended the shortest path algorithms to cost-algebraic heuristic search algorithms, which can handle more general cost structures.
They introduced the notion of cost algebras, which define the cost of a path using a binary operator to combine edge weights and an operation to select the best value.
Following their approach, first, we define a monoid.

\begin{definition} \label{def:monoid}
    Let $A$ be a set, $\times: A \times A \rightarrow A$ be a binary operator, and $\mathbf{1} \in A$.
    A tuple $\langle A, \times, \mathbf{1} \rangle$ is a \emph{monoid} if the following conditions are satisfied.
    \begin{itemize}
        \item $x \times y \in A$ for $x, y \in A$ (closure).
        \item $x \times (y \times z) = (x \times y) \times z$ for $x, y, z, \in A$ (associativity).
        \item $x \times \mathbf{1} = \mathbf{1} \times x = x$ for $x \in A$ (identity).
    \end{itemize}
\end{definition}

Next, we define isotonicity, a property of a set and a binary operator with regard to comparison.
Since minimization or maximization over rational numbers is sufficient for our use case, we restrict the set $A$ to rational numbers, and the comparison operator to $\leq$.
The original paper by \citeauthor{Edelkamp2005}~\cite{Edelkamp2005} is more general.

\begin{definition}[Isotonicity] \label{def:isotone}
    Given a set $A \subseteq \mathbb{Q} \cup \{ -\infty, \infty \}$ and a binary operator $\times: A \times A \rightarrow A$, $A$ is \emph{isotone} if
    $x \leq y \rightarrow x \times z \leq y \times z$ and
    $x \leq y \rightarrow z \times x \leq z \times y$
    for $x, y, z \in A$.
\end{definition}

With a monoid and isotonicity, we define a cost algebra.

\begin{definition} \label{def:cost-algebra}
    Let $\langle A, \times, \mathbf{1} \rangle$ be a monoid where $A \subseteq \mathbb{Q} \cup \{ -\infty, \infty \}$ is isotone.
    The monoid $\langle A, \times, \mathbf{1} \rangle$ is a \emph{cost algebra} if $\forall x \in A, \mathbf{1} \leq x$ for minimization or $\forall x \in A, \mathbf{1} \geq x$ for maximization.
\end{definition}

\subsection{Cost-Algebraic DyPDL Models} \label{sec:cost-algebraic-dypdl}

To apply cost-algebraic heuristic search, we focus on DyPDL models where cost expressions satisfy particular conditions.
First, we define a monoidal DyPDL model, where cost expressions are represented by a binary operator in a monoid.

\begin{definition} \label{def:monoidal-model}
    Let $\langle A, \times, \mathbf{1} \rangle$ be a monoid where $A \subseteq \mathbb{Q} \cup \{ -\infty, \infty \}$.
    A DyPDL model $\langle \mathcal{V}, S^0, \mathcal{T}, \mathcal{B}, \mathcal{C} \rangle$ is \emph{monoidal} with $\langle A, \times, \mathbf{1} \rangle$ if the cost expression of every transition $\tau \in \mathcal{T}$ is represented as $\mathsf{cost}_\tau(x, S) = w_\tau(S) \times x$ where $w_\tau : \mathcal{S} \to A \setminus \{ -\infty, \infty \}$ is a numeric expression , and the cost $\mathsf{base\_cost}_B$ of each base case $B \in \mathcal{B}$ returns a value in $A \setminus \{ -\infty, \infty \}$.
\end{definition}

We also define a cost-algebraic DyPDL model, which requires stricter conditions.

\begin{definition} \label{def:cost-algebaic-model}
    A monoidal DyPDL model $\langle \mathcal{V}, S^0, \mathcal{T}, \mathcal{B}, \mathcal{C} \rangle$ with a monoid $\langle A, \times, \mathbf{1}\rangle$ is \emph{cost-algebraic} if $\langle A, \times, \mathbf{1}\rangle$ is a cost algebra.
\end{definition}

For example, the DP model for TSPTW is cost-algebraic with a cost algebra $\langle \mathbb{Q}_0^+, +, 0 \rangle$ since the cost expression of each transition is defined as $(x, S) \mapsto c_{S[i],j} + x$ with $c_{S[i],j} \geq 0$.

When a model is monoidal, we can associate a weight to each edge in the state transition graph.
The weight of a path can be computed by repeatedly applying the binary operator to the weights of the edges in the path.

\begin{definition} \label{def:path-cost}
    Given a monoidal DyPDL model with $\langle A, \times, \mathbf{1} \rangle$, the \emph{weight of an edge} $(S, S', \tau)$ is $w_\tau(S)$.
    The \emph{weight of a path} $\langle (S, S^1, \sigma_1), (S^1, S^2, \sigma_2), ..., (S^{m-1}, S^m, \sigma_m) \rangle$  defined by a sequence of transitions $\sigma$ is
    \begin{equation*}
        w_\sigma(S) = w_{\sigma_1}(S) \times w_{\sigma_2}(S^1) \times ... \times w_{\sigma_m}(S^{m-1}).
    \end{equation*}
    For an empty path $\langle \rangle$, the weight is $\mathbf{1}$.
\end{definition}

The order of applications of the binary operator $\times$ does not matter due to the associativity.
Differently from the original cost-algebraic heuristic search, the weight of a path corresponding to an $S$-solution may not be equal to the cost of the $S$-solution in Definition~\ref{def:dypdl} due to our inclusion of the cost of a base state.
In the following lemma, we associate the weight of a path with the cost of a solution.

\begin{lemma} \label{lem:path-cost}
    Given a monoidal DyPDL model with a monoid $\langle A, \times, \mathbf{1} \rangle$ and a state $S$, let $\sigma$ be an $S$-solution.
    For minimization, $\mathsf{solution\_cost}(\sigma, S) = w_\sigma(S) \times \min_{B \in \mathcal{B} : S[\![\sigma]\!] \models C_B} \mathsf{base\_cost}_B(S[\![\sigma]\!])$.
    For maximization, we replace $\min$ with $\max$.
\end{lemma}

\begin{proof}
    If $\sigma$ is an empty sequence, since $w_\sigma(S) = \mathbf{1}$ and $S[\![\sigma]\!] = S$,
    \begin{equation*}
        w_\sigma(S) \times \min_{B \in \mathcal{B} : S[\![\sigma]\!] \models C_B} \mathsf{base\_cost}_B(S[\![\sigma]\!]) = \min_{B \in \mathcal{B} : S \models C_B} \mathsf{base\_cost}_B(S) = \mathsf{solution\_cost}(\sigma, S)
    \end{equation*}
    by Definition~\ref{def:dypdl}.
    Otherwise, let $\sigma = \langle \sigma_1, ..., \sigma_m \rangle$, $S^1 = S[\![\sigma_1]\!]$, and $S^{i+1} = S^i[\![\sigma_{i+1}]\!]$ for $i = 1, ..., m-1$.
    Following Definitions~\ref{def:dypdl} and \ref{def:monoidal-model},
    \begin{equation*}
        \mathsf{solution\_cost}(\sigma, S) = \mathsf{cost}_{\sigma_1}(\mathsf{solution\_cost}(\langle \sigma_2, ..., \sigma_m \rangle, S^1), S) = w_{\sigma_1}(S) \times \mathsf{solution\_cost}(\langle \sigma_2, ..., \sigma_m \rangle, S^1).
    \end{equation*}
    For $2 \leq i \leq m$, we get
    \begin{equation*}
        \begin{split}
            \mathsf{solution\_cost}(\langle \sigma_i, ..., \sigma_m \rangle, S^{i-1}) & = \mathsf{cost}_{\sigma_i}(\mathsf{solution\_cost}(\langle \sigma_{i+1}, ..., \sigma_m \rangle, S^i), S^{i-1}) \\
                                                                                      & = w_{\sigma_i}(S^{i-1}) \times \mathsf{solution\_cost}(\langle \sigma_{i+1}, ..., \sigma_m \rangle, S^i).
        \end{split}
    \end{equation*}
    For $i=m+1$, $\mathsf{solution\_cost}(\langle \rangle, S^m) = \min_{B \in \mathcal{B} : S^m \models C_B} \mathsf{base\_cost}_B(S^m)$.
    Thus, we get the equation in the lemma by Definition~\ref{def:path-cost}.
    The proof for maximization is similar.
\end{proof}

We show that isotonicity is sufficient for the Principle of Optimality in Definition~\ref{def:principle}.
First, we prove its generalized version in Theorem~\ref{thm:principle}.
In what follows, we denote the concatenation of sequences of transitions $\sigma$ and $\sigma'$ by $\langle \sigma; \sigma' \rangle$.

\begin{theorem} \label{thm:principle}
    Consider a monoidal DyPDL model with $\langle A, \times, \mathbf{1} \rangle$ such that $A \subseteq \mathbb{Q} \cup \{ -\infty, \infty \}$ and $A$ is isotone.
    Let $S'$ and $S''$ be states reachable from $S$ with sequences of transitions $\sigma'$ and $\sigma''$, respectively, with $w_{\sigma'}(S) \leq w_{\sigma''}(S)$.
    For minimization, if there exist $S'$- and $S''$-solutions $\sigma^1$ and $\sigma^2$ with $\mathsf{solution\_cost}(\sigma^1, S') \leq \mathsf{solution\_cost}(\sigma^2, S'')$, then $\langle \sigma'; \sigma^1 \rangle$ and $\langle \sigma''; \sigma^2 \rangle$, are $S$-solutions with $\mathsf{solution\_cost}(\langle \sigma'; \sigma^1 \rangle, S) \leq \mathsf{solution\_cost}(\langle \sigma''; \sigma^2 \rangle, S)$.
    For maximization, we replace $\leq$ with $\geq$.
\end{theorem}

\begin{proof}
    The sequences $\langle \sigma'; \sigma^1 \rangle$ and $\langle \sigma''; \sigma^2 \rangle$ are $S$-solutions by Definition~\ref{def:dypdl}.
    Since $\mathsf{solution\_cost}(\langle \sigma'; \sigma^1 \rangle, S) = w_{\sigma'}(S) \times \mathsf{solution\_cost}(\sigma^1, S')$ and $A$ is isotone,
    \begin{equation*}
        \begin{split}
            \mathsf{solution\_cost}(\langle \sigma'; \sigma^1 \rangle, S) & = w_{\sigma'}(S) \times \mathsf{solution\_cost}(\sigma^1, S') \leq w_{\sigma''}(S) \times \mathsf{solution\_cost}(\sigma^1, S')      \\
                                                                          & \leq w_{\sigma''}(S) \times \mathsf{solution\_cost}(\sigma^2, S'') = \mathsf{solution\_cost}(\langle \sigma''; \sigma^2 \rangle, S).
        \end{split}
    \end{equation*}
    The proof for maximization is similar.
\end{proof}

\begin{corollary}
    Let $\langle A, \times, \mathbf{1} \rangle$ be a monoid where $A \subseteq \mathbb{Q} \cup \{ -\infty, \infty \}$ and $A$ is isotone.
    A monoidal DyPDL model with $\langle A, \times, \mathbf{1} \rangle$ satisfies the Principle of Optimality in Definition~\ref{def:principle}.
\end{corollary}

Intuitively, the Principle of Optimality or its sufficient condition, isotonicity, ensures that an optimal path can be constructed by extending an optimal subpath.
Thus, a state space search algorithm can discard suboptimal paths to reach each node in the state transition graph.
Without this property, a state space search algorithm may need to enumerate suboptimal paths to a node since they may lead to an optimal path to a goal node.

\subsection{Formalization of Heuristic Search for DyPDL} \label{sec:dypdl-heuristic-search}

A state space search algorithm searches for a path between nodes in a graph.
In particular, we focus on unidirectional search algorithms, which visit nodes by traversing edges from one node (the initial node) to find a path to one of the nodes satisfying particular conditions (goal nodes).
Moreover, we focus on heuristic search algorithms, which use a heuristic function to estimate the path cost from a state to a goal node using a heuristic function $h$.
For a node $S$, a unidirectional heuristic search algorithms maintains $g(S)$ (the $g$-value), the best path cost from the initial node to $S$, and $h(S)$ (the $h$-value), the estimated path cost from $S$ to a goal node.
These values are used in two ways: search guidance and pruning.

For search guidance, typically, the priority of a node $S$ is computed from $g(S)$ and $h(S)$, and the node to visit next is selected based on it.
For pruning, a heuristic function needs to be \emph{admissible}: $h(S)$ is a lower bound of the shortest path weight from a node $S$ to a goal node.
In the conventional shortest path problem, if a heuristic function $h$ is admissible, $g(S) + h(S)$ is a lower bound on the weight of a path from the initial node to a goal node via $S$.
Therefore, when we have found a path from the initial node to a goal node with weight $\overline{\gamma}$, we can prune the path to $S$ if $g(S) + h(S) \geq \overline{\gamma}$.
With this pruning, a heuristic search algorithm can be considered a branch-and-bound algorithm \cite{Ibaraki1978,Nau1984}.

While the above two functionalities of a heuristic function are fundamentally different, it is common that a single admissible heuristic function is used for both purposes.
In particular, A* \cite{Hart1968} visits the node that minimizes the $f$-value, $f(S) = g(S) + h(S)$.
While A* does not explicitly prune paths, if the weights of edges are nonnegative, it never expands a state $S$ with path $\sigma(S)$ such that $g(S) + h(S) > \gamma^*$, where $\gamma^*$ is the shortest path weight from the initial node to a goal node.\footnote{While $f^*$ is conventionally used to represent the optimal path weight, we use $\gamma^*$ to explicitly distinguish it from $f$-values.}
Thus, A* implicitly prunes non-optimal paths while guiding the search with the $f$-values.
However, in general, we can use different functions for the two purposes, and the one used for search guidance need not be admissible.
Such multi-heuristic search algorithms have been developed particularly for the bounded-suboptimal setting, where we want to find a solution whose suboptimality is bounded by a constant factor, and the anytime setting, where we want to find increasingly better solutions until proving optimality \cite{Pearl1982,Chakrabarti1989,Baier2009,Thayer2011,Aine2016,Fickert2022}.

In DyPDL, a dual bound function can be used for search guidance, but we may use a different heuristic function.
In this section, we do not introduce heuristic functions for search guidance and do not specify how to select the next node to visit.
Instead, we provide a generic heuristic search algorithm that uses a dual bound function \emph{only} for pruning and discuss its completeness and optimality.
To explicitly distinguish pruning from search guidance, for a dual bound function, we use $\eta$ as in Definition~\ref{def:dual-bound} instead of $h$ and do not use $f$.

We show generic pseudo-code of a heuristic search algorithm for a monoidal DyPDL model in Algorithm~\ref{alg:search}.
The algorithm starts from the target state $S^0$ and searches for a path to a base state by traversing edges in the state transition graph.
The open list $O$ stores candidate states to search.
The set $G$ stores generated states to detect duplicate or dominated states.
If the model satisfies isotonicity, with Theorem~\ref{thm:principle}, we just need to consider the best path to each state in terms of the weight.
The sequence of transitions $\sigma(S)$ represents the best path found so far from the target state $S^0$ to $S$.
The $g$-value of $S$, $g(S)$, is the weight of the path $\sigma(S)$.
The function $\eta$ is a dual bound function, which underestimates the cost of an $S$-solution by the $\eta$-value of $S$, $\eta(S)$.
The best solution found so far, $\overline{\sigma}$, and its cost $\overline{\gamma}$ (i.e., the primal bound) is also maintained.
Algorithm~\ref{alg:search} is for minimization, and $\infty$ is replaced with $-\infty$, $\min$ is replaced with $\max$, $<$ is replaced with $>$, and $\geq$ and $\leq$ are swapped for maximization.
All the theoretical results shown later can be easily adapted to maximization.

\begin{algorithm}[t]
    \caption{
        Heuristic search for minimization with a monoidal DyPDL model $\langle \mathcal{V}, S^0, \mathcal{T}, \mathcal{B}, \mathcal{C} \rangle$ with $\langle A, \times, \mathbf{1} \rangle$.
        An approximate dominance relation $\preceq_a$ and a dual bound function $\eta$ are given as input.
    }
    \begin{algorithmic}[1]
        \If{$S^0 \not\models \mathcal{C}$}
        \Return $\text{NULL}$ \label{alg:search:target-constr}
        \EndIf
        \State $\overline{\gamma} \leftarrow \infty, \overline{\sigma} \leftarrow \text{NULL}$ \label{alg:search:init-solutions} \Comment{Initialize the solution.}
        \State $\sigma(S^0) \leftarrow \langle \rangle$, $g(S^0) \leftarrow \mathbf{1}$ \label{alg:search:init-functions} \Comment{Initialize the $g$-value.}
        \State $G, O \leftarrow \{ S^0 \}$ \label{alg:search:init-lists} \Comment{Initialize the open list.}
        \While{$O \neq \emptyset$} \label{alg:search:while}
        \State Let $S \in O$ \label{alg:search:select} \Comment{Select a state.}
        \State $O \leftarrow O \setminus \{ S \}$ \label{alg:search:remove} \Comment{Remove the state.}
        \If{$\exists B \in \mathcal{B}, S \models C_B$} \label{alg:search:check-base}
        \State $\text{current\_cost} \leftarrow g(S) \times \min_{B \in \mathcal{B} : S \models C_B} \mathsf{base\_cost}_B(S)$ \label{alg:search:solution-cost} \Comment{Compute the solution cost.}
        \If{$\text{current\_cost} < \overline{\gamma}$} \label{alg:search:check-solution}
        \State $\overline{\gamma} \leftarrow \text{current\_cost}$, $\overline{\sigma} \leftarrow \sigma(S)$ \label{alg:search:update-solution} \Comment{Update the best solution.}
        \State $O \leftarrow \{ S' \in O \mid g(S') \times \eta(S') < \overline{\gamma} \}$ \label{alg:search:prune-open} \Comment{Prune states in the open list.}
        \EndIf
        \Else
        \ForAll{$\tau \in \mathcal{T}^*(S) : S[\![\tau]\!] \models \mathcal{C}$} \label{alg:search:gen}
        \State $g_{\text{current}} \leftarrow g(S) \times w_\tau(S)$ \Comment{Compute the $g$-value.} \label{alg:search:compute-g}
        \If{$\not\exists S' \in G$ such that $S[\![\tau]\!] \preceq_a S'$ and $g_{\text{current}} \geq g(S')$} \label{alg:search:dominating}
        \If{$g_{\text{current}} \times \eta(S[\![\tau]\!]) < \overline{\gamma}$} \label{alg:search:check-bound}
        \If{$\exists S' \in G$ such that $S' \preceq_a S[\![\tau]\!]$ and $g_{\text{current}} \leq g(S')$} \label{alg:search:dominated}
        \State $G \leftarrow G \setminus \{ S' \}$, $O \leftarrow O \setminus \{ S' \}$ \label{alg:search:remove-dominated} \Comment{Remove a dominated state.}
        \EndIf
        \State $\sigma(S[\![\tau]\!]) \leftarrow \langle \sigma(S); \tau \rangle$, $g(S[\![\tau]\!]) \leftarrow g_{\text{current}}$ \label{alg:search:update-g}
        \State $G \leftarrow G \cup \{ S[\![\tau]\!] \}$, $O \leftarrow O \cup \{ S[\![\tau]\!] \}$ \label{alg:search:insert} \Comment{Insert the successor state.}
        \EndIf
        \EndIf
        \EndFor
        \EndIf
        \EndWhile
        \State \Return $\overline{\sigma}$ \label{alg:search:return} \Comment{Return the solution.}
    \end{algorithmic}
    \label{alg:search}
\end{algorithm}

If the target state $S^0$ violates the state constraints, the model does not have a solution, so we return $\text{NULL}$ (line~\ref{alg:search:target-constr}).
Otherwise, the open list $O$ and $G$ are initialized with $S^0$ (line~\ref{alg:search:init-lists}).
The $g$-value of $S^0$ is initialized to $\mathbf{1}$ following Definition~\ref{def:path-cost} (line~\ref{alg:search:init-functions}).
Initially, the solution cost $\overline{\gamma} = \infty$, and $\overline{\sigma} = \text{NULL}$ (line~\ref{alg:search:init-solutions}).
When $O$ is empty, $\overline{\sigma}$ is returned (line~\ref{alg:search:return}).
In such a case, the state transition graph is exhausted, and the current solution $\overline{\sigma}$ is an optimal solution, or the model does not have a solution if $\overline{\sigma} = \text{NULL}$.

When $O$ is not empty, a state $S \in O$ is selected and removed from $O$ (lines~\ref{alg:search:select} and \ref{alg:search:remove}).
We do not specify how to select $S$ in Algorithm~\ref{alg:search} as it depends on the concrete heuristic search algorithms implemented.
If $S$ is a base state, $\sigma(S)$ is a solution, so we update the best solution if $\sigma(S)$ is better (lines~\ref{alg:search:check-base}--\ref{alg:search:update-solution}).
If the best solution is updated, we prune a state $S'$ in $O$ such that $g(S') \times \eta(S')$ is not better than the new solution cost since the currently found paths to such states do not lead to a better solution (line~\ref{alg:search:prune-open}).

If $S$ is not a base state, $S$ is \emph{expanded}.
We define a set of applicable transitions considering forced transitions as
\begin{equation}
    \mathcal{T}^*(S) = \begin{cases}
        \{ \tau \}     & \text{if } \exists \tau \in \mathcal{T}(S), \tau \text{ is identified to be a forced transition} \\
        \mathcal{T}(S) & \text{otherwise.}
    \end{cases}
\end{equation}
We expect that identifying all forced transitions is not practical but identifying some of them is feasible, e.g., based on sufficient conditions defined by a user as in the talent scheduling example in Section~\ref{sec:forced}.
In the first line, when multiple forced transitions are identified, we assume that one of them is selected.
For example, we can select the one defined first in the model in practice.
A successor states $S[\![\tau]\!]$ is generated for each transition $\tau \in \mathcal{T}^*(S)$ (line~\ref{alg:search:gen}).
In doing so, successor states violating state constraints are discarded.
For each successor state, we check if a state $S'$ that dominates $S[\![\tau]\!]$ and has a better or equal $g$-value is already generated (line~\ref{alg:search:dominating}).
In such a case, $\sigma(S')$ leads to a better or equal solution, so we prune $S[\![\tau]\!]$.
Since $S[\![\tau]\!]$ itself dominates $S[\![\tau]\!]$, this check also works as duplicate detection.
If a dominating state in $G$ is not detected, and $g_{\text{current}} \times \eta(S[\![\tau]\!])$ is better than the primal bound (line~\ref{alg:search:check-bound}), we insert it into $G$ and $O$ (line~\ref{alg:search:insert}).
The best path to $S[\![\tau]\!]$ is updated to $\langle \sigma(S); \tau \rangle$, which is an extension of $\sigma(S)$ with $\tau$ (line~\ref{alg:search:update-g}).
Before doing so, we remove an existing state $S'$ from $G$ if $S'$ is dominated by $S[\![\tau]\!]$ with a worse or equal $g$-value (line~\ref{alg:search:dominated}).

Algorithm~\ref{alg:search} terminates in finite time if the model is finite and cost-algebraic.
In addition, even if the model is not cost-algebraic, if it is acyclic, it still terminates in finite time.
First, we show the termination for a finite and acyclic model.
Intuitively, with such a model, Algorithm~\ref{alg:search} enumerates a finite number of paths from the target state, so it eventually terminates.

\begin{theorem} \label{thm:search-terminate-acyclic}
    Given a finite, acyclic, and monoidal DyPDL model, Algorithm~\ref{alg:search} terminates in finite time.
\end{theorem}

\begin{proof}
    Unless the target state violates the state constraints, the algorithm terminates when $O$ becomes an empty set.
    In each iteration of the loop in lines~\ref{alg:search:while}--\ref{alg:search:insert}, at least one state is removed from $O$ by line~\ref{alg:search:remove}.
    However, multiple successor states can be added to $O$ in each iteration by line~\ref{alg:search:insert}.
    We prove that the number of iterations that reach line~\ref{alg:search:insert} is finite.
    With this property, $O$ eventually becomes an empty set with finite iterations.

    A successor state $S[\![\tau]\!]$ is inserted into $O$ if it is not dominated by a state in $G$ with a better or equal $g$-value, and $g_{\text{current}} \times \eta(S[\![\tau]\!])$ is less than the current solution cost.
    Suppose that $S[\![\tau]\!]$ was inserted into $O$ and $G$ in line~\ref{alg:search:insert}, and now the algorithm generates $S[\![\tau]\!]$ again in line~\ref{alg:search:gen}.
    Suppose that $g_{\text{current}} = g(S) \times w_\tau(S) \geq g(S[\![\tau]\!])$.
    If $S[\![\tau]\!] \in G$, then we do not add $S[\![\tau]\!]$ to $O$ due to line~\ref{alg:search:dominated}.
    If $S[\![\tau]\!] \not\in G$, then $S[\![\tau]\!]$ was removed from $G$, so we should have generated a state $S'$ such that $S[\![\tau]\!] \preceq_a S'$ and $g(S') \leq g(S[\![\tau]\!])$ (lines~\ref{alg:search:dominating} and \ref{alg:search:remove-dominated}).
    It is possible that $S'$ was also removed from $G$, but in such a case, we have another state $S'' \in G$ such that $S[\![\tau]\!] \preceq_a S' \preceq_a S''$ and $g(S'') \leq g(S') \leq g(S[\![\tau]\!])$, so $S[\![\tau]\!]$ is not inserted into $O$ again.
    Thus, if $S[\![\tau]\!]$ was ever inserted into $G$, then $S[\![\tau]\!]$ is inserted into $O$ in line~\ref{alg:search:insert} only if $g_{\text{current}} < g(S[\![\tau]\!])$.
    We need to find a better path from $S^0$ to $S[\![\tau]\!]$.
    Since the model is finite and acyclic, the number of paths from $S^0$ to each state is finite.
    Therefore, each state is inserted into $O$ finite times.
    Since the model is finite, the number of reachable states is finite.
    By line~\ref{alg:search:gen}, we only generate reachable states.
    Thus, we reach line~\ref{alg:search:insert} finitely many times.
\end{proof}

When the state transition graph contains cycles, there can be an infinite number of paths even if the graph is finite.
However, if the model is cost-algebraic, the cost monotonically changes along a path, so having a cycle does not improve a solution.
Thus, the algorithm terminates in finite time by enumerating a finite number of acyclic paths.
We start with the following lemma, which confirms that the $g$-value is the weight of the path from the target state.

\begin{lemma} \label{lem:sigma}
    After line~\ref{alg:search:init-lists} of Algorithm~\ref{alg:search}, for each state $S \in O$, $S$ is the target state $S^0$, or $S$ is reachable from $S^0$ with $\sigma(S)$ such that $g(S) = w_{\sigma(S)}(S^0)$ at all lines except for \ref{alg:search:update-g}--\ref{alg:search:insert}.
\end{lemma}

\begin{proof}
    Assume that the following condition holds at the beginning of the current iteration:
    for each state $S \in O$, $S$ is the target state $S^0$ with $g(S^0) = \mathbf{1}$, or $S$ is reachable from $S^0$ with $\sigma(S)$ and $g(S) = w_{\sigma(S)}(S^0)$.
    In the first iteration, $O = \{ S^0 \}$, so the assumption holds.
    When the assumption holds, the condition continues to hold until reaching lines~\ref{alg:search:update-g}--\ref{alg:search:insert}, where the $g$-value is updated, and a new state is added to $O$.
    If we reach these lines, a non-base state $S$ was removed from $O$ in line~\ref{alg:search:remove}.
    Each successor state $S[\![\tau]\!]$ is reachable from $S$ with $\langle \tau \rangle$.
    By the assumption, $S = S^0$, or $S$ is reachable from $S^0$ with $\sigma(S)$.
    Therefore, $S[\![\tau]\!]$ is reachable from $S^0$ with $\langle \sigma(S); \tau \rangle$.
    If $S[\![\tau]\!]$ is inserted into $O$, then $\sigma(S[\![\tau]\!]) = \langle \sigma(S); \tau \rangle$.
    If $S = S^0$,
    \begin{equation*}
        g(S[\![\tau]\!]) = g(S^0) \times w_\tau(S^0) = \mathbf{1} \times w_\tau(S^0) = w_\tau(S^0) = w_{\sigma(S[\![\tau]\!])}(S^0).
    \end{equation*}
    If $S$ is not the target state, since $g(S) = w_{\sigma(S)}(S^0)$, by Definition~\ref{def:path-cost},
    \begin{equation*}
        g(S[\![\tau]\!]) = g(S) \times w_\tau(S) = w_{\sigma(S)}(S^0) \times w_\tau(S) = w_{\sigma(S[\![\tau]\!])}(S^0).
    \end{equation*}
    Thus, $S[\![\tau]\!]$ is reachable from $S^0$ with $\sigma(S[\![\tau]\!])$ and $g(S[\![\tau]\!]) = w_{\sigma(S[\![\tau]\!])}(S^0)$, so the condition holds after line~\ref{alg:search:insert}.
    By mathematical induction, the lemma is proved.
\end{proof}

\begin{theorem} \label{thm:search-terminate-cost-algebraic}
    Given a finite and cost-algebraic DyPDL model, Algorithm~\ref{alg:search} terminates in finite time.
\end{theorem}

\begin{proof}
    The proof is almost the same as the proof of Theorem~\ref{thm:search-terminate-acyclic}.
    However, now, there may be an infinite number of paths to a state since the state transition graph may contain cycles.
    We show that the algorithm never considers a path containing cycles when the model is cost-algebraic.
    Assume that for each state $S$, the best found path $\sigma(S)$ is acyclic up to the current iteration.
    This condition holds at the beginning since $\sigma(S^0) = \langle \rangle$ is acyclic.
    Suppose that the algorithm generates a successor state $S[\![\tau]\!]$ that is already included in the path $\sigma(S)$.
    Then, $S[\![\tau]\!]$ was generated before.
    In addition, $S[\![\tau]\!]$ is not a base state since it has a successor state on $\sigma(S)$.
    Since $\sigma(S)$ is acyclic, $S[\![\tau]\!]$ is included only once.
    Let $\sigma(S) = \langle \sigma^1; \sigma^2 \rangle$ where $\sigma^1$ is the path from $S^0$ to $S[\![\tau]\!]$.
    By Lemma~\ref{lem:sigma}, we have
    \begin{equation*}
        g_{\text{current}} = g(S) \times w_\tau(S) = w_{\sigma^1}(S^0) \times w_{\sigma^2}(S[\![\tau]\!]) \times w_\tau(S).
    \end{equation*}
    If $g(S[\![\tau]\!])$ and $\sigma(S[\![\tau]\!])$ were updated after $S[\![\tau]\!]$ was generated with $\sigma(S[\![\tau]\!]) = \sigma^1$, then a path from $S^0$ to $S[\![\tau]\!]$ with a smaller weight was found by line~\ref{alg:search:dominating}.
    Thus, $g(S[\![\tau]\!]) \leq w_{\sigma^1}(S^0) = w_{\sigma^1}(S^0) \times \mathbf{1}$.
    By Definition~\ref{def:cost-algebra}, $\mathbf{1} \leq w_{\sigma^2}(S[\![\tau]\!]) \times w_\tau(S)$.
    Since $A$ is isotone,
    \begin{equation*}
        g_{\text{current}} = w_{\sigma^1}(S^0) \times w_{\sigma^2}(S[\![\tau]\!]) \times w_\tau(S) \geq w_{\sigma^1}(S^0) \times \mathbf{1} \geq g(S[\![\tau]\!]).
    \end{equation*}
    Therefore, $S[\![\tau]\!]$ is not inserted into $O$, and $\sigma(S[\![\tau]\!])$ remains acyclic.
    Thus, by mathematical induction, for each state, the number of insertions into $O$ is at most the number of acyclic paths to that state, which is finite.
\end{proof}

We confirm that $\overline{\sigma}$ is a solution for a model when it is not $\text{NULL}$ even during execution.
In other words, Algorithm~\ref{alg:search} is an anytime algorithm that can return a solution before proving the optimality.

\begin{theorem} \label{thm:search-solution}
    After line~\ref{alg:search:update-solution} of Algorithm~\ref{alg:search}, if $\overline{\sigma} \neq \text{NULL}$, then $\overline{\sigma}$ is a solution for the DyPDL model with $\overline{\gamma} = \mathsf{solution\_cost}(\overline{\sigma})$.
\end{theorem}

\begin{proof}
    The solution $\overline{\sigma}$ is updated in line~\ref{alg:search:update-solution} when a base state $S$ is removed from $O$ in line~\ref{alg:search:remove}.
    If $S = S^0$, then $\overline{\sigma} = \langle \rangle$, which is a solution.
    Since $g(S^0) = \mathbf{1}$, $\overline{\gamma} = \min_{B \in \mathcal{B} : S^0 \models C_B} \mathsf{base\_cost}_B(S^0) = \mathsf{solution\_cost}(\overline{\sigma})$ by Definition~\ref{def:dypdl}.
    If $S$ is not the target state, $\overline{\sigma} = \sigma(S)$, which is a solution since $S$ is reachable from $S^0$ with $\sigma(S)$ by Lemma~\ref{lem:sigma}, and $S$ is a base state.
    Since $g(S) = w_{\sigma(S)}(S^0)$, it holds that $\overline{\gamma} = w_{\sigma(S)}(S^0) \times \min_{B \in \mathcal{B} : S \models C_B} \mathsf{base\_cost}_B(S) = \mathsf{solution\_cost}(\overline{\sigma})$ by Lemma~\ref{lem:path-cost}.
\end{proof}

Finally, we prove the optimality of Algorithm~\ref{alg:search}.
Our proof is based on the following lemma, whose proof is presented in \ref{sec:proofs}.

\begin{lemma} \label{lem:open-not-empty}
    In Algorithm~\ref{alg:search}, suppose that a solution exists for the DyPDL model, and let $\hat{\gamma}$ be its cost.
    When reaching line~\ref{alg:search:while}, at least one of the following two conditions is satisfied:
    \begin{itemize}
        \item $\overline{\gamma} \leq \hat{\gamma}$.
        \item The open list $O$ contains a state $\hat{S}$ such that an $\hat{S}$-solution $\hat{\sigma}$ exists, and $\langle \sigma(\hat{S}); \hat{\sigma} \rangle$ is a solution for the model with $\mathsf{solution\_cost}(\langle \sigma(\hat{S}); \hat{\sigma} \rangle) \leq \hat{\gamma}$.
    \end{itemize}
\end{lemma}

\begin{theorem} \label{thm:search-optimality}
    Let $\langle A, \times, \mathbf{1} \rangle$ be a monoid where $A \subseteq \mathbb{Q} \cup \{ -\infty, \infty \}$ and $A$ is isotone.
    Given a monoidal DyPDL model $\langle \mathcal{V}, S^0, \mathcal{T}, \mathcal{B}, \mathcal{C} \rangle$ with $\langle A, \times, \mathbf{1} \rangle$, if an optimal solution exists for the model, and Algorithm~\ref{alg:search} returns a solution that is not $\text{NULL}$, then the solution is optimal.
    If Algorithm~\ref{alg:search} returns $\text{NULL}$, then the model is infeasible.
\end{theorem}

\begin{proof}
    Suppose that a solution exists, and let $\hat{\gamma}$ be its cost.
    By Lemma~\ref{lem:open-not-empty}, when we reach line~\ref{alg:search:while} with $O = \emptyset$, $\overline{\gamma} \leq \hat{\gamma}$.
    Since $\overline{\gamma} \neq \infty$, it holds that $\overline{\sigma} \neq \text{NULL}$ by line~\ref{alg:search:update-solution}.
    Therefore, if a solution exists, $\text{NULL}$ is never returned, i.e., $\text{NULL}$ is returned only if the model is infeasible.
    Suppose that an optimal solution exists, and let $\gamma^*$ be its cost.
    Now, consider the above discussion with $\hat{\gamma} = \gamma^*$.
    When we reach line~\ref{alg:search:while} with $O = \emptyset$, $\overline{\gamma} \leq \gamma^*$.
    By Theorem~\ref{thm:search-solution}, $\overline{\sigma}$ is a solution with $\mathsf{solution\_cost}(\overline{\sigma}) = \overline{\gamma}$.
    Since $\mathsf{solution\_cost}(\overline{\sigma}) \geq \gamma^*$,  $\overline{\gamma} = \gamma^*$ and $\overline{\sigma}$ is an optimal solution.
    Therefore, if an optimal solution exists, and the algorithm returns a solution, the solution is optimal.
\end{proof}

\begin{corollary}
    Given a finite and cost-algebraic DyPDL model, the model has an optimal solution, or the model is infeasible.
    A problem to decide if a solution whose cost is less (greater) than a given rational number exists for minimization (maximization) is decidable.
\end{corollary}

Note that Theorem~\ref{thm:search-optimality} does not require a model to be finite, acyclic, or cost-algebraic.
While the algorithm terminates in finite time if the model is finite and acyclic or cost-algebraic, there is no guarantee in general due to the undecidability in Theorem~\ref{thm:undecidable}.
However, even for such a model, if the algorithm terminates, the optimality or infeasibility is proved.

As shown in the proof of Lemma~\ref{lem:open-not-empty}, when an optimal solution exists, a state $\hat{S}$ such that there exists an optimal solution extending $\sigma(\hat{S})$ is included in the open list.
For minimization (maximization), by taking the minimum (maximum) $g(S) \times \eta(S)$ value in the open list, we can obtain a dual bound, i.e., a lower (upper) bound on the optimal solution cost.

\begin{theorem} \label{thm:search-dual-bound}
    Let $\langle A, \times, \mathbf{1} \rangle$ be a monoid where $A \subseteq \mathbb{Q} \cup \{ -\infty, \infty \}$ and $A$ is isotone.
    Given a monoidal DyPDL model $\langle \mathcal{V}, S^0, \mathcal{T}, \mathcal{B}, \mathcal{C} \rangle$ with $\langle A, \times, \mathbf{1} \rangle$, if an optimal solution exists for the model and has the cost $\gamma^*$, and $O$ is not empty in line~\ref{alg:search:while}, for minimization,
    \begin{equation*}
        \min_{S \in O} g(S) \times \eta(S) \leq \gamma^*.
    \end{equation*}
\end{theorem}

\begin{proof}
    By Lemma~\ref{lem:open-not-empty}, if $O \neq \emptyset$, then $\overline{\gamma} = \gamma^*$, or there exists a state $\hat{S} \in O$ on an optimal solution, i.e., there exists an $\hat{S}$-solution $\hat{\sigma}$ such that $\langle \sigma(\hat{S}); \hat{\sigma} \rangle$ is an optimal solution.
    If $\overline{\gamma} = \gamma^*$, by lines~\ref{alg:search:prune-open} and \ref{alg:search:check-bound}, $\min_{S \in O} g(S) \times \eta(S) < \gamma^*$.
    Otherwise, $\hat{S} \in O$.
    Since $\eta(\hat{S}) \leq \mathsf{solution\_cost}(\hat{\sigma}, \hat{S})$ and $A$ is isotone,
    \begin{equation*}
        \min_{S \in O} g(S) \times \eta(S) \leq g(\hat{S}) \times \eta(\hat{S}) \leq g(\hat{S}) \times \mathsf{solution\_cost}(\hat{\sigma}, \hat{S}) = \gamma^*.
    \end{equation*}
\end{proof}

\subsection{Heuristic Search Algorithms for DyPDL} \label{sec:algorithms}

We introduce existing heuristic search algorithms as instantiations of Algorithm~\ref{alg:search} so that we can use them for DyPDL.
In particular, each algorithm differs in how to select a state $S$ to remove from the open list $O$ in line~\ref{alg:search:select}.
In addition to A*, which is the most fundamental heuristic search algorithm, we select anytime algorithms that have been applied to combinatorial optimization problems in problem-specific settings.
For detailed descriptions of the algorithms, please refer to the papers that proposed them.
Similar to A*, in our configuration, these algorithms use a heuristic function $h$ and guide the search with the $f$-value, which is computed as $f(S) = g(S) \times h(S)$, where $\times$ is a binary operator such as $+$.
As we discussed in Section~\ref{sec:dypdl-heuristic-search}, $h$ is not necessarily a dual bound function and not necessarily admissible.

\subsubsection{CAASDy: Cost-Algebraic A* Solver for DyPDL} \label{sec:caasdy}

A* selects a state with the best $f$-value in line~\ref{alg:search:remove} (i.e., the minimum $f$-value for minimization and the maximum $f$-value for maximization).
If there are multiple states with the best $f$-value, one is selected according to a tie-breaking strategy.
Among states with the same $f$-value, we select a state with the best $h$-value (with ``best'' defined accordingly to the best $f$-value).
In what follows, if we select a state according to the $f$-values in other algorithms, we also assume that a state with the best $f$-value is selected, and ties are broken by the $h$-values.
If there are multiple states with the best $f$- and $h$-values, we use another tie-breaking strategy, which is not specified here and discussed later when we describe the implementation.
We call our solver cost-algebraic A* solver for DyPDL (CAASDy) as we originally proposed it only for cost-algebraic models \cite{Kuroiwa2023CAASDy}.
However, as shown in Theorems~\ref{thm:search-solution} and \ref{thm:search-optimality}, CAASDy is applicable to monoidal and acyclic models with a monoid $\langle A, \times, \mathbf{1} \rangle$ if $A$ is isotone.

In original A*, if $h$ is admissible, the first solution found is optimal.
Previous work has generalized A* to cost-algebraic heuristic search with this property \cite{Edelkamp2005}.
In our case, if a model is not cost-algebraic, the first solution may not be an optimal solution.
In addition, even if a model is cost-algebraic, our problem setting is slightly different from \citeauthor{Edelkamp2005}~\cite{Edelkamp2005}:
a base case has a cost, so the cost of a solution can be different from the weight of the corresponding path.
If a model is cost-algebraic and the costs of the base cases do not matter, we can prove that the first solution found by CAASDy is optimal.

\begin{theorem} \label{thm:a*}
    Given a cost-algebraic DyPDL model with a monoid $\langle A, \times, \mathbf{1} \rangle$, let $h$ be an admissible heuristic function, i.e., given any reachable state $S$ and any $S$-solution $\sigma$, $h(S) \leq \mathsf{solution\_cost}(\sigma, S)$ for minimization.
    If an optimal solution exists for the model, the costs of base cases are $\mathbf{1}$, i.e., $\forall B \in \mathcal{B}, \mathsf{base\_cost}_B(S) = \mathbf{1}$, and $h(S) \geq \mathbf{1}$ for any reachable state $S$, then the first found solution by CAASDy is optimal.
\end{theorem}

\begin{proof}
    Let $\overline{\sigma} = \sigma(S)$ be the first found solution with the cost $\overline{\gamma}$ in line~\ref{alg:search:update-solution}.
    Since $\min\limits_{B \in \mathcal{B} : S \models C_B} \mathsf{base\_cost}_B(S) = \mathbf{1}$, $\overline{\gamma} = g(S)$.
    Since $\mathbf{1} \leq h(S) \leq \min\limits_{B \in \mathcal{B} : S \models C_B} \mathsf{base\_cost}_B(S) = \mathbf{1}$, $f(S) = g(S) \times h(S) = g(S) = \overline{\gamma}$.
    If $\sigma(S)$ is not an optimal solution, by Lemma~\ref{lem:open-not-empty}, $O$ contains a state $\hat{S} \in O$ such that there exists an $\hat{S}$-solution $\hat{\sigma}$ with $\mathsf{solution\_cost}(\langle \sigma(\hat{S}); \hat{\sigma} \rangle) = \gamma^* < \overline{\gamma}$.
    Since $A$ is isotone, $\mathsf{solution\_cost}(\langle \sigma(\hat{S}); \hat{\sigma} \rangle) = g(\hat{S}) \times \mathsf{solution\_cost}(\hat{\sigma}, \hat{S}) \geq g(\hat{S}) \times h(\hat{S})$.
    Therefore,
    \begin{equation*}
        f(\hat{S}) = g(\hat{S}) \times h(\hat{S}) \leq \mathsf{solution\_cost}(\langle \sigma(\hat{S}); \hat{\sigma} \rangle) < \overline{\gamma} = f(S).
    \end{equation*}
    Thus, $\hat{S}$ should have been expanded before $S$, which is a contradiction.
\end{proof}

\subsubsection{Depth-First Branch-and-Bound (DFBnB)}

Theorem~\ref{thm:a*} indicates a potential disadvantage of A*:
it does not find any feasible solutions before proving the optimality, which may take a long time.
In practice, having a subopitmal solution is better than having no solution.
In some combinatorial optimization problems, we can find a solution by applying a fixed number of transitions.
For example, in talent scheduling, any sequence of scenes is a solution.
With this observation, prioritizing depth in the state transition graph possibly leads to quickly finding a solution.

Depth-first branch-and-bound (DFBnB) expands states in the depth-first order, i.e., a state $S$ that maximizes $|\sigma(S)|$ is selected in line~\ref{alg:search:select}.
Concretely, the open list $O$ is implemented as a stack, and the state on the top of the stack (the state added to $O$ most recently) is selected in line~\ref{alg:search:select}.
Successor states of the same state have the same priority, so ties are broken by the $f$-values.
DFBnB has been applied to state-based formulations of combinatorial optimization problems such as the sequential ordering problem (SOP) \cite{Libralesso2020}, the traveling salesperson problem (TSP), and single machine scheduling \cite{Vadlamudi2012,Vadlamudi2016}.

\subsubsection{Cyclic Best-First Search (CBFS)}

In DFBnB, in terms of search guidance, $f$-values are used to break ties between successor states of the same state.
Cyclic best-first search (CBFS) \cite{Kao2009} is similar to DFBnB, but it relies more on $f$-values to guide the search and can be viewed as a hybridization of A* and DFBnB.
CBFS partitions the open list $O$ into layers $O_i$ for each depth $i$.
A state $S$ is inserted into $O_i$ if $\sigma(S)$ has $i$ transitions.
At the beginning, $O_0 = \{ S^0 \}$ and $O_i = \emptyset$ for $i > 0$.
Starting with $i=0$, if $O_i \neq \emptyset$, CBFS selects a state having the best priority from $O_i$ in line~\ref{alg:search:select} and inserts successor states into $O_{i+1}$ in line~\ref{alg:search:insert}.
We use the $f$-value as the priority.
After that, CBFS increases $i$ by $1$.
However, when $i$ is the maximum depth, CBFS resets $i$ to $0$ instead of incrementing it.
The maximum depth is usually known in a problem-specific setting, but we do not use a fixed parameter in our setting.
Instead, we set $i$ to $0$ when a new best solution is found after line~\ref{alg:search:update-solution}, or $O_j = \emptyset$ for all $j \geq i$.
In problem specific settings, CBFS was used in single machine scheduling \cite{Kao2009} and the simple assembly line balancing problem (SALBP-1) \cite{Sewell2012,Morrison2014}.

\subsubsection{Anytime Column Search (ACS)}

Anytime column search (ACS) \cite{Vadlamudi2012} can be considered a generalized version of CBFS, expanding $b$ states at each depth.
ACS also partitions the open list $O$ into $O_i$ for each depth $i$ and selects a state from $O_i$ in line~\ref{alg:search:select}, starting with $i = 0$ and $O_0 = \{ S_0 \}$.
ACS increases $i$ by $1$ after removing $b$ states from $O_i$ or when $O_i$ becomes empty, where $b$ is a parameter.
We remove the best $b$ states according to the $f$-values.

Anytime column progressive search (ACPS) \cite{Vadlamudi2012} is a non-parametric version of ACS, which starts from $b=1$ and increases $b$ by $1$ when it reaches the maximum depth.
Similarly to CBFS, we set $i$ to $0$ and increase $b$ by $1$ when a new best solution is found or $\forall j \geq i, O_j = \emptyset$.
For combinatorial optimization, ACS and ACPS were evaluated on TSP \cite{Vadlamudi2012}.

\subsubsection{Anytime Pack Search (APS)}

Anytime pack search (APS) \cite{Vadlamudi2016} hybridizes A* and DFBnB in a different way from CBFS and ACS.
It maintains the set of the best states $O_b \subseteq O$, initialized with $\{ S^0 \}$, the set of the best successor states $O_c \subseteq O$, and a suspend list $O_s \subseteq O$.
APS expands all states from $O_b$ in line~\ref{alg:search:select} and inserts the best $b$ successor states according to a priority into $O_c$ and other successor states into $O_s$.
When there are fewer than $b$ successor states, all of them are inserted into $O_c$.
After expanding all states in $O_b$, APS swaps $O_b$ and $O_c$ and continues the procedure.
If $O_b$ and $O_c$ are empty, the best $b$ states are moved from $O_s$ to $O_b$.
We use the $f$-value as the priority to select states.

Anytime pack progressive search (APPS) \cite{Vadlamudi2016} starts from $b=1$ and increases $b$ by $\delta$ if $b < \overline{b}$ when the best $b$ states are moved from $O_s$ to $O_b$, where $\delta$ and $\overline{b}$ are parameters.
We use $\delta = 1$ and $\overline{b} = \infty$ following the configuration in TSP and single machine scheduling \cite{Vadlamudi2016}.

\subsubsection{Discrepancy-Based Search}

We consider a search strategy originating from the CP community, discrepancy-based search \cite{Harvey1995}, which assumes that successor states of the same state are assigned priorities based on some estimation.
In our case, we use the $f$-value as the priority.
The idea of discrepancy-based search is that the estimation may make mistakes, but the number of mistakes made by a strong guidance heuristic is limited.
For each path, the discrepancy, the number of deviations from the estimated best path, is maintained.
In our case, the target state has a discrepancy of $0$.
When a state $S$ has a discrepancy of $d$, the successor states with the best priority has the discrepancy of $d$.
Other successor states have the discrepancy of $d+1$.
Discrepancy-based search algorithms search states whose discrepancy is smaller than an upper bound and iteratively increase the upper bound.

Discrepancy-bounded depth-first search (DBDFS) \cite{Beck2000} performs depth-first search that only expands states having the discrepancy between $(i-1)k$ and $ik-1$ inclusive, where $i$ starts from $1$ and increases by $1$ when all states within the range are expanded, and $k$ is a parameter.
The open list is partitioned into two sets $O_0$ and $O_1$, and $O_0 = \{ S^0 \}$ and $O_1 = \emptyset$ at the beginning.
A state is selected from $O_0$ in line~\ref{alg:search:select}.
Successor states with the discrepancy between $(i-1)k$ and $ik-1$ are added to $O_0$, and other states are added to $O_1$.
When $O_0$ becomes empty, it is swapped with $O_1$, and $i$ is increased by $1$.
The discrepancy of states in $O_1$ is $ik$ because the discrepancy is increased by at most $1$ at a successor state.
Therefore, after swapping $O_0$ with $O_1$, the discrepancy of states in $O_0$ falls between the new bounds, $ik$ and $(i+1)k - 1$.
For depth-first search, when selecting a state to remove from $O_0$, we break ties by the $f$-values.
We use $k=1$ in our configuration.
Discrepancy-based search was originally proposed as tree search \cite{Harvey1995,Beck2000} and later applied to state space search for SOP \cite{Libralesso2020}.

\subsubsection{Complete Anytime Beam Search (CABS)} \label{sec:cabs}

While the above algorithms except for A* are similar to depth-first search, we consider beam search, a heuristic search algorithm that searches the state transition graph layer by layer, similar to breadth-first search.
Previous work in heuristic search has shown that breadth-first search can be beneficial in saving memory by discarding states in previous layers \cite{Zhou2006}.
However, breadth-first search may take a long time to find a solution, particularly in our setting.
In DP models for combinatorial optimization problems such as TSPTW and talent scheduling, all solutions have the same length, and an algorithm needs to reach the last layer to find a solution.
For such settings, breadth-first search may need to expand many states before reaching the last layer.
Beam search mitigates this issue by expanding at most $b$ states at each layer, where $b$ is a parameter called a beam width, while losing completeness.

Since beam search cannot be considered an instantiation of Algorithm~\ref{alg:search}, we provide dedicated pseudo-code in Algorithm~\ref{alg:beam}.
Beam search maintains states in the same layer, i.e., states that are reached with the same number of transitions, in the open list $O$, which is initialized with the target state.
Beam search expands all states in $O$, inserts the best $b$ successor states into $O$, and discards the remaining successor states, where $b$ is a parameter called a beam width.
Beam search may discard all successor states leading to solutions, so it is incomplete, i.e., it may not find a solution.

\begin{algorithm}[t]
    \caption{
        Beam search for minimization with a monoidal DyPDL model $\langle \mathcal{V}, S^0, \mathcal{T}, \mathcal{B}, \mathcal{C} \rangle$ with $\langle A, \times, \mathbf{1} \rangle$.
        An approximate dominance relation $\preceq_a$, a dual bound function $\eta$, a primal bound $\overline{\gamma}$, and a beam width $b$ are given as input.
    }
    \begin{algorithmic}[1]
        \If{$S^0 \not\models \mathcal{C}$}
        \Return $\text{NULL}$, $\top$ \label{alg:beam:target-constr}
        \EndIf
        \State $\overline{\sigma} \leftarrow \text{NULL}$, $\text{complete} \leftarrow \top$ \label{alg:beam:init-solutions} \Comment{Initialize the solution.}
        \State $l \leftarrow 0$, $\sigma^l(S^0) \leftarrow \langle \rangle$, $g^l(S^0) \leftarrow \mathbf{1}$ \label{alg:beam:init-functions} \Comment{Initialize the $g$-value.}
        \State $O \leftarrow \{ S^0 \}$ \label{alg:beam:init-lists} \Comment{Initialize the open list.}
        \While{$O \neq \emptyset$ and $\overline{\sigma} = \text{NULL}$} \label{alg:beam:while}
        \State $G \leftarrow \emptyset$ \label{alg:beam:reset-g} \Comment{Initialize the set of states.}
        \ForAll{$S \in O$} \label{alg:beam:for}
        \If{$\exists B \in \mathcal{B}, S \models C_B$} \label{alg:beam:check-base}
        \State $\text{current\_cost} \leftarrow g^l(S) \times \min_{B \in \mathcal{B} : S \models C_B} \mathsf{base\_cost}_B(S)$ \label{alg:beam:solution-cost} \Comment{Compute the solution cost.}
        \If{$\text{current\_cost} < \overline{\gamma}$} \label{alg:beam:check-solution}
        \State $\overline{\gamma} \leftarrow \text{current\_cost}$, $\overline{\sigma} \leftarrow \sigma^l(S)$ \label{alg:beam:update-solution} \Comment{Update the best solution.}
        \EndIf
        \Else
        \ForAll{$\tau \in \mathcal{T}^*(S) : S[\![\tau]\!] \models \mathcal{C}$} \label{alg:beam:gen}
        \State $g_{\text{current}} \leftarrow g^l(S) \times w_\tau(S)$ \Comment{Compute the $g$-value.}
        \If{$\not\exists S' \in G$ such that $S[\![\tau]\!] \preceq_a S'$ and $g_{\text{current}} \geq g^{l+1}(S')$} \label{alg:beam:dominating}
        \If{$g_{\text{current}} \times \eta(S[\![\tau]\!]) < \overline{\gamma}$} \label{alg:beam:check-bound}
        \If{$\exists S' \in G$ such that $S' \preceq_a S[\![\tau]\!]$ and $g_{\text{current}} \leq g^{l+1}(S')$} \label{alg:beam:dominated}
        \State $G \leftarrow G \setminus \{ S' \}$ \label{alg:beam:remove-dominated} \Comment{Remove a dominated state.}
        \EndIf
        \State $\sigma^{l+1}(S[\![\tau]\!]) \leftarrow \langle \sigma^l(S); \tau \rangle$, $g^{l+1}(S[\![\tau]\!]) \leftarrow g_{\text{current}}$ \label{alg:beam:update-g} \Comment{Update the $g$-value.}
        \State $G \leftarrow G \cup \{ S[\![\tau]\!] \}$ \label{alg:beam:insert} \Comment{Insert the successor state.}
        \EndIf
        \EndIf
        \EndFor
        \EndIf
        \EndFor
        \State $l \leftarrow l + 1$ \Comment{Proceed to the next layer.} \label{alg:beam:increment-l}
        \State $O \leftarrow \{ S \in G \mid g^l(S) \times \eta(S) < \overline{\gamma} \}$ \label{alg:beam:prune-open} \Comment{Prune states by the bound.}
        \If{$|O| > b$} \label{alg:beam:check-size}
        \State $O \leftarrow$ the best $b$ states in $O$, $\text{complete} \leftarrow \bot$ \label{alg:beam:prune-width} \Comment{Keep the best $b$ states.}
        \EndIf
        \EndWhile
        \If{$\text{complete}$ and $O \neq \emptyset$} \label{alg:beam:check-empty}
        \State $\text{complete} \leftarrow \bot$ \label{alg:beam:update-complete} \Comment{A better solution may exist.}
        \EndIf
        \State \Return $\overline{\sigma}$, $\text{complete}$ \label{alg:beam:return} \Comment{Return the solution.}
    \end{algorithmic}
    \label{alg:beam}
\end{algorithm}

\begin{algorithm}[t]
    \caption{
        CABS for minimization with a monoidal DyPDL model $\langle \mathcal{V}, S^0, \mathcal{T}, \mathcal{B}, \mathcal{C} \rangle$ with $\langle A, \times, \mathbf{1} \rangle$.
        An approximate dominance relation $\preceq_a$ and a dual bound function $\eta$ are given as input.
    }
    \begin{algorithmic}[1]
        \State $\overline{\gamma} \leftarrow \infty$, $\overline{\sigma} \leftarrow \text{NULL}$, $b \leftarrow 1$ \label{alg:cabs:init} \Comment{Initialization.}
        \Loop
        \State $\sigma, \text{complete} \leftarrow$ \textsc{BeamSearch}{($\langle \mathcal{V}, S^0, \mathcal{T}, \mathcal{B}, \mathcal{C} \rangle$, $\preceq_a$, $\eta$, $\overline{\gamma}$, $b$)} \label{alg:cabs:call} \Comment{Execute Algorithm~\ref{alg:beam}.}
        \If{$\mathsf{solution\_cost}(\sigma)< \overline{\gamma}$} \label{alg:cabs:check-solution}
        \State $\overline{\gamma} \leftarrow \mathsf{solution\_cost}(\sigma)$, $\overline{\sigma} \leftarrow \sigma$ \label{alg:cabs:update-solution} \Comment{Update the solution.}
        \EndIf
        \If{$\text{complete}$} \label{alg:cabs:check-completeness}
        \State \Return $\overline{\sigma}$ \label{alg:cabs:return} \Comment{Return the solution.}
        \EndIf
        \State $b \leftarrow 2b$ \label{alg:cabs:double-beam} \Comment{Double the beam width.}
        \EndLoop
    \end{algorithmic}
    \label{alg:cabs}
\end{algorithm}

Complete anytime beam search (CABS) \cite{Zhang1998} is a complete version of beam search.
In our configuration, CABS iteratively executes beam search with a beam width $b$ starting from $b=1$ and doubles $b$ after each iteration until finding an optimal solution or proving infeasibility.
With this strategy, CABS tries to quickly find a solution with small $b$ and eventually converges to breadth-first search when $b$ is large enough.
Note that this configuration follows \citeauthor{Libralesso2020}~\cite{Libralesso2020,Libralesso2022}, who applied CABS to SOP and permutation flowshop.
Originally, \citeauthor{Zhang1998}~\cite{Zhang1998} considered a generalized version of beam search, i.e., successor states inserted into $O$ are decided by a user-provided pruning rule, which can be different from selecting the best $b$ states.
In this generalized version, CABS repeats beam search while relaxing the pruning rule and terminates when it finds a satisfactory solution according to some criterion.

In Algorithm~\ref{alg:beam}, beam search maintains the set of states in the current layer using the open list $O$, and the set of states in the next layer using $G$.
The open list $O$ is updated to $G$ after generating all successor states while pruning states based on the bound (line~\ref{alg:beam:prune-open}).
If $O$ contains more than $b$ states, only the best $b$ states are kept.
This operation may prune optimal solutions, so the flag $\text{complete}$, which indicates the completeness of beam search, becomes $\bot$.
Beam search terminates when $O$ becomes empty, or a solution is found (line~\ref{alg:beam:while}).
Even if $\text{complete} = \top$, and a solution is found, when $O$ is not empty, we may miss a better solution (line~\ref{alg:beam:check-empty}).
Therefore, we update $\text{complete}$ to $\bot$ in such a case (line~\ref{alg:beam:update-complete}).
We can derive the maximization version of Algorithm~\ref{alg:beam} in a similar way as Algorithm~\ref{alg:search}.

Beam search in Algorithm~\ref{alg:beam} has several properties that are different from Algorithm~\ref{alg:search}.
\begin{enumerate}
    \item The set $G$, which is used to detect dominance, contains only states in the next layer.
    \item A state $S$ may be dropped from the open list $O$ even if $g(S) \times \eta(S) < \overline{\gamma}$ (line~\ref{alg:beam:prune-width}).
    \item Beam search may terminate when a solution is found even if $O \neq \emptyset$.
\end{enumerate}
With Property (1), beam search can potentially save memory compared to Algorithm~\ref{alg:search}.
This method can be considered layered duplicate detection as proposed in previous work \cite{Zhou2006}.
With this strategy, we do not detect duplicates when the same states appear in different layers.
When a generated successor state $S[\![\tau]\!]$ in the next layer is the same as a state $S'$ in the current layer, in line~\ref{alg:beam:update-g}, we do not want to update $g(S')$ and $\sigma(S')$ since we do not check if a better path to $S'$ is found.
Thus, we maintain $l$, which is incremented by 1 after each layer (line~\ref{alg:beam:increment-l}), and use $g^l$ and $\sigma^l$ to differentiate $g$ and $\sigma$ for different layers.

Our layered duplicate detection mechanism prevents us from using beam search when the state transition graph contains cycles;
beam search cannot store states found in the previous layers, so it continues to expand states in a cycle.
This issue can be addressed by initializing $G$ with $\{ S^0 \}$ outside the while loop, e.g., just after line~\ref{alg:beam:init-lists}, and removing line~\ref{alg:beam:reset-g}.
With this modification, beam search can be used for a cyclic but cost-algebraic DyPDL model.

By Properties (2) and (3), there is no guarantee that beam search proves the optimality or infeasibility unless $\text{complete} = \top$.
However, CABS (Algorithm~\ref{alg:cabs}) has the guarantee of optimality as it repeats beam search until $\text{complete}$ becomes $\top$.
In what follows, we formalize the above points.
Once again, we present the theoretical results for minimization, but they can be easily adapted to maximization.

\begin{theorem} \label{thm:beam-terminate-acyclic}
    Given a finite, acyclic, and monoidal DyPDL model, beam search terminates in finite time.
\end{theorem}

\begin{proof}
    Suppose that we have generated a successor state $S[\![\tau]\!]$, which was generated before.
    The difference from Algorithm~\ref{alg:search} is that we need to consider is Property (1).
    If $S[\![\tau]\!]$ was generated before as a successor state of a state in the current layer, by the proof of Theorem~\ref{thm:search-terminate-acyclic}, there exists a state $S' \in G$ with $S[\![\tau]\!] \preceq_a S'$.
    The successor state $S[\![\tau]\!]$ is inserted into $G$ again only if we find a better path to $S[\![\tau]\!]$.
    If $S[\![\tau]\!]$ was generated before as a successor state of a state in a previous layer, the path to $S[\![\tau]\!]$ at that time was shorter (in terms of the number of transitions) than that of the current path.
    Thus, the current path is different from the previous path.
    The successor state $S[\![\tau]\!]$ may be inserted into $G$ since it is not included in $G$.
    In either case, $S[\![\tau]\!]$ is inserted into $G$ again only if we find a new path to it.
    Since the number of paths to $S[\![\tau]\!]$ is finite, we insert $S[\![\tau]\!]$ into $G$ finite times.
    The rest of the proof follows that of Theorem~\ref{thm:search-terminate-acyclic}.
\end{proof}

As we discussed above, with a slight modification, we can remove Property (1) and prove the termination of beam search for a cost-algebraic DyPDL model.

Since Properties (1)--(3) do not affect the proofs of Lemma~\ref{lem:sigma} and Theorem~\ref{thm:search-solution}, the following theorem holds.

\begin{theorem} \label{thm:beam-solution}
    After line~\ref{alg:beam:update-solution} of Algorithm~\ref{alg:beam}, if $\overline{\sigma} \neq \text{NULL}$, then $\overline{\sigma}$ is a solution for the model with $\overline{\gamma} = \mathsf{solution\_cost}(\overline{\sigma})$.
\end{theorem}

We also prove the optimality of beam search when $\text{complete} = \top$.

\begin{theorem} \label{thm:beam-optimality}
    Let $\langle A, \times, \mathbf{1} \rangle$ be a monoid where $A \subseteq \mathbb{Q} \cup \{ -\infty, \infty \}$ and $A$ is isotone.
    Given a monoidal DyPDL model $\langle \mathcal{V}, S^0, \mathcal{T}, \mathcal{B}, \mathcal{C} \rangle$ with $\langle A, \times, \mathbf{1} \rangle$ and $\overline{\gamma} \in A$, if an optimal solution exists for the model, and beam search returns $\overline{\sigma} \neq \text{NULL}$ and $\text{complete} = \top$, then $\overline{\sigma}$ is an optimal solution.
    If beam search returns $\overline{\sigma} = \text{NULL}$ and $\text{complete} = \top$, then there does not exist a solution whose cost is less than $\overline{\gamma}$.
\end{theorem}

\begin{proof}
    When $\text{complete} = \top$ is returned, during the execution, beam search never reached lines~\ref{alg:beam:prune-width} and \ref{alg:beam:update-complete}.
    Therefore, we can ignore Properties (2) and (3).
    If we modify Algorithm~\ref{alg:beam} so that $G$ contains states in all layers as discussed above, we can also ignore Property (1).
    By ignoring Properties (1)--(3), we can consider beam search as an instantiation of Algorithm~\ref{alg:search}.
    If the model is infeasible, or an optimal solution exists with the cost $\gamma^*$ and $\overline{\gamma} > \gamma^*$ at the beginning, the proof is exactly the same as that of Theorem~\ref{thm:search-optimality}.
    If $\overline{\gamma} \leq \gamma^*$ was given as input, beam search has never updated $\overline{\sigma}$ and $\overline{\gamma}$, and $\text{NULL}$ is returned if it terminates.
    In such a case, indeed, no solution has a cost less than $\overline{\gamma} \leq \gamma^*$.

    The above proof is for beam search with the modification.
    We confirm that it is also valid with beam search in Algorithm~\ref{alg:beam} without modification, i.e., we consider Property (1).
    The proof of Theorem~\ref{thm:search-optimality} depends on Lemma~\ref{lem:open-not-empty}, which claims that when a solution with a cost $\hat{\gamma}$ exists and $\overline{\gamma} > \hat{\gamma}$, the open list contains a state $\hat{S}$ such that there exists an $\hat{S}$-solution $\hat{\sigma} = \langle \hat{\sigma}_1, ..., \hat{\sigma}_m \rangle$ with $\mathsf{solution\_cost}(\langle \sigma(\hat{S}); \hat{\sigma} \rangle) \leq \hat{\gamma}$.
    At the beginning, $\hat{S} = S^0$ exists in $O$.
    When such a state $\hat{S}$ exists in the current layer, it is expanded.
    First, we show that a successor state of $\hat{S}$ satisfying the condition is generated.
    If no applicable forced transitions are identified in $\hat{S}$, a successor state $\hat{S}[\![\hat{\sigma}_1]\!]$ with $\sigma(\hat{S}[\![\hat{\sigma}_1]\!]) = \langle \sigma(\hat{S}); \hat{\sigma}_1 \rangle$ is generated, and
    \begin{equation*}
        \mathsf{solution\_cost}(\langle \sigma(\hat{S}[\![\hat{\sigma}_1]\!]); \hat{\sigma}_2, ..., \hat{\sigma}_m \rangle) = \mathsf{solution\_cost}(\langle \sigma(\hat{S}); \hat{\sigma} \rangle) \leq \hat{\gamma}.
    \end{equation*}
    If an applicable forced transitions is identified, only one successor $\hat{S}[\![\tau]\!]$ is generated with a forced transition $\tau$.
    By Definition~\ref{def:forced}, there exists an $\hat{S}$-solution $\langle \tau; \hat{\sigma}' \rangle$ with $\mathsf{solution\_cost}(\langle \tau, \hat{\sigma}' \rangle, \hat{S}) \leq \mathsf{solution\_cost}(\hat{\sigma}, \hat{S})$.
    Since $A$ is isotone,
    \begin{equation*}
        \begin{split}
            \mathsf{solution\_cost}(\langle \sigma(\hat{S}[\![\tau]\!]); \hat{\sigma}' \rangle) & = \mathsf{solution\_cost}( \langle \sigma(\hat{S}); \tau; \hat{\sigma}' \rangle) = w_{\sigma(\hat{S})}(S^0) \times \mathsf{solution\_cost}(\langle \tau, \hat{\sigma}' \rangle, \hat{S}) \\
                                                                                                & \leq w_{\sigma(\hat{S})}(S^0) \times \mathsf{solution\_cost}(\hat{\sigma}, \hat{S}) = \mathsf{solution\_cost}(\langle \sigma(\hat{S}); \hat{\sigma} \rangle) \leq \hat{\gamma}.
        \end{split}
    \end{equation*}

    If the successor state $\hat{S}[\![\tau]\!]$ (or $\hat{S}[\![\hat{\sigma}_1]\!]$) is not inserted into $G$ in line~\ref{alg:beam:insert}, another state $S' \in G$ dominates $\hat{S}[\![\tau]\!]$ with a better or equal $g$-value, so there exists a solution extending $\sigma(S')$ with the cost at most $\hat{\gamma}$.
    Thus, $S'$ can be considered a new $\hat{S}$.
    When $\hat{S}[\![\tau]\!]$ or $S'$ is removed from $G$ by line~\ref{alg:beam:remove-dominated}, another state $S''$ that dominates $\hat{S}[\![\tau]\!]$ or $S'$ with a better or equal $g$-value is inserted into $G$, and there exists a solution extending $\sigma(S'')$ with the cost at most $\hat{\gamma}$.
\end{proof}

For CABS, since beam search returns an optimal solution or proves the infeasibility when $\text{complete} = \top$ by Theorem~\ref{thm:beam-optimality}, the optimality is straightforward by line~\ref{alg:cabs:check-completeness} of Algorithm~\ref{alg:cabs}.

\begin{corollary} \label{cor:cabs-optimality}
    Let $\langle A, \times, \mathbf{1} \rangle$ be a monoid where $A \subseteq \mathbb{Q} \cup \{ -\infty, \infty \}$ and $A$ is isotone.
    Given a monoidal DyPDL model $\langle \mathcal{V}, S^0, \mathcal{T}, \mathcal{B}, \mathcal{C} \rangle$ with $\langle A, \times, \mathbf{1} \rangle$, if an optimal solution exists for the model, and CABS returns a solution that is not $\text{NULL}$, then it is an optimal solution.
    If CABS returns $\text{NULL}$, then the model is infeasible.
\end{corollary}

We prove that CABS terminates when a DyPDL model is finite, monoidal, and acyclic.

\begin{theorem} \label{thm:cabs-terminate-acyclic}
    Given a finite, acyclic, and monoidal DyPDL model, CABS terminates in finite time.
\end{theorem}

\begin{proof}
    When the beam width $b$ is sufficiently large, e.g., equal to the number of reachable states in the model, beam search never reaches line~\ref{alg:beam:prune-width}.
    Since the number of reachable states is finite, $b$ eventually becomes such a large number with finite iterations.
    Suppose that we call beam search with sufficiently large $b$.
    If $\text{complete} = \top$ is returned, we are done.
    Otherwise, beam search should have found a new solution whose cost is better than $\overline{\gamma}$ in line~\ref{alg:beam:update-solution} and reached line~\ref{alg:beam:update-complete}.
    In this case, there exists a solution for the model.
    Since the state transition graph is finite and acyclic, there are a finite number of solutions, and there exists an optimal solution with the cost $\gamma^*$.
    Since $\overline{\gamma}$ decreases after each call if $\text{complete} = \bot$, eventually, $\overline{\gamma}$ becomes $\gamma^*$, and $\text{complete} = \top$ is returned with finite iterations.
    By Theorem~\ref{thm:beam-terminate-acyclic}, each call of beam search terminates in finite time.
    Therefore, CABS terminates in finite time.
\end{proof}

To obtain a dual bound from beam search, we need to slightly modify Theorem~\ref{thm:search-dual-bound};
since beam search may discard states leading to optimal solutions, we need to keep track of the minimum (or maximum for maximization) $g^l(S) \times \eta(S)$ value for all discarded states in addition to states in $O$.

\begin{theorem} \label{thm:beam-dual-bound}
    Let $\langle A, \times, \mathbf{1} \rangle$ be a monoid where $A \subseteq \mathbb{Q} \cup \{ -\infty, \infty \}$ and $A$ is isotone.
    Given a monoidal DyPDL model $\langle \mathcal{V}, S^0, \mathcal{T}, \mathcal{B}, \mathcal{C} \rangle$ with $\langle A, \times, \mathbf{1} \rangle$ and $\overline{\gamma} \in A$, let $D_m$ be the set of states dropped in layer $m \leq l - 1$ by line~\ref{alg:beam:prune-width} of Algorithm~\ref{alg:beam}.
    If an optimal solution for the model exists and has the cost $\gamma^*$, just after line~\ref{alg:beam:prune-open},
    \begin{equation*}
        \min\left\{ \overline{\gamma}, \min_{S \in O} g^l(S) \times \eta(S), \min_{m = 1, ..., l-1} \min_{S \in D_m} g^m(S) \times \eta(S) \right\} \leq \gamma^*
    \end{equation*}
    where we assume $\min_{S \in O} g^l(S) \times \eta(S) = \infty$ if $O = \emptyset$ and $\min_{S \in D_m} g^m(S) \times \eta(S) = \infty$ if $D_m = \emptyset$.
\end{theorem}

\begin{proof}
    If $\overline{\gamma} \leq \gamma^*$, the inequality holds trivially, so we assume $\overline{\gamma} > \gamma^*$.
    We prove that there exists a state $\hat{S} \in O \cup \bigcup_{m=1}^{l-1} D_m$ on an optimal path, i.e., there exists an $\hat{S}$-solution $\hat{\sigma}$ such that $\langle \sigma^m(\hat{S}); \hat{\sigma} \rangle$ is an optimal solution where $m \in \{ 0, ..., l \}$.
    Initially, $O = \{ S^0 \}$, so the condition is satisfied.
    Suppose that a state $\hat{S}$ on an optimal path is included in $O$ just before line~\ref{alg:beam:for}.
    If $\hat{S}$ is a base state, we reach line~\ref{alg:beam:solution-cost}, and $\text{current\_cost} = \gamma^*$.
    Since $\overline{\gamma} > \gamma^*$, $\overline{\gamma}$ is updated to $\gamma^*$ in line~\ref{alg:beam:update-solution}.
    Then, $\overline{\gamma} = \gamma^* \leq \gamma^*$ will hold after line~\ref{alg:beam:prune-open}.
    If $\hat{S}$ is not a base state, by a similar argument to the proof of Theorem~\ref{thm:beam-optimality} (or Theorem~\ref{thm:search-optimality} if we consider the modified version where states in all layers are kept in $G$), a state on an optimal path, $S'$, will be included in $G$ just before line~\ref{alg:beam:prune-open}.
    Since $g^l(S') \times \eta(S') \leq \gamma^* < \overline{\gamma}$,
    $S' \in O$ holds after line~\ref{alg:beam:prune-open}, and $\min_{S \in O} g^l(S) \times \eta(S) \leq g^l(S') \times \eta(S') \leq \gamma^*$.
    After line~\ref{alg:beam:prune-width}, $S'$ will be included in either $O$ or $D_l$, which can be considered a new $\hat{S}$.
    Suppose that $\hat{S} \in D_m$ just before line~\ref{alg:beam:for}.
    Since $\hat{S}$ is never removed from $D_m$, $\min_{S \in D_m} g^m(S) \times \eta(S) \leq g^m(\hat{S}) \times \eta(\hat{S}) \leq \gamma^*$ always holds.
    By mathematical induction, the theorem is proved.
\end{proof}

\section{DyPDL Models for Combinatorial Optimization Problems} \label{sec:models}

To show the flexibility of DyPDL, in addition to TSPTW and talent scheduling, we formulate DyPDL models for NP-hard combinatorial optimization problems from different application domains such as routing, packing, scheduling, and manufacturing.
We select problem classes whose DyPDL models can be solved by our heuristic search solvers.
The models are monoidal with isotonicity, which guarantees the optimality of the heuristic search solvers, as shown in Theorem~\ref{thm:search-optimality} and Corollary~\ref{cor:cabs-optimality}.
While some models are cost-algebraic and others are not, all of them are acyclic, so the heuristic search solvers terminate in finite time, as shown in Theorems~\ref{thm:search-terminate-acyclic} and \ref{thm:cabs-terminate-acyclic}.
We diversify the problem classes with the following criteria:
\begin{itemize}
        \item Both minimization and maximization problems are included.
        \item DyPDL models with different binary operators for the cost expressions are included: addition ($+$) and taking the maximum ($\max$).
        \item Each of DIDP, MIP, and CP outperforms the others in at least one problem class as shown in Section~\ref{sec:experiment}.
\end{itemize}
We present DyPDL models for six problem classes satisfying the above criteria in this section and three in \ref{sec:cp-models}.
For some of the problem classes, problem-specific DP approaches were previously proposed, and our DyPDL models are based on them.
To concisely represent the models, we only present the Bellman equations since all models are finite and acyclic and satisfy the Principle of Optimality in Definition~\ref{def:principle}.
The YAML-DyPDL files for the models are publicly available in our repository.\footnote{\url{https://github.com/Kurorororo/didp-models}}

\subsection{Capacitated Vehicle Routing Problem (CVRP)} \label{sec:cvrp}

In the capacitated vehicle routing problem (CVRP) \cite{Dantzig1959}, customers $N = \{ 0,...,n-1 \}$, where $0$ is the depot, are given, and each customer $i \in N \setminus \{ 0 \}$ has the demand $d_i \geq 0$.
A solution is a tour to visit each customer in $N \setminus \{ 0 \}$ exactly once using $m$ vehicles, which start from and return to the depot.
The sum of demands of customers visited by a single vehicle must be less than or equal to the capacity $q$.
We assume $d_i \leq q$ for each $i \in N$.
Visiting customer $j$ from $i$ requires the travel time $c_{ij} \geq 0$, and the objective is to minimize the total travel time.
CVRP is strongly NP-hard because it generalizes TSP \cite{Toth2002}.

We formulate the DyPDL model based on the giant-tour representation \cite{Gromicho2012}.
We sequentially construct tours for the $m$ vehicles.
Let $U$ be a set variable representing unvisited customers, $i$ be an element variable representing the current location, $l$ be a numeric variable representing the current load, and $k$ be a numeric variable representing the number of used vehicles.
Both $l$ and $k$ are resource variables where less is preferred.
At each step, one customer $j$ is visited by the current vehicle or a new vehicle.
When a new vehicle is used, $j$ is visited via the depot, $l$ is reset, and $k$ is increased.
Similar to TSPTW, let $c^{\text{in}}_j = \min_{k \in N \setminus \{ j \}} c_{kj}$ and $c^{\text{out}}_j = \min_{k \in N \setminus \{ j \}} c_{jk}$.
\begin{align}
         & \text{compute } V(N \setminus \{ 0 \}, 0, 0, 1)                                                                                                                                                                                                                                                                            \\
         & V(U, i, l, k) =                                                                                                                                               \begin{cases}
                                                                                                                                                                                 \infty                                                                                   & \text{if } (m - k + 1)q < l + \sum\limits_{j \in U} d_j \\
                                                                                                                                                                                 c_{i0}                                                                                   & \text{else if } U = \emptyset                           \\
                                                                                                                                                                                 \min\left\{ \begin{array}{l}
                                    \min\limits_{j \in U : l + d_j \leq q} c_{ij} + V(U \setminus \{ j \}, j, l + d_j, k) \\
                                    \min\limits_{j \in U} c_{i0} + c_{0j} + V(U \setminus \{ j \}, j, d_j, k+1)
                            \end{array} \right. & \text{else if } \exists j \in U, l + d_j \leq q \land k < m                                                  \\
                                                                                                                                                                                 \min\limits_{j \in U : l + d_j \leq q} c_{ij} + V(U \setminus \{ j \}, j, l + d_j, k)    & \text{else if } \exists j \in U : l + d_j \leq q        \\
                                                                                                                                                                                 \min\limits_{j \in U} c_{i0} + c_{0j} + V(U \setminus \{ j \}, j, d_j, k+1)              & \text{else if } k < m                                   \\
                                                                                                                                                                                 \infty                                                                                   & \text{else}
                                                                                                                                                                         \end{cases} \label{eqn:cvrp:bellman} \\
         & V(U, i, l, k) \leq V(U, i, l', k') ~ \quad \quad \quad \quad \quad \quad \quad \quad \quad \quad \quad \quad \quad \quad \text{if } l \leq l' \land k \leq k' \label{eqn:cvrp:dominance}                                                                                                                                   \\
         & V(U, i, l, k) \geq \max\left\{ \sum_{j \in U \cup \{ 0 \}} c^{\text{in}}_j, \sum_{j \in U \cup \{ i \}} c^{\text{out}}_j  \right\}. \label{eqn:cvrp:dual-bound}
\end{align}
The first line of Equation~\eqref{eqn:cvrp:bellman} represents a state constraint:
in a state, if the sum of capacities of the remaining vehicles  ($(m-k+1)q$) is less than the sum of the current load ($l$) and the demands of the unvisited customers ($\sum_{j \in U} d_j$), it does not lead to a solution.
The second line is a base case where all customers are visited.
The model has two types of transitions:
directly visiting customer $j$, which is applicable when the current vehicle has sufficient space ($l + d_j \leq q$), and visiting $j$ with a new vehicle from the depot, which is applicable when there is an unused vehicle ($k < m$).
The third line is active when both of them are possible, and the fourth and fifth lines are active when only one of them is possible.
Recall that a state $S$ dominates another state $S'$ iff for any $S'$-solution, there exists an equal or better $S$-solution with an equal or shorter length in Definition~\ref{def:dominance}.
If $l \leq l'$ and $k \leq k'$, any $(U, i, l', k')$-solution is also a $(U, i, l, k)$-solution, so the dominance implied by Inequality~\eqref{eqn:cvrp:dominance} satisfies this condition.
Inequality~\eqref{eqn:cvrp:dual-bound} is a dual bound function defined in the same way as Inequality~\eqref{eqn:tsptw:bound} of the DyPDL model for TSPTW.
Similar to the DyPDL model for TSPTW, this model is cost-algebraic with a cost algebra $\langle \mathbb{Q}^+_0, +, 0 \rangle$.
However, since the base cost $c_{i0}$ is not necessarily zero, the first solution found by CAASDy may not be optimal (Theorem~\ref{thm:a*}).

\subsection{Multi-Commodity Pickup and Delivery TSP (m-PDTSP)} \label{sec:m-pdtsp}

A one-to-one multi-commodity pickup and delivery traveling salesperson problem (m-PDTSP) \cite{Hernandez-Perez2009} is to pick up and deliver commodities using a single vehicle.
Similar to CVRP, m-PDTSP is a generalization of TSP and is strongly NP-hard.
In this problem, customers $N = \{ 0, ..., n-1 \}$, edges $A \subseteq N \times N$, and commodities $M = \{ 0, ..., m-1 \}$ are given.
The vehicle can visit customer $j$ directly from customer $i$ with the travel time $c_{ij} \geq 0$ if $(i, j) \in A$.
Each commodity $k \in M$ is picked up at customer $p_k \in N$ and delivered to customer $d_k \in N$.
The load increases (decreases) by $w_k$ at $p_k$ ($d_k$) and must not exceed the capacity $q$.
The vehicle starts from $0$, visits each customer once, and stops at $n-1$.
We assume that cyclic dependencies between commodities, e.g., $p_k = d_{k'}$ and $p_{k'} = d_k$, do not exist.

We propose a DyPDL model based on the 1-PDTSP reduction \cite{Gouveia2015} and the DP model by \citeauthor{Castro2020}~\cite{Castro2020}.
In a state, a set variable $U$ represents the set of unvisited customers, an element variable $i$ represents the current location, and a numeric resource variable $l$ represents the current load.
The net change of the load at customer $j$ is represented by $\delta_j = \sum_{k \in M : p_k = j} w_k - \sum_{k \in M : d_k = j} w_k$, and the customers that must be visited before $j$ is represented by $P_j = \{ p_k \mid k \in M : d_k = j \}$, both of which can be precomputed.
The set of customers that can be visited next is $X(U, i, l) = \{ j \in U \mid (i, j) \in A \land l + \delta_j \leq q \land P_j \cap U = \emptyset \}$.
Let $c^{\text{in}}_j = \min_{k \in N : (k, j) \in A} c_{kj}$ and $c^{\text{out}}_j = \min_{k \in N : (j, k) \in A} c_{jk}$.
\begin{align}
         & \text{compute } V(N \setminus \{ 0, n-1 \}, 0, 0)                                                                                                                \\
         & V(U, i, l) = \begin{cases}
                                c_{i,n-1}                                                                       & \text{if } U = \emptyset \land (i, n-1) \in A \\
                                \min\limits_{j \in X(U, i, l)} c_{ij} + V(U \setminus \{ j \}, j, l + \delta_j) & \text{else if } X(U, i, l) \neq \emptyset     \\
                                \infty                                                                          & \text{else}
                        \end{cases}                     \\
         & V(U, i, l) \leq V(U, i, l') \quad \quad \quad \quad \quad \quad \quad \quad \quad \quad \ \ \text{if } l \leq l' \label{eqn:m-pdtsp:dominance}                   \\
         & V(U, i, l) \geq \max\left\{\sum_{j \in U \cup \{ n-1 \} } c^{\text{in}}_j, \sum_{j \in U \cup \{ i \}} c^{\text{out}}_j \right\}. \label{eqn:m-pdtsp:dual-bound}
\end{align}
Similarly to CVRP, Inequalities \eqref{eqn:m-pdtsp:dominance} and \eqref{eqn:m-pdtsp:dual-bound} represent resource variables and a dual bound function, the model is cost-algebraic with a cost algebra $\langle \mathbb{Q}^+_0, +, 0 \rangle$, and the base cost $c_{i,n-1}$ is not necessary zero.

\subsection{Orienteering Problem with Time Windows (OPTW)} \label{sec:optw}

In the orienteering problem with time windows (OPTW) \cite{Kantor1992}, customers $N = \{ 0,...,n-1 \}$ are given, where $0$ is the depot.
Visiting customer $j$ from $i$ requires the travel time $c_{ij} > 0$ while producing the integer profit $p_j \geq 0$.
Each customer $j$ can be visited only in the time window $[a_j, b_j]$, and the vehicle needs to wait until $a_j$ upon earlier arrival.
The objective is to maximize the total profit while starting from the depot at time $t=0$ and returning to the depot by $b_0$.
OPTW is strongly NP-hard because it is a generalization of the orienteering problem, which is NP-hard \cite{Golden1987}.

Our DyPDL model is similar to the DP model by \citeauthor{Righini2009}~\cite{Righini2009} but designed for DIDP with forced transitions and a dual bound function.
A set variable $U$ represents the set of customers to visit, an element variable $i$ represents the current location, and a numeric resource variable $t$ represents the current time, where less is preferred.
We visit customers one by one using transitions.
Customer $j$ can be visited next if it can be visited and the depot can be reached by the deadline after visiting $j$.
Let $c^*_{ij}$ be the shortest travel time from $i$ to $j$.
Then, the set of customers that can be visited next is $X(U, i, t) = \{ j \in U \mid t + c_{ij} \leq b_j \land t + c_{ij} + c^*_{j0} \leq b_0 \}$.
In addition, we remove a customer that can no longer be visited using a forced transition.
If $t + c^*_{ij} > b_j$, then we can no longer visit customer $j$.
If $t + c^*_{ij} + c^*_{j0} > b_0$, then we can no longer return to the depot after visiting $j$.
Thus, the set of unvisited customers that can no longer be visited is represented by $Y(U, i, t) = \{ j \in U \mid t + c^*_{ij} > b_j \lor t + c^*_{ij} + c^*_{j0} > b_0 \}$.
The set $Y(U, i, t)$ is not necessarily equivalent to $U \setminus X(U, i, t)$ since it is possible that $j$ cannot be visited directly from $i$ but can be visited via another customer when the triangle inequality does not hold.

If we take the sum of profits over $U \setminus Y(U, i, t)$, we can compute an upper bound on the value of the current state.
In addition, we use another upper bound considering the remaining time limit $b_0 - t$.
We consider a relaxed problem, where the travel time to customer $j$ is always $c^{\text{in}}_j = \min_{k \in N \setminus \{ j \}} c_{kj}$.
This problem can be viewed as the well-known 0-1 knapsack problem \cite{Martello1990,Kellerer2004}, which is to maximize the total profit of items included in a knapsack such that the total weight of the included items does not exceed the capacity of the knapsack.
Each customer $j \in U \setminus Y(U, i, t)$ is an item with the profit $p_j$ and the weight $c^{\text{in}}_j$, and the capacity of the knapsack is $b_0 - t - c^{\text{in}}_0$ since we need to return to the depot.
Then, we can use the Dantzig upper bound \cite{Dantzig1957}, which sorts items in the descending order of the efficiency $e^{\text{in}}_j = p_j / c^{\text{in}}_j$ and includes as many items as possible.
When an item $k$ exceeds the remaining capacity $q$, then it is included fractionally, i.e., the profit is increased by $\lfloor q e^{\text{in}}_k \rfloor$.
This procedural upper bound is difficult to represent efficiently with the current YAML-DyPDL due to its declarative nature.
Therefore, we further relax the problem by using $\max_{j \in U \setminus Y(U, i, t)} e^{\text{in}}_j$ as the efficiencies of all items, i.e., we use $\lfloor (b_0 - t - c^{\text{in}}_0) \max_{j \in U \setminus Y(U, i, t)} e^{\text{in}}_j \rfloor$ as an upper bound.
Similarly, based on $c^{\text{out}}_j = \min_{k \in N \setminus \{ j \}} c_{jk}$, the minimum travel time from $j$, we also use $\lfloor (b_0 - t - c^{\text{out}}_i) \max_{j \in U \setminus Y(U, i, t)} e^{\text{out}}_j \rfloor$ where $e^{\text{out}}_j = p_j / c^{\text{out}}_j$.
\begin{align}
         & \text{compute } V(N \setminus \{ 0 \}, 0, 0)                                                                                                                                                                                                                                                                                        \\
         & V(U, i, t) = \begin{cases}
                                0                                                                                         & \text{if } t + c_{i0} \leq b_0 \land U = \emptyset           \\
                                V(U \setminus \{ j \}, i, t)                                                              & \text{else if } \exists j \in Y(U, i, t)                     \\
                                V(U \setminus \{ j \}, i, t)                                                              & \text{else if } \exists j \in U \land X(U, i, t) = \emptyset \\
                                \max\limits_{j \in X(U, i, t)} p_j + V(U \setminus \{ j \}, j, \max\{ t + c_{ij}, a_j \}) & \text{else if } X(U, i, t) \neq \emptyset                    \\
                                -\infty                                                                                   & \text{else}
                        \end{cases} \label{eqn:optw:bellman}                                                                                                                                                               \\
         & V(U, i, t) \geq V(U, i, t') ~ \quad \quad \quad \quad \quad \quad \quad \quad \quad \quad \quad \quad \quad \quad \text{if } t \leq t' \label{eqn:optw:dominance}                                                                                                                                                                   \\
         & V(U, i, t) \leq \min\left\{ \sum_{j \in U \setminus Y(U, i, t)} p_j, \left\lfloor (b_0 - t - c^{\text{in}}_0) \max_{j \in U \setminus Y(U, i, t)} e^{\text{in}}_j \right\rfloor, \left\lfloor (b_0 - t - c^{\text{out}}_i) \max_{j \in U \setminus Y(U, i, t)} e^{\text{out}}_j \right\rfloor \right\}. \label{eqn:optw:dual-bound}
\end{align}
The second line of Equation~\eqref{eqn:optw:bellman} removes an arbitrary customer $j$ in $Y(U, i, t)$, which is a forced transition.
The third line also defines a forced transition to remove a customer $j$ in $U$ when no customer can be visited directly ($X(U, i, t) = \emptyset$);
in such a case, even if $j \in U \setminus Y(U, i, t)$, i.e., $j + c^*_{ij} \leq b_j$, the shortest path to customer $j$ is not available.
The base case (the first line of Equation~\eqref{eqn:optw:bellman}) becomes active when all customers are visited or removed.
This condition forces the vehicle to visit as many customers as possible.
Since each transition removes one customer from $U$, and all customers must be removed in a base state, all $(U, i, t)$- and $(U, i, t')$-solutions have the same length.
If $t \leq t'$, more customers can potentially be visited, so $(U, i, t)$ leads to an equal or better solution than $(U, i, t')$.
Thus, the dominance implied by Inequality~\eqref{eqn:optw:dominance} satisfies Definition~\ref{def:dominance}.
The cost expressions are represented by the addition of nonnegative values, so the model is monoidal with a monoid $\langle \mathbb{Q}_0^+, +, 0 \rangle$, and $\mathbb{Q}_0^+$ is isotone.
However, the model is not cost-algebraic since it is maximization and $\forall x \in \mathbb{Q}_0^+, 0 \geq x$ does not hold.
Thus, the first solution found by CAASDy may not be optimal.

\subsection{Bin Packing} \label{sec:bin-packing}

In the bin packing problem \cite{Martello1990}, items $N = \{ 0, ..., n-1 \}$ are given, and each item $i$ has weight $w_i$.
The objective is to pack items in bins with the capacity $q$ while minimizing the number of bins.
We assume $q \geq w_i$ for each $i \in N$.
Bin packing is strongly NP-hard \cite{Garey1990}.

In our DyPDL model, we pack items one by one.
A set variable $U$ represents the set of unpacked items, and a numeric resource variable $r$ represents the remaining space in the current bin, where more is preferred.
In addition, we use an element resource variable $k$ representing the number of used bins, where less is preferred.
The model breaks symmetry by packing item $i$ in the $i$-th or an earlier bin.
Thus, $X(U, r, k) = \{ i \in U \mid r \geq w_i \land i + 1 \geq k \}$ represents items that can be packed in the current bin.
When $\forall j \in U, r < w_j$, then a new bin is opened, and any item in $Y(U, k) = \{ i \in U \mid i \geq k \}$ can be packed;
it is a forced transition.

For a dual bound function, we use lower bounds, LB1, LB2, and LB3, used by \citeauthor{Jonhson1988}~\cite{Jonhson1988}.
The first lower bound, LB1, is $\left\lceil (\sum_{i \in U} w_i - r) / q \right\rceil$, which relaxes the problem by allowing splitting an item across multiple bins.
The second lower bound, LB2, only considers large items in $\{ i \in U \mid w_i \geq q/2 \}$.
If $w_i > q/2$, item $i$ cannot be packed with other large items.
If $w_i = q/2$, at most one additional item $j$ with $w_j = q/2$ can be packed.
Let $a_i = 1$ if $w_i > q/2$ and $a_i = 0$ otherwise.
Let $b_i = 1/2$ if $w_i = q/2$ and $b_i = 0$ otherwise.
The number of bins is lower bounded by $\sum_{i \in U} a_i + \left\lceil \sum_{i \in U} b_i \right\rceil - \mathbbm{1}\left(r \geq \frac{q}{2}\right)$, where $\mathbbm{1}$ is an indicator function that returns 1 if the given condition is true and 0 otherwise.
The last term considers the case when an item with $w_i \geq q/2$ can be packed in the current bin.
Similarly, LB3 only considers items in $\{ i \in U \mid w_i \geq q/3 \}$.
Let $c = 1$ if $w_i > 2q/3$, $c_i = 2/3$ if $w_i = 2q/3$, $c_i = 1/2$ if $q/3 < w_i < 2q/3$, $c_i = 1/3$ if $w_i = q/3$, and $c_i = 0$ otherwise.
The number of bins is lower bounded by $\left\lceil \sum_{i \in U} c_i \right\rceil - \mathbbm{1}\left(r \geq \frac{q}{3}\right)$.

\begin{align}
         & \text{compute } V(N, 0, 0)                                                                                                                            \\
         & V(U, r, k) =  \begin{cases}
                                 0                                                               & \text{if } U = \emptyset                                             \\
                                 1 + V(U \setminus \{ i \}, q-w_i, k + 1)                        & \text{else if } \exists i \in Y(U, k) \land \forall j \in U, r < w_j \\
                                 \min\limits_{i \in X(U, r, k)} V(U \setminus \{ i \}, r-w_i, k) & \text{else if } X(U, r, k) \neq \emptyset                            \\
                                 \infty                                                          & \text{else}
                         \end{cases}  \\
         & V(U, r, k) \leq V(U, r', k') ~ \quad \quad \quad \quad \quad \quad \quad \quad \text{if } r \geq r' \land k \leq k' \label{eqn:bin-packing:dominance} \\
         & V(U, r, k) \geq  \max \left\{ \begin{array}{l}
                                                 \left\lceil (\sum_{i \in U} w_i - r) / q \right\rceil                                                         \\
                                                 \sum_{i \in U} a_i + \left\lceil \sum_{i \in U} b_i \right\rceil - \mathbbm{1}\left(r \geq \frac{q}{2}\right) \\
                                                 \left\lceil \sum_{i \in U} c_i \right\rceil - \mathbbm{1}\left(r \geq \frac{q}{3}\right).
                                         \end{array} \right. \label{eqn:bin-packing:dual-bound}
\end{align}
Since each transition packs one item, any $(U, r, k)$- and $(U, r', k')$ solutions have the same length.
It is easy to see that $(U, r, k)$ leads to an equal or better solution than $(U, r', k')$ if $r \geq r'$ and $k \leq k'$, so the dominance implied by Inequality~\eqref{eqn:bin-packing:dominance} is valid.
This model is cost-algebraic with a cost algebra $\langle \mathbb{Z}^+_0, +, 0 \rangle$.
Since the base cost is always zero, the first solution found by CAASDy is optimal by Theorem~\ref{thm:a*}.

\subsection{Simple Assembly Line Balancing Problem (SALBP-1)} \label{sec:salbp-1}

The variant of the simple assembly line balancing problem (SALBP) called SALBP-1 \cite{Salveson1955,Baybars1986} is the same as bin packing except for precedence constraints.
In SALBP-1, we are given set of tasks $N = \{ 0, ..., n-1 \}$, and each task $i$ has a processing time $w_i$.
A task is scheduled in a station, and the sum of the processing times of tasks in a station must not exceed the cycle time $q$.
Stations are ordered, and each task must be scheduled in the same or later station than its predecessors $P_i \subseteq N$.
SALBP-1 is strongly NP-hard because it is a generalization of bin packing \cite{Alvalez-Miranda2019}.

We formulate a DyPDL model based on that of bin packing and inspired by a problem-specific heuristic search method for SALBP-1 \cite{Sewell2012,Morrison2014}.
Due to the precedence constraint, we cannot schedule an arbitrary item when we open a station unlike bin packing.
Thus, we do not use an element resource variable $k$.
Now, the set of tasks that can be scheduled in the current station is represented by $X(U, r) = \{ i \in U \mid r \geq w_i \land P_i \cap U = \emptyset \}$.
We introduce a transition to open a new station only when $X(U, r) = \emptyset$, which is called a maximum load pruning rule in the literature \cite{Jackson1956,Scholl1997}.
Since bin packing is a relaxation of SALBP-1, we can use the dual bound function for bin packing.
\begin{align}
         & \text{compute } V(N, 0)                                                                                                                  \\
         & V(U, r) = \begin{cases}
                             0                                                         & \text{if } U = \emptyset            \\
                             1 + V(U, q)                                               & \text{else if } X(U, r) = \emptyset \\
                             \min\limits_{i \in X(U, r)} V(U \setminus \{ i \}, r-w_i) & \text{else}
                     \end{cases}                                \\
         & V(U, r) \leq V(U, r')
        ~~~ \quad \quad \quad \quad \quad \quad \quad \text{if } r \geq r' \label{eqn:salbp-1:dominance}                                            \\
         & V(U, r) \geq  \max \left\{ \begin{array}{l}
                                              \left\lceil (\sum_{i \in U} w_i - r) / q \right\rceil                                                         \\
                                              \sum_{i \in U} a_i + \left\lceil \sum_{i \in U} b_i \right\rceil - \mathbbm{1}\left(r \geq \frac{q}{2}\right) \\
                                              \left\lceil \sum_{i \in U} c_i \right\rceil - \mathbbm{1}\left(r \geq \frac{q}{3}\right).
                                      \end{array} \right.
\end{align}
The length of a $(U, r)$-solution is the sum of $|U|$ and the number of stations opened, which is the cost of that solution.
Therefore, if $r \geq r'$, then state $(U, r)$ leads to an equal or better and shorter solution than $(U, r')$, so the dominance implied by Inequality~\eqref{eqn:salbp-1:dominance} is valid.
Similar to bin packing, this model is cost-algebraic, and the base cost is zero, so the first solution found by CAASDy is optimal.

\subsection{Graph-Clear} \label{sec:graph-clear}

In the graph-clear problem \cite{Kolling2007}, an undirected graph $(N, E)$ with the node weight $a_i$ for $i \in N$ and the edge weight $b_{ij}$ for $\{ i, j \} \in E$ is given.
In the beginning, all nodes are contaminated.
In each step, one node $c$ can be made clean by sweeping it using $a_c$ robots and blocking each edge $\{ c, i \}$ using $b_{ci}$ robots.
However, while sweeping a node, an already swept node becomes contaminated if it is connected by a path of unblocked edges to a contaminated node.
The optimal solution minimizes the maximum number of robots per step to make all nodes clean.
This optimization problem is NP-hard because finding a solution whose cost is smaller than a given value is NP-complete \cite{Kolling2007}.

Previous work \cite{Morin2018} proved that there exists an optimal solution in which a swept node is never contaminated again.
Based on this observation, the authors developed a state-based formula as the basis for MIP and CP models.
We use the state-based formula directly as a DyPDL model.
A set variable $C$ represents swept nodes, and one node in $N \setminus C$ is swept at each step.
We block all edges connected to $c$ and all edges from contaminated nodes to already swept nodes.
We assume that $b_{ij} = 0$ if $\{ i, j \} \notin E$.
\begin{align}
         & \text{compute } V(\emptyset)   \\
        \begin{split}
                 & V(C) = \begin{cases}
                                  0                                                                                                                                                                                                & \text{if } C = N \\
                                  \min\limits_{c \in N \setminus C} \max\left\{ a_c + \sum\limits_{i \in N} b_{ci} + \sum\limits_{i \in C}\sum\limits_{j \in (N \setminus C) \setminus \{ c \}} b_{ij}, V(C \cup \{ c \}) \right\} & \text{else}
                          \end{cases}
        \end{split} \\
         & V(C) \geq 0.
\end{align}
Viewing the maximum of two values ($\max$) as a binary operator, $\langle \mathbb{Z}_0^+, \max, 0 \rangle$ is a monoid since $\max\{x ,y \} \in \mathbb{Z}_0^+$, $\max\{ x, \max\{ y, z \} \} = \max\{ \max\{ x, y \}, z \}$, and $\max\{ x, 0 \} = \max\{ 0, x \} = x$ for $x, y, z \in \mathbb{Z}_0^+$.
It is isotone since $x \leq y \rightarrow \max\{ x, z \} \leq \max \{ y, z \}$ and $x \leq y \rightarrow \max\{ z, x \} \leq \max\{ z, y \}$.
Since $\forall x \in \mathbb{Z}_0^+, 0 \leq x$, $\langle \mathbb{Z}_0^+, \max, 0 \rangle$ is a cost algebra, so the DyPDL model is cost-algebraic.
Since the base cost is always zero, the first solution found by CAASDy is optimal.

\section{Experimental Evaluation} \label{sec:experiment}

We implement and experimentally evaluate DIDP solvers using the heuristic search algorithms described in Section~\ref{sec:algorithms}.
We compare our DIDP solvers with commercial MIP and CP solvers, Gurobi 11.0.2 \cite{gurobi} and IBM ILOG CP Optimizer 22.1.0 \cite{Laborie2018}.
We select state-of-the-art MIP and CP models in the literature when multiple models exist and develop a new model when we do not find an existing one.
We also compare DIDP with existing state-based general-purpose solvers, domain-independent AI planners, a logic programming language, and a decision diagram-based (DD-based) solver.

\subsection{Software Implementation of DIDP} \label{sec:software}

We develop didp-rs v0.7.0,\footnote{\url{https://github.com/domain-independent-dp/didp-rs/releases/tag/v0.7.0}} a software implementation of DIDP in Rust.
It has four components, dypdl,\footnote{\url{https://crates.io/crates/dypdl}} dypdl-heuristic-search,\footnote{\url{https://crates.io/crates/dypdl-heuristic-search}} didp-yaml,\footnote{\url{https://crates.io/crates/didp-yaml}} and DIDPPy.\footnote{\url{https://didppy.readthedocs.io}}
The library dypdl is for modeling, and dypdl-heuristic-search is a library for heuristic search solvers.
The commandline interface didp-yaml takes YAML-DyPDL domain and problem files and a YAML file specifying a solver as input and returns the result.
DIDPPy is a Python interface whose modeling capability is equivalent to didp-yaml.
In our experiment, we use didp-yaml.

As DIDP solvers, dypdl-heuristic-search implements CAASDy, DFBnB, CBFS, ACPS, APPS, DBDFS, and CABS.
These solvers can handle monoidal DyPDL models with a monoid $\langle A, \times, \mathbf{1} \rangle$ where $A \subseteq \mathbb{Q} \cup \{-\infty, \infty\}$, $\times \in \{ +, \max \}$, and $\mathbf{1} = 0$ if $\times = +$ or $\mathbf{1}$ is the minimum value in $A$ if $\times = \max$.

In all solvers, we use the dual bound function provided with a DyPDL model as a heuristic function.
Thus, $f(S) = g(S) \times h(S) = g(S) \times \eta(S)$.
By Theorem~\ref{thm:search-dual-bound}, the best $f$-value in the open list is a dual bound.
In CAASDy, states in the open list are ordered by the $f$-values in a binary heap, so a dual bound can be obtained by checking the top of the binary heap.
Similarly, in DFBnB, CBFS, and ACPS, since states with each depth are ordered by the $f$-values, by keeping track of the best $f$-value in each depth, we can compute a dual bound.
In APPS, when the set of the best states $O_b$ and the set of the best successor states $O_c$ become empty, the best $f$-value of states in the suspend list $O_s$ is a dual bound, where states are ordered by the $f$-values.
In DBDFS, we keep track of the best $f$-value of states inserted into $O_1$ and use it as a dual bound when $O_0$ becomes empty.
In CABS, based on Theorem~\ref{thm:beam-dual-bound}, the best $f$-value of discarded states is maintained, and a dual bound is computed after generating all successor states in a layer.
In CAASDy, CBFS, ACPS, APPS, and CABS, when the $f$- and $h$-values of two states are the same, the tie is broken according to the implementation of the binary heap that is used to implement the open list.
In DFBnB and DBDFS, the open list is implemented with a stack, and successor states are sorted before being pushed to the stack, so the tie-breaking depends on the implementation of the sorting algorithm.
While a dual bound function is provided in each DyPDL model used in our experiment, it is not required in general;
when no dual bound function is provided, the DIDP solvers use the $g$-value instead of the $f$-value to guide the search and do not perform pruning.

As explained in Section~\ref{sec:yaml-transitions}, forced transitions can be explicitly defined in a DyPDL model.
If such transitions are applicable in a state, our solvers keeps only the first defined one $\tau$ in the set of applicable transitions, i.e., $\mathcal{T}^*(S) = \{ \tau \}$ in Algorithms~\ref{alg:search} and \ref{alg:beam}.
Otherwise, no forced transitions are considered, i.e., $\mathcal{T}^*(S) = \mathcal{T}(S)$.

\subsection{Benchmarks} \label{sec:benchmarks}

We describe benchmark instances and MIP and CP models for nine problem classes.
The DyPDL model for TSPTW is presented in Section~\ref{sec:dp} as a running example, and the models for CVRP, m-PDTSP, OPTW, bin packing, SALPB-1, and graph-clear are in Section~\ref{sec:models}.
The other DyPDL models are presented in \ref{sec:cp-models}.
All benchmark instances are in text format, so they are converted to YAML-DyPDL problem files by a Python script.
All instances in one problem class share the same YAML-DyPDL domain file except for the multi-dimensional knapsack problem, where the number of state variables depends on an instance, and thus a domain file is generated for each instance by the Python script.
All instances generated by us, MIP and CP models, YAML-DyPDL domain files, and the Python scripts are available from our repository.\footnote{\url{https://github.com/Kurorororo/didp-models}}

\subsubsection{TSPTW} \label{sec:tsptw-benchmarks}

For TSPTW, we use 340 instances from \citeauthor{Dumas1995}~\cite{Dumas1995}, \citeauthor{Gendreau1998}~\cite{Gendreau1998}, \citeauthor{Ohlmann2007}~\cite{Ohlmann2007}, and \citeauthor{Ascheuer1995}~\cite{Ascheuer1995}, where travel times are integers;
while didp-rs can handle floating point numbers, the CP solver we use, CP Optimizer, does not.
In these instances, the deadline to return to the depot, $b_0$, is defined, but $\forall i \in N, b_i + c_{i0} \leq b_0$ holds, i.e., we can always return to the depot after visiting the final customer.
Thus, in our DyPDL model (Equation~\eqref{eqn:tsptw:transitions} with redundant information in Inequalities~\eqref{eqn:tsptw:infeasibility}, \eqref{eqn:tsptw:dominance}, and \eqref{eqn:tsptw:bound}), $b_0$ is not considered.
For MIP, we use Formulation (1) proposed by \citeauthor{Hungerlander2018}~\cite{Hungerlander2018}.
When there are zero-cost edges, flow-based subtour elimination constraints \cite{Gavish1978} are added.
We adapt a CP model for a single machine scheduling problem with time windows and sequence-dependent setup times \cite{Booth2016} to TSPTW, where an interval variable represents the time to visit a customer.
We change the objective to the sum of travel costs (setup time in their model) and add a $\mathsf{First}$ constraint ensuring that the depot is visited first.

\subsubsection{CVRP}

We use 207 instances in A, B, D, E, F, M, P, and X sets from CVRPLIB \cite{Uchoa2017}.
We use the DyPDL model in Section~\ref{sec:cvrp}, a MIP model proposed by \citeauthor{Gadegaard2021}~\cite{Gadegaard2021}, and a CP model proposed by \citeauthor{Saadaoui2019}~\cite{Saadaoui2019}.

\subsubsection{m-PDTSP}

We use 1178 instances from \citeauthor{Hernandez-Perez2009}~\cite{Hernandez-Perez2009}, which are divided into class1, class2, and class3 sets.
We use the DyPDL model in Section~\ref{sec:m-pdtsp}, the MCF2C+IP formulation for MIP \cite{Letchford2016}, and the CP model proposed by \citeauthor{Castro2020}~\cite{Castro2020}.
In all models, unnecessary edges are removed by a preprocessing method \cite{Letchford2016}.

\subsubsection{OPTW}

We use 144 instances from \citeauthor{Righini2006}~\cite{Righini2006,Righini2008}, \citeauthor{Motemanni2009}~\cite{Motemanni2009}, and \citeauthor{Vansteenwegen2009}~\cite{Vansteenwegen2009}.
In these instances, service time $s_i$ spent at each customer $i$ is defined, so we incorporate it in the travel time, i.e., we use $s_i + c_{ij}$ as the travel time from $i$ to $j$.
We use the MIP model described in \citeauthor{Vansteenwegen2011}~\cite{Vansteenwegen2011}.
For CP, we develop a model similar to that of TSPTW, described in \ref{sec:optw-cp}.

\subsubsection{Multi-Dimensional Knapsack Problem (MDKP)}

We use 276 instances of the multi-dimensional knapsack problem (MDKP) \cite{Martello1990,Kellerer2004} from OR-Library \cite{Beasley1990}, excluding one instance that has fractional item weights;
while the DIDP solvers can handle fractional weights, the CP solver does not.
We use a DyPDL model in \ref{sec:mdkp-dp} and the MIP model described in \citeauthor{Cacchiani2022}~\cite{Cacchiani2022}.
For CP, we develop a model using the $\mathsf{Pack}$ global constraint \cite{Shaw2004} for each dimension (see \ref{sec:mdkp-cp}).

\subsubsection{Bin Packing}

We use 1615 instances in BPPLIB \cite{Delorme2018}, proposed by \citeauthor{Falkenauer1996}~\cite{Falkenauer1996} (Falkenauer U and Falkenauer T), \citeauthor{SchollKJ1997}~\cite{SchollKJ1997} (Scholl 1, Scholl 2, and Scholl 3), \citeauthor{Wascher1996}~\cite{Wascher1996} (W\"{a}scher), \citeauthor{Schwerin1997}~\cite{Schwerin1997} (Schwerin 1 and Schwerin 2), and \citeauthor{Schoenfield2002}~\cite{Schoenfield2002} (Hard28).
We use the DyPDL model in Section~\ref{sec:bin-packing} and the MIP model by \citeauthor{Martello1990}~\cite{Martello1990} extended with inequalities ensuring that bins are used in order of index and item $j$ is packed in the $j$-th bin or earlier as described in \citeauthor{Delorme2016}~\cite{Delorme2016}.
We implement a CP model using $\mathsf{Pack}$ while ensuring that item $j$ is packed in bin $j$ or before.
For MIP and CP models, the upper bound on the number of bins is computed by the first-fit decreasing heuristic.
We show the CP model in \ref{sec:bpp-cp}.

\subsubsection{SALBP-1}

We use 2100 instances proposed by \citeauthor{Morrison2014}~\cite{Morrison2014}.
We use the DyPDL model in Section~\ref{sec:salbp-1} and the NF4 formulation for MIP \cite{Ritt2018}.
Our CP model is based on \citeauthor{Bukchin2018}~\cite{Bukchin2018} but is implemented with the global constraint $\mathsf{Pack}$ in CP Optimizer as it performs better than the original model (see \ref{sec:salbp-1-cp}).
In addition, the upper bound on the number of stations is computed in the same way as the MIP model instead of using a heuristic.

\subsubsection{Single Machine Total Weighted Tardiness}

We use 375 instances of single machine scheduling to minimize total weighted tardiness ($1||\sum w_i T_i$) \cite{Emmons1969} in OR-Library \cite{Beasley1990} with 40, 50, and 100 jobs.
We use a DyPDL model in \ref{sec:wt-dp} and the formulation with assignment and positional date variables (F4) for MIP \cite{Keha2009}.
For CP, we formulate a model using interval variables, as described in \ref{sec:wt-cp}.
We extract precedence relations between jobs using the method proposed by \citeauthor{Kanet2007}~\cite{Kanet2007} and incorporate them into the DyPDL and CP models but not into the MIP model as its performance is not improved.

\subsubsection{Talent Scheduling}

\citeauthor{DeLaBanda2007}~\cite{DeLaBanda2007} considered instances with 8, 10, 12, 14, 16, 18, 20, 22 actors and 16, 18, ..., 64 scenes, resulting in 200 configurations in total.
For each configuration, they randomly generated 100 instances.
We use the first five instances for each configuration, resulting in 1000 instances in total.
We use an extended version of the DyPDL model presented in Section~\ref{sec:forced} (see \ref{sec:talent-dp}) and a MIP model described in \citeauthor{Qin2016}~\cite{Qin2016}.
For CP, we extend the model used in \citeauthor{Chu2015}~\cite{Chu2015} with the $\mathsf{AllDifferent}$ global constraint \cite{Lauriere1978}, which is redundant but slightly improves the performance in practice, as described in \ref{sec:talent-cp}.
In all models, a problem is simplified by preprocessing as described in \citeauthor{DeLaBanda2007}~\cite{DeLaBanda2007}.

\subsubsection{Minimization of Open Stacks Problem (MOSP)}

We use 570 instances of the minimization of open stacks problem (MOSP) \cite{Yuen1995} from four sets: Constraint Modelling Challenge \cite{ConstraintModellingChallenge}, SCOOP Project,\footnote{\url{https://cordis.europa.eu/project/id/32998}} \citeauthor{Faggioli1998}~\cite{Faggioli1998}, and \citeauthor{Chu2009}~\cite{Chu2009}.
We use a DyPDL model based on a problem-specific algorithm \cite{Chu2009} presented in \ref{sec:mosp-dp}.
The MIP and CP models are proposed by \citeauthor{Martin2021}~\cite{Martin2021}.
From their two MIP models, we select MOSP-ILP-I as it solves more instances optimally in their paper.

\subsubsection{Graph-Clear}

We generated 135 instances using planar and random graphs in the same way as \citeauthor{Morin2018}~\cite{Morin2018}, where the number of nodes in a graph is 20, 30, or 40.
For planar instances, we use a planar graph generator \cite{Fusy2009} with the input parameter of 1000.
We use the DyPDL model in Section~\ref{sec:graph-clear} and MIP and CP models proposed by \citeauthor{Morin2018}~\cite{Morin2018}.
From the two proposed CP models, we select CPN as it solves more instances optimally.

\subsection{Comparison with MIP and CP} \label{sec:result}

We use Rust 1.70.0 for didp-rs and Python 3.10.2 for the Python scripts to convert instances to YAML-DyPDL files and the Python interfaces of Gurobi and CP Optimizer.
All experiments are performed on an Intel Xeon Gold 6418 processor with a single thread, an 8 GB memory limit, and a 30-minute time limit using GNU Parallel \cite{GNUParallel}.

\subsubsection{Coverage}

\begin{table}[tb]
    \caption[Coverage and the number of instances where the memory limit is reached.]{
        Coverage (c.) and the number of instances where the memory limit is reached (m.) in each problem class.
        The coverage of a DIDP solver is in bold if it is higher than MIP and CP, and the higher of MIP and CP is in bold if there is no better DIDP solver.
        The highest coverage is underlined.
    }
    \small
    \centering
    \tabcolsep=0.17em
    \begin{tabular}{lrrrrrrrrrrrrrrrrrrrrr}
        \toprule
                                  & \multicolumn{2}{c}{MIP}  & \multicolumn{2}{c}{CP} & \multicolumn{2}{c}{CAASDy} & \multicolumn{2}{c}{DFBnB} & \multicolumn{2}{c}{CBFS} & \multicolumn{2}{c}{ACPS} & \multicolumn{2}{c}{APPS} & \multicolumn{2}{c}{DBDFS} & \multicolumn{2}{c}{CABS} & \multicolumn{2}{c}{CABS/0}                                                                                                                                                                   \\
        \cmidrule(lr){2-3} \cmidrule(lr){4-5} \cmidrule(lr){6-7} \cmidrule(lr){8-9} \cmidrule(lr){10-11} \cmidrule(lr){12-13} \cmidrule(lr){14-15} \cmidrule(lr){16-17} \cmidrule(lr){18-19} \cmidrule(lr){20-21}
                                  & c.                       & m.                     & c.                         & m.                        & c.                       & m.                       & c.                       & m.                        & c.                       & m.                         & c.                      & m.  & c.                      & m.  & c.                      & m.   & c.                        & m. & c.                       & m. \\
        \midrule
        TSPTW (340)               & 224                      & 0                      & 47                         & 0                         & \textbf{257}             & 83                       & \textbf{242}             & 34                        & \textbf{257}             & 81                         & \textbf{257}            & 82  & \textbf{257}            & 83  & \textbf{256}            & 83   & \underline{\textbf{259}}  & 0  & \underline{\textbf{259}} & 0  \\
        CVRP (207)                & \underline{\textbf{28}}  & 5                      & 0                          & 0                         & 6                        & 201                      & 6                        & 187                       & 6                        & 201                        & 6                       & 201 & 6                       & 201 & 6                       & 201  & 6                         & 0  & 5                        & 3  \\
        m-PDTSP (1178)            & 940                      & 0                      & \underline{\textbf{1049}}  & 0                         & 952                      & 226                      & 985                      & 193                       & 988                      & 190                        & 988                     & 190 & 988                     & 190 & 987                     & 191  & 1035                      & 0  & 988                      & 15 \\
        OPTW (144)                & 16                       & 0                      & 49                         & 0                         & \underline{\textbf{64}}  & 79                       & \underline{\textbf{64}}  & 60                        & \underline{\textbf{64}}  & 80                         & \underline{\textbf{64}} & 80  & \underline{\textbf{64}} & 80  & \underline{\textbf{64}} & 78   & \underline{\textbf{64}}   & 0  & -                        & -  \\
        MDKP (276)                & \underline{\textbf{168}} & 0                      & 6                          & 0                         & 4                        & 272                      & 4                        & 272                       & 5                        & 271                        & 5                       & 271 & 5                       & 271 & 4                       & 272  & 5                         & 1  & -                        & -  \\
        Bin Packing (1615)        & 1160                     & 0                      & \underline{\textbf{1234}}  & 0                         & 922                      & 632                      & 526                      & 1038                      & 1115                     & 431                        & 1142                    & 405 & 1037                    & 520 & 426                     & 1118 & 1167                      & 4  & 242                      & 14 \\
        SALBP-1 (2100)            & 1431                     & 250                    & 1584                       & 0                         & \textbf{1657}            & 406                      & \textbf{1629}            & 470                       & 1484                     & 616                        & \textbf{1626}           & 474 & \textbf{1635}           & 465 & 1404                    & 696  & \underline{\textbf{1802}} & 0  & 1204                     & 1  \\
        $1 || \sum w_i T_i$ (375) & 107                      & 0                      & 150                        & 0                         & \textbf{270}             & 105                      & \textbf{233}             & 8                         & \textbf{272}             & 103                        & \textbf{272}            & 103 & \textbf{265}            & 110 & \textbf{268}            & 107  & \underline{\textbf{288}}  & 0  & -                        & -  \\
        Talent Scheduling (1000)  & 0                        & 0                      & 0                          & 0                         & \textbf{207}             & 793                      & \textbf{189}             & 388                       & \textbf{214}             & 786                        & \textbf{214}            & 786 & \textbf{206}            & 794 & \textbf{205}            & 795  & \underline{\textbf{239}}  & 0  & \textbf{231}             & 0  \\
        MOSP (570)                & 241                      & 14                     & 437                        & 0                         & \textbf{483}             & 87                       & \textbf{524}             & 46                        & \textbf{523}             & 47                         & \textbf{524}            & 46  & \textbf{523}            & 47  & \textbf{522}            & 48   & \underline{\textbf{527}}  & 0  & -                        & -  \\
        Graph-Clear (135)         & 26                       & 0                      & 4                          & 0                         & \textbf{78}              & 57                       & \textbf{99}              & 36                        & \textbf{101}             & 34                         & \textbf{101}            & 34  & \textbf{99}             & 36  & \textbf{82}             & 53   & \underline{\textbf{103}}  & 19 & -                        & -  \\
        \bottomrule
    \end{tabular}
    \label{tab:coverage}
\end{table}

Since all solvers are exact, we evaluate \emph{coverage}, the number of instances where an optimal solution is found and its optimality is proved within time and memory limits.
We include the number of instances where infeasibility is proved in coverage.
We show the coverage of each method in each problem class in Table~\ref{tab:coverage}.
In the table, if a DIDP solver has higher coverage than MIP and CP, it is emphasized in bold.
If MIP or CP is better than all DIDP solvers, its coverage is in bold.
The highest coverage is underlined.
We explain CABS/0 in Tables~\ref{tab:coverage}--\ref{tab:primal-integral} later in Section~\ref{sec:ablation-dual-bound}.

CAASDy, ACPS, APPS, and CABS outperform both MIP and CP in seven problem classes: TSPTW, OPTW, SALBP-1, $1||\sum w_i T_i$, talent scheduling, MOSP, and graph-clear.
In addition, the DIDP solvers except for CAASDy have higher coverage than MIP and CP in the class1 instances of m-PDTSP (145 (CABS) and 144 (others) vs. 128 (MIP and CP)).
Comparing the DIDP solvers, CABS has the highest coverage in all problem classes.
As shown, each DIDP solver except for CABS reaches the memory limit in most of the instances it is unable to solve while CABS rarely reaches the memory limit.
This difference is possibly because CABS needs to store only states in the current and next layers, while other solvers need to store all generated and not dominated states in the open list.

MIP has the highest coverage in CVRP and MDKP, and CP in m-PDTSP and bin packing.
MIP runs out of memory in some instances while CP never does.
In particular, in the MIP model for SALBP-1, the number of decision variables and constraints is quadratic in the number of tasks in the worst case, and MIP reaches the memory limit in 250 instances with 1000 tasks.

\subsubsection{Optimality Gap}

\begin{table}[tb]
    \caption[Average optimality gap in each problem class.]{
        Average optimality gap in each problem class.
        The optimality gap of a DIDP solver is in bold if it is lower than MIP and CP, and the lower of MIP and CP is in bold if there is no better DIDP solver.
        The lowest optimality gap is underlined.
    }
    \small
    \centering
    \tabcolsep=0.53em
    \begin{tabular}{lrrrrrrrrrr}
        \toprule
                                  & MIP                         & CP                          & CAASDy          & DFBnB           & CBFS                        & ACPS                        & APPS            & DBDFS           & CABS                        & CABS/0          \\
        \midrule
        TSPTW (340)               & 0.2200                      & 0.7175                      & 0.2441          & \textbf{0.1598} & \textbf{0.1193}             & \textbf{0.1194}             & \textbf{0.1217} & \textbf{0.1408} & \underline{\textbf{0.1151}} & \textbf{0.2085} \\
        CVRP (207)                & 0.8647                      & 0.9868                      & 0.9710          & \textbf{0.7484} & \textbf{0.7129}             & \textbf{0.7123}             & \textbf{0.7164} & \textbf{0.7492} & \underline{\textbf{0.6912}} & 0.9111          \\
        m-PDTSP (1178)            & 0.1838                      & \underline{\textbf{0.1095}} & 0.2746          & 0.2097          & 0.1807                      & 0.1807                      & 0.1840          & 0.2016          & 0.1599                      & 0.1878          \\
        OPTW (144)                & 0.6650                      & 0.2890                      & 0.5556          & 0.3583          & \underline{\textbf{0.2683}} & \underline{\textbf{0.2683}} & \textbf{0.2778} & 0.3359          & \textbf{0.2696}             & -               \\
        MDKP (276)                & \underline{\textbf{0.0008}} & 0.4217                      & 0.9855          & 0.4898          & 0.4745                      & 0.4745                      & 0.4742          & 0.4854          & 0.4676                      & -               \\
        Bin Packing (1615)        & 0.0438                      & \underline{\textbf{0.0043}} & 0.4291          & 0.0609          & 0.0083                      & 0.0075                      & 0.0105          & 0.0651          & 0.0049                      & 0.7386          \\
        SALBP-1 (2100)            & 0.2712                      & 0.0108                      & 0.2100          & 0.0257          & 0.0115                      & \textbf{0.0096}             & \textbf{0.0094} & 0.0273          & \underline{\textbf{0.0057}} & 0.3695          \\
        $1 || \sum w_i T_i$ (375) & 0.4981                      & 0.3709                      & \textbf{0.2800} & 0.3781          & \textbf{0.2678}             & \textbf{0.2679}             & \textbf{0.2878} & \textbf{0.2845} & \underline{\textbf{0.2248}} & -               \\
        Talent Scheduling (1000)  & 0.8867                      & 0.9509                      & \textbf{0.7930} & \textbf{0.2368} & \textbf{0.1884}             & \textbf{0.1884}             & \textbf{0.2003} & \textbf{0.2462} & \underline{\textbf{0.1697}} & \textbf{0.6667} \\
        MOSP (570)                & 0.3169                      & 0.1931                      & \textbf{0.1526} & \textbf{0.0713} & \textbf{0.0362}             & \textbf{0.0359}             & \textbf{0.0392} & \textbf{0.0655} & \underline{\textbf{0.0200}} & -               \\
        Graph-Clear (135)         & 0.4465                      & 0.4560                      & \textbf{0.4222} & \textbf{0.2359} & \textbf{0.0995}             & \textbf{0.0996}             & \textbf{0.1089} & \textbf{0.2636} & \underline{\textbf{0.0607}} & -               \\
        \bottomrule
    \end{tabular}
    \label{tab:optimality-gap}
\end{table}

We also evaluate \emph{optimality gap}, a relative difference between primal and dual bounds.
The optimality gap measures how close a solver is to prove the optimality in instances that are not optimally solved.
Let $\overline{\gamma}$ be a primal bound and $\underline{\gamma}$ be a dual bound found by a solver.
We define the optimality gap, $\delta\left(\overline{\gamma}, \underline{\gamma}\right)$, as follows:
\begin{equation*}
    \delta\left(\overline{\gamma}, \underline{\gamma}\right) = \begin{cases}
        0                                                                                                                & \text{if } \overline{\gamma} = \underline{\gamma} = 0 \\
        \frac{|\overline{\gamma} - \underline{\gamma}|}{\max\left\{ |\overline{\gamma}|, |\underline{\gamma}|  \right\}} & \text{else.}                                          \\
    \end{cases}
\end{equation*}
The optimality gap is $0$ when the optimality is proved and positive otherwise.
We also use $0$ as the optimality gap when the infeasibility is proved.
In the second line, when the signs of the primal and dual bounds are the same, since $|\overline{\gamma} - \underline{\gamma}| \leq \max\{ |\overline{\gamma}|, |\underline{\gamma}| \}$, the optimality gap never exceeds $1$.
In practice, we observe that primal and dual bounds found are always nonnegative in our experiment.
Therefore, we use $1$ as the optimality gap when either a primal or dual bound is not found.

We show the average optimality gap in Table~\ref{tab:optimality-gap}.
Similar to Table~\ref{tab:coverage}, the optimality gap of a DIDP solver is in bold if it is better than MIP and CP, and the best value is underlined.
ACPS, APPS, and CABS achieve a better optimality gap than MIP and CP in the seven problem classes where they have higher coverage.
In addition, the DIDP solvers except for CAASDy outperform MIP and CP in CVRP, where MIP has the highest coverage;
in large instances of CVRP, MIP fails to find feasible solutions, which results in a high average optimality gap.
Comparing the DIDP solvers, CABS is the best in all problem classes except for OPTW, where CBFS and ACPS are marginally better.
CAASDy is the worst among the DIDP solvers in all problem classes except for $1||\sum w_i T_i$;
similar to MIP in CVRP, CAASDy does not provide a primal bound for any unsolved instances except for eleven instances of m-PDTSP.

\subsubsection{Primal Integral}

\begin{table}[tb]
    \caption[Average primal integral in each problem class.]{
        Average primal integral in each problem class.
        The primal integral of a DIDP solver is in bold if it is lower than MIP and CP, and the lower of MIP and CP is in bold if there is no better DIDP solver.
        The lowest primal integral is underlined.
    }
    \small
    \centering
    \tabcolsep=0.51em
    \begin{tabular}{lrrrrrrrrrr}
        \toprule
                                  & MIP                       & CP                         & CAASDy  & DFBnB           & CBFS            & ACPS            & APPS            & DBDFS          & CABS                        & CABS/0          \\
        \midrule
        TSPTW (340)               & 479.03                    & 48.97                      & 458.26  & \textbf{46.31}  & \textbf{9.49}   & \textbf{10.06}  & \textbf{29.36}  & 56.65          & \underline{\textbf{9.25}}   & \textbf{13.71}  \\
        CVRP (207)                & 1127.55                   & 482.89                     & 1748.23 & \textbf{420.66} & \textbf{423.45} & \textbf{418.29} & \textbf{440.59} & 523.42         & \underline{\textbf{333.68}} & \textbf{335.75} \\
        m-PDTSP (1178)            & 177.60                    & 26.04                      & 333.53  & \textbf{23.37}  & \textbf{6.51}   & \textbf{6.49}   & \textbf{9.23}   & \textbf{17.87} & \underline{\textbf{5.24}}   & \textbf{5.31}   \\
        OPTW (144)                & 438.06                    & \underline{\textbf{15.58}} & 1018.23 & 175.49          & 54.05           & 54.29           & 74.37           & 139.64         & 57.95                       & -               \\
        MDKP (276)                & \underline{\textbf{0.65}} & 15.86                      & 1773.92 & 236.12          & 211.69          & 211.62          & 211.84          & 237.99         & 201.72                      & -               \\
        Bin Packing (1615)        & 88.07                     & 8.05                       & 778.60  & 104.46          & 9.98            & 8.40            & 13.82           & 111.85         & \underline{\textbf{5.04}}   & 11.56           \\
        SALBP-1 (2100)            & 538.79                    & 28.43                      & 383.35  & 35.59           & \textbf{10.83}  & \textbf{7.28}   & \textbf{6.74}   & 38.80          & \underline{\textbf{1.92}}   & \textbf{19.42}  \\
        $1 || \sum w_i T_i$ (375) & 64.89                     & \underline{\textbf{3.49}}  & 513.24  & 136.99          & 111.19          & 103.76          & 105.97          & 97.34          & 71.21                       & -               \\
        Talent Scheduling (1000)  & 106.10                    & \underline{\textbf{18.91}} & 1435.12 & 119.03          & 40.72           & 40.39           & 60.12           & 143.45         & 25.41                       & 50.78           \\
        MOSP (570)                & 95.20                     & 13.01                      & 275.48  & \textbf{4.41}   & \textbf{1.39}   & \textbf{1.20}   & \textbf{1.37}   & \textbf{7.72}  & \underline{\textbf{0.31}}   & -               \\
        Graph-Clear (135)         & 334.87                    & 83.49                      & 764.00  & \textbf{4.63}   & \textbf{0.70}   & \textbf{0.74}   & \textbf{3.45}   & 87.90          & \underline{\textbf{0.37}}   & -               \\
        \bottomrule
    \end{tabular}
    \label{tab:primal-integral}
\end{table}

To evaluate the performance of anytime solvers, we use the \emph{primal integral} \cite{Berthold2013}, which considers the balance between the solution quality and computational time.
For an optimization problem, let $\sigma^t$ be a solution found by a solver at time $t$, $\sigma^*$ be an optimal (or best-known) solution, and $\gamma$ be a function that returns the solution cost.
Assuming $\gamma(\sigma^t) \geq 0$, the \emph{primal gap} function $p$ is
\begin{equation*}
    p(t) = \begin{cases}
        0                                                                                                           & \text{if } \gamma(\sigma^t) = \gamma(\sigma^*) = 0 \\
        1                                                                                                           & \text{if no } \sigma^t                             \\
        \frac{\left| \gamma(\sigma^*) - \gamma(\sigma^t) \right|}{\max\{ |\gamma(\sigma^*)|, |\gamma(\sigma^t)| \}} & \text{else.}
    \end{cases}
\end{equation*}
The primal gap takes a value in $[0, 1]$, and lower is better.
Let $t_i \in [0, T]$ for $i = 1, ..., l-1$ be a time point when a new better solution is found by a solver, $t_0 = 0$, and $t_l = T$ be the time limit.
The primal integral is defined as $P(T) = \sum_{i=1}^l p(t_{i-1}) \cdot (t_i - t_{i-1})$.
It takes a value in $[0, T]$, and lower is better.
$P(T)$ decreases if the same solution cost is achieved faster or a better solution is found with the same computational time.
When an instance is proved to be infeasible at time $t$, we use $p(t) = 0$, so $P(T)$ corresponds to the time to prove infeasibility.
For TSPTW, CVRP, and $1|| \sum w_i T_i$, we use the best-known solutions provided with the instances to compute the primal gap.
For other problems, we use the best solutions found by the solvers evaluated.

We show the average primal integral in Table~\ref{tab:primal-integral}.
Similar to Tables~\ref{tab:coverage} and \ref{tab:optimality-gap}, the primal integral of a DIDP solver is in bold if it is better than MIP and CP, and the best value is underlined.
CBFS, ACPS, and APPS outperform MIP and CP in six problem classes (TSPTW, CVRP, m-PDTSP, SALBP-1, MOSP, and graph-clear).
In addition to these problem classes, CABS achieves a better primal integral than CP in bin packing.
Comparing the DIDP solvers, CABS is the best in all problem classes except for OPTW, where CBFS and ACPS are better.
As mentioned above, CAASDy does not find feasible solutions for almost all unsolved instances, resulting in the worst primal integral in all problem classes.

In CVRP, while MIP solves more instances optimally, the DIDP solvers except for CAASDy achieve a better average primal integral since MIP fails to find primal bounds for large instances as mentioned.
In m-PDTSP, the DIDP solvers except for CAASDy have a lower primal integral than CP, which has the highest coverage.
In contrast, in OPTW, $1||\sum w_i T_i$, and talent scheduling, where the DIDP solvers solve more instances, CP has a better primal integral.

\subsection{Performance of DIDP Solvers and Problem Characteristics}

In CVRP, MIP solves more instances than the DIDP solvers even though the DyPDL model is similar to other routing problems (i.e., TSPTW, m-PDTSP, and OPTW) where the DIDP solvers are better.
Similarly, in bin packing, CP solves more instances even though the DyPDL model is similar to that of SALBP-1, where CABS is the best.
One common feature in the subset of the problem classes where DIDP is better than MIP or CP is a sequential dependency.
In TSPTW and OPTW, a solution is a sequence of visited customers, the time when a customer is visited depends on the partial sequence of customers visited before, and time window constraints restrict possible sequences.
Similarly, in m-PDTSP and SALBP-1, precedence constraints restrict possible sequences.
In the DyPDL models, these constraints restrict possible paths in the state transition graphs and thus reduce the number of generated states.
In contrast, in CVRP, as long as the capacity constraint is satisfied for each vehicle, customers can be visited in any order.
In the DyPDL model for bin packing, while item $i$ must be packed in the $i$-th or earlier bin and an arbitrary item is packed in a new bin by a forced transition, the remaining items can be packed in any order.
We conjecture that this difference, whether a sequential dependency exists or not, may be important for the performance of the DIDP solvers observed in our experiments.
However, sequential dependencies do not appear to be the only factor: DIDP also outperforms the other solvers in talent scheduling, MOSP, and Graph-Clear which do not exhibit such sequential dependencies.
Detailed analysis of model characteristics that affect the performance of DIDP solvers is left for future work.

\subsection{Evaluating the Importance of Dual Bound Functions} \label{sec:ablation-dual-bound}

As we described above, our DIDP solvers use the dual bound function defined in a DyPDL model as an admissible heuristic function, which is used for both search guidance and state pruning.
In $1||\sum w_i T_i$, MOSP, and graph-clear, we use a trivial dual bound function, which always returns 0.
Nevertheless, the DIDP solvers show better performance than MIP and CP.
This result suggests that representing these problems using state transition systems provides a fundamental advantage on these problems while raising the question of the impact of non-trivial dual bound functions on problem solving performance.
To investigate, we evaluate the performance of CABS with DyPDL models where the dual bound function is replaced with a function that always returns 0.
In other words, beam search keeps the best $b$ states according to the $g$-values and prunes a state $S$ if $g(S) \geq \overline{\gamma}$ in minimization, where $\overline{\gamma}$ is a primal bound.
Since the zero dual bound function is not valid for OPTW and MDKP, where the DyPDL models maximize the nonnegative total profit,
we use only TSPTW, CVRP, m-PDTSP, bin packing, SALBP-1, and talent scheduling.
We call this configuration CABS/0 and show results in Tables~\ref{tab:coverage}--\ref{tab:primal-integral}.
CABS/0 has a lower coverage than CABS in all problem classes except for TSPTW, where it achieves the same coverage.
In terms of the optimality gap and primal integral, CABS is better than CABS/0 in all problem classes, and the difference is particularly large in bin packing and SALBP-1.
The result confirms the importance of a non-trivial dual bound function for the current DIDP solvers.

\subsection{Comparison with Other State-Based Approaches} \label{sec:state-based-main}

We compare DIDP with domain-independent AI planning and Picat, a logic programming language that has an AI planning module.
For domain-independent AI planning, we formulate numeric planning models for TSPTW, CVRP, m-PDTSP, bin packing, and SALBP-1 based on the DyPDL models using PDDL 2.1.
We use NLM-CutPlan Orbit \cite{NLMCutPlan}, the winner of the optimal numeric track in the International Planning Competition (IPC) 2023,\footnote{\url{https://ipc2023-numeric.github.io/}} to solve the models.
For MOSP, we use a PDDL model for classical planning that was previously used in IPC from 2006 to 2008 with Ragnarok \cite{ragnarok}, the winner of the optimal classical track in IPC 2023.\footnote{\url{https://ipc2023-classical.github.io/}}
In these PDDL models, we are not able to model redundant information represented by state constraints, resource variables, dual bound functions, and forced transitions.
For Picat, we formulate models for TSPTW, CVRP, m-PDTSP, bin packing, SALBP-1, $1||\sum w_i T_i$, and talent scheduling using the AI planning module with the \textsf{best_plan_bb} predicate, which performs a branch-and-bound algorithm.
These models are equivalent to the DyPDL models except that they do not have resource variables.
For OPTW, MDKP, MOSP, and graph-clear, we use tabling, a feature to cache the evaluation results of predicates, and the models do not include resource variables and dual bound functions.
We provide more details for the PDDL and Picat models in \ref{sec:state-based-comparison}, and the implementations are available in our repository.\footnote{\url{https://github.com/Kurorororo/didp-models}}

For NLMCut-Plan and Ragnarok, a problem instance is translated to PDDL files by a Python script.
For Picat, a problem instance of CVRP, m-PDTSP, OPTW, SALBP-1, $1||\sum w_i T_i$, and talent scheduling is preprocessed and formatted by a Python script so that Picat can easily parse it.
We use GCC 12.3 for NLM-CutPlan and Ragnarok, IBM ILOG CPLEX 22.1.1 as a linear programming solver for Ragnarok, and Picat 3.6.

\begin{table}[tb]
    \caption[Coverage of MIP, CP, PDDL planners, Picat, and CABS.]{
        Coverage of MIP, CP, PDDL planners, Picat, and CABS.
        For PDDL, Ragnarok is used for MOSP, and NLM-CutPlan Orbit is used for the other problem classes.
        The coverage of a solver is in bold if it is higher than MIP and CP, and the higher of MIP and CP is in bold if there is no better solver.
        The highest coverage is underlined.
    }
    \small
    \centering
    \begin{tabular}{lrrrrr}
        \toprule
                                  & MIP                      & CP                        & PDDL & Picat         & CABS                      \\
        \midrule
        TSPTW (340)               & 224                      & 47                        & 61   & 210           & \underline{\textbf{259}}  \\
        CVRP (207)                & \underline{\textbf{28}}  & 0                         & 1    & 6             & 6                         \\
        m-PDTSP (1178)            & 940                      & \underline{\textbf{1049}} & 1031 & 804           & 1035                      \\
        OPTW (144)                & 16                       & 49                        & -    & 26            & \underline{\textbf{64}}   \\
        MDKP (276)                & \underline{\textbf{168}} & 6                         & -    & 3             & 5                         \\
        Bin Packing (1615)        & 1160                     & \underline{\textbf{1234}} & 18   & 895           & 1167                      \\
        SALBP-1 (2100)            & 1431                     & 1584                      & 871  & \textbf{1590} & \underline{\textbf{1802}} \\
        $1 || \sum w_i T_i$ (375) & 107                      & 150                       & -    & \textbf{199}  & \underline{\textbf{288}}  \\
        Talent Scheduling (1000)  & 0                        & 0                         & -    & \textbf{84}   & \underline{\textbf{239}}  \\
        MOSP (570)                & 241                      & 437                       & 193  & 162           & \underline{\textbf{527}}  \\
        Graph-Clear (135)         & 26                       & 4                         & -    & \textbf{45}   & \underline{\textbf{103}}  \\
        \bottomrule
    \end{tabular}
    \label{tab:planning}
\end{table}

Table~\ref{tab:planning} compares MIP, CP, the PDDL planners, Picat, and CABS.
Since the PDDL planners and Picat return only an optimal solution, we evaluate only coverage for them.
CABS has higher or equal coverage than the planners and Picat in all problem classes.
This result is not surprising since the planners and the AI planning module and tabling in Picat are not designed for combinatorial optimization.
It might be possible to improve the PDDL and Picat models so that they are more suited for these approaches.
Moreover, different PDDL planners might be better for combinatorial optimization.
However, our point is to show that the performance achieved by the DIDP solvers is not a trivial consequence of the state-based modeling approach, and DIDP is doing something that existing approaches are not able to easily do.

We also compare DIDP with ddo, a DD-based solver \cite{Gillard2020}, using TSPTW and talent scheduling, for which previous work developed models for ddo \cite{Gillard2021,Coppe2023,Coppe2024,Coppe2024b}.
For TSPTW, while we minimize the total travel time, which does not include the waiting time, the model for ddo minimizes the makespan, which is the time spent until returning to the depot.
Therefore, we adapt our DyPDL model to minimize the makespan:
when visiting customer $j$ from the current location $i$ with time $t$, we increase the cost by $\max\{ c_{ij}, a_j - t \}$ instead of $c_{ij}$.
To avoid confusion, in what follows, we call TSPTW to minimize the makespan TSPTW-M.

Since defining a merging operator required by ddo is a non-trivial task, we do not evaluate ddo for problem classes other than TSPTW-M and talent scheduling.
When two states are merged into one state, solutions for the original states must also be a solution for the merged state with the same or better cost.
In the DP model for TSPTW-M, three state variables, the set of unvisited customers $U$, the current location $i$, and the current time $t$, are used.
When two states with the sets of unvisited customers $U$ and $U'$ are merged, the set of customers that must be visited should be $U \cap U'$, but the set of customers that may be visited should be $U \cup U'$.
Thus, in the model for ddo, $U$ is replaced with two state variables, one representing the set of customers that must be visited and another representing the set of customers that may be visited, and these two variables have the same values in a non-merged state.
Similarly, the current locaton is represented by a set of locations that can be considered as the current location.
As in this example, defining a merging operator requires a significant change in the DP model.

\begin{table}[tb]
    \caption[Coverage and the average optimality gap of ddo and CABS in TSPTW-M and talent scheduling.]{
        Coverage and the average optimality gap of ddo and CABS in TSPTW-M and talent scheduling.
        For TSPTW-M, the optimality gap is not presented since ddo runs out of 8 GB memory in all unsolved instances and does not report intermediate solutions.
        For talent scheduling, the average optimality gap is computed from 976 instances where ddo does not reach the memory limit.
        A value of ddo or CABS is in bold if it is better than MIP and CP, and the better one of MIP and CP is in bold otherwise.
        The best value is underlined.
    }
    \small
    \centering
    \begin{tabular}{lrrrrrrrr}
        \toprule
                                 & \multicolumn{4}{c}{Coverage} & \multicolumn{4}{c}{Optimality Gap}                                                                                                             \\
        \cmidrule(lr){2-5} \cmidrule(lr){6-9}
                                 & MIP                          & CP                                 & ddo          & CABS                     & MIP    & CP     & ddo                         & CABS            \\
        \midrule
        TSPTW-M (340)            & 114                          & \underline{\textbf{331}}           & 213          & 260                      &        &        & -                           & -               \\
        Talent Scheduling (1000) & 0                            & 0                                  & \textbf{210} & \underline{\textbf{239}} & 0.8871 & 0.9513 & \underline{\textbf{0.1424}} & \textbf{0.1730} \\
        \bottomrule
    \end{tabular}
    \label{tab:ddo}
\end{table}

We evaluate ddo 2.0 using Rust 1.70.0 and present the result in Table~\ref{tab:ddo}.
We use the ddo models for TSPTW-M and talent scheduling obtained from the published repository of ddo.\footnote{\url{https://github.com/xgillard/ddo/tree/b2e68bfc085af7cc09ece38cc9c81acb0da6e965/ddo/examples}}
We also adapt the MIP and CP models for TSPTW to TSPTW-M and evaluate them.
While ddo returns the best solution and dual bound found within the time limit, it does not return intermediate solutions and bounds during solving.
Since we manage the memory limit using an external process, when ddo reaches the memory limit, it is killed without returning the best solution.
In TSPTW-M, ddo reaches the memory limit in all unsolved instances.
Therefore, we evaluate only coverage in TSPTW-M and present the average optimality gap computed from 976 out of 1000 talent scheduling instances where ddo does not reach the memory limit.
CABS is competitive with ddo:
it is better in TSPTW-M and talent scheduling in coverage, but ddo has a better average optimality gap in talent scheduling.
Note that when we adapt TSPTW to minimize makespan, CP performance increases considerably from the coverage of 47 (see Table~\ref{tab:coverage}) to 331 instances.
We conjecture that the improvement is a result of the strong back-propagation of the maximum objective function with specialized global constraints for scheduling \cite{Baptiste2001}.

\section{Conclusion and Future Work} \label{sec:conclusion}

We proposed domain-independent dynamic programming (DIDP), a novel model-based paradigm for combinatorial optimization based on dynamic programming (DP).
We introduced Dynamic Programming Description Language (DyPDL), a modeling formalism for DP, and YAML-DyPDL, a modeling language for DyPDL.
We developed seven DIDP solvers using heuristic search algorithms and experimentally showed that DIDP outperforms mixed-integer programming and constraint programming in a number of combinatorial optimization problem classes.
This result shows that DIDP is promising and complements existing model-based paradigms.

The significance of DIDP is that it is the first model-based paradigm designed for combinatorial optimization based on DP.
DIDP is based on two different fields, artificial intelligence (AI) and operations research (OR).
In particular, we focused on the state-based representations of problems, which are common in AI planning and DP but not previously exploited in a model-based paradigm for combinatorial optimization.
In AI planning, PDDL, the state-based modeling language, is commonly used, and some combinatorial optimization problems such as the minimization of open stacks problem were modeled in PDDL and used in International Planning Competitions.
However, PDDL and AI planners are not specifically designed for combinatorial optimization.
In OR, DP and state space search methods were used in problem-specific settings, but little work has developed a model-based paradigm based on DP.
DIDP bridges these gaps, benefitting from both AI and OR.
Since DIDP has a state-based modeling formalism similar to AI planning, we can apply heuristic search algorithms studied in AI to various combinatorial optimization problems.
Since DIDP follows the OR approach that allows a user to incorporate redundant information into optimization models, we can develop efficient models for application problems built upon problem-specific DP methods studied in OR.

DIDP opens up new research opportunities.
As shown in the experimental result, state-based paradigms such as DIDP and constraint-based paradigms such as mixed-integer programming and constraint programming are suited to different problem classes.
Even within state-based paradigms, DIDP and decision diagram-based solvers have different strengths.
Analyzing the characteristics of problems that make DIDP superior to others is an interesting direction.
DIDP also makes it possible to investigate better DP models, for example, by incorporating redundant information.

There is also a significant opportunity to improve DIDP.
One of the most important directions is to develop better heuristic functions for the heuristic search solvers.
The current solvers use a dual bound function provided in a DyPDL model for two roles: search guidance and state pruning.
While we demonstrated the importance of a dual bound function in Section~\ref{sec:ablation-dual-bound}, using it for search guidance is not necessarily justified as discussed in Section~\ref{sec:dypdl-heuristic-search};
we may improve the anytime behavior by using an inadmissible heuristic function for search guidance.
Disentangling search guidance from pruning and developing better functions for each role is one of our future plans.
In particular, we are considering developing methods to automatically compute heuristic functions from a DyPDL model as studied in AI planning.
In addition, decision diagrams (DDs) can be a source of heuristic functions;
in the existing DD-based solver \cite{Bergman2016,Gillard2020}, relaxed DDs, where multiple nodes are merged together to make the graph smaller, are used to compute a dual bound.
To obtain a relaxed DD, the DD-based solver requires a user to provide a merging operator.
Thus, for DIDP, there are two directions:
developing a domain-independent merging operator for DyPDL and extending DyPDL so that a user can declaratively incorporate a merging operator into a model.

In contrast to applying AI techniques to DIDP, using DIDP for AI planning is also possible.
In proving the undecidability of DyPDL, we showed that numeric planning tasks can be modeled in DyPDL.
As this result implies that we can automatically transform a numeric planning task into a DyPDL model, investigating more efficient DyPDL models for each planning domain is an interesting direction for future work.

\appendix

\section{Proof of Lemma~\ref{lem:open-not-empty}} \label{sec:proofs}

We first restate Lemma~\ref{lem:open-not-empty}.

\begin{quote}
    In Algorithm~\ref{alg:search}, suppose that a solution exists for the DyPDL model, and let $\hat{\gamma}$ be its cost.
    When reaching line~\ref{alg:search:while}, at least one of the following two conditions is satisfied:
    \begin{itemize}
        \item $\overline{\gamma} \leq \hat{\gamma}$.
        \item The open list $O$ contains a state $\hat{S}$ such that an $\hat{S}$-solution $\hat{\sigma}$ exists, and $\langle \sigma(\hat{S}); \hat{\sigma} \rangle$ is a solution for the model with $\mathsf{solution\_cost}(\langle \sigma(\hat{S}); \hat{\sigma} \rangle) \leq \hat{\gamma}$.
    \end{itemize}
\end{quote}

Recall that $O$ is the open list storing states to explore, $G$ is the set of generated and non-dominated states, $\sigma(S)$ is the current best path to state $S$, $\overline{\sigma}$ is the current best solution with the cost $\overline{\gamma}$, and $\hat{\gamma}$ is the cost of some solution for the DyPDL model.
The lemma claims that $\overline{\gamma} \leq \hat{\gamma}$ holds or $O$ contains a state $\hat{S}$ that leads to a solution with the cost at most $\hat{\gamma}$ at the beginning of each iteration of the while loop in Algorithm~\ref{alg:search}.

Once $\overline{\gamma} \leq \hat{\gamma}$ holds, $\overline{\gamma}$ never increases, so the lemma continues to hold.
We only consider $\overline{\gamma} > \hat{\gamma}$ in the current iteration and examine if the lemma will hold in the next iteration.

Now, we further specify our assumption.
Intuitively, we assume that among states in $G$ that lead to a solution with the cost at most $\hat{\gamma}$, the one $\hat{S}$ with the minimum number of transitions is included in $O$.
\begin{assumption} \label{asm:open-not-empty}
    When $\overline{\gamma} > \hat{\gamma}$, then $O$ contains a state $\hat{S}$ such that an $\hat{S}$-solution $\hat{\sigma}$ exists, $\langle \sigma(\hat{S}); \hat{\sigma} \rangle$ is a solution for the model with $\mathsf{solution\_cost}(\langle \sigma(\hat{S}); \hat{\sigma} \rangle) \leq \hat{\gamma}$, and $|\hat{\sigma}| \leq |\sigma'|$ for each $S'$-solution $\sigma'$ with $\mathsf{solution\_cost}(\langle \sigma(S'); \sigma' \rangle) \leq \hat{\gamma}$ for each $S' \in G$.
    Here, we denote the number of transitions in a solution $\sigma$ as $|\sigma|$.
\end{assumption}
At the beginning, $O$ and $G$ contain only one state $S^0$, so the assumption holds.
In line~\ref{alg:search:remove}, $S$ is removed from $O$.
We consider two cases, where $S$ is a base state or not, and prove that the claim will hold in the next iteration in either case.

\paragraph{Case 1: $S$ is a base state}
Since $\sigma(S)$ is a solution, if $\mathsf{solution\_cost}(\hat{\sigma}) \leq \hat{\gamma} < \overline{\gamma}$, then $\overline{\gamma}$ becomes $\mathsf{solution\_cost}(\hat{\sigma})$.
Now, $\overline{\gamma} \leq \hat{\gamma}$, and the assumption will hold in the next iteration.
Otherwise, we confirm that $\hat{S}$ is not removed from $O$ in the current iteration.
First, $\hat{S} \neq S$ because there exists a solution extending $\sigma(\hat{S})$ with the cost at most $\hat{\gamma}$ by the assumption while $\mathsf{solution\_cost}(\sigma(S)) \geq \overline{\gamma} > \hat{\gamma}$.
Second, $\hat{S}$ is not removed from $O$ in line~\ref{alg:search:prune-open} by the following reason.
By Lemma~\ref{lem:sigma}, $g(\hat{S}) \times \eta(\hat{S}) = w_{\sigma(\hat{S})}(S^0) \times \eta(\hat{S})$, where $\eta$ is the dual bound function.
By Definition~\ref{def:dual-bound}, $\eta(\hat{S}) \leq \mathsf{solution\_cost}(\hat{\sigma}, \hat{S})$.
Since $A$ is isotone, $g(\hat{S}) \times \eta(\hat{S}) \leq w_{\sigma(\hat{S})}(S^0) \times \mathsf{solution\_cost}(\hat{\sigma}, \hat{S}) \leq \hat{\gamma} < \overline{\gamma}$.

\paragraph{Case 2: $S$ is not a base state}
In this case, the successor states of $S$ are generated in lines~\ref{alg:search:gen}--\ref{alg:search:insert}.
When a successor state $S[\![\tau]\!]$ is inserted into $G$ by line~\ref{alg:search:insert}, it is also inserted into $O$.
We first show the following sublemma.

\begin{sublemma}
    When $O$ contains a state $\hat{S}$ specified by Assumption~\ref{asm:open-not-empty} at the beginning of an iteration of the for loop in lines~\ref{alg:search:gen}--\ref{alg:search:insert}, then the assumption continues to hold after finishing the current iteration.
\end{sublemma}
\begin{proof}
    If lines~\ref{alg:search:dominated}--\ref{alg:search:insert} are not reached, $O$ and $G$ are not changed, so the assumption continues to hold.
    Otherwise, the successor state $S[\![\tau]\!]$ is added to $G$ and $O$.
    We consider two cases, where $\hat{S}$ is not removed from $O$ by state dominance in line~\ref{alg:search:remove-dominated} and where it is removed.

    When $\hat{S}$ is not removed from $O$, if no $S[\![\tau]\!]$-solution $\sigma'$ with $\mathsf{solution\_cost}(\langle \sigma(S'); \sigma' \rangle) \leq \hat{\gamma}$ and $|\sigma'| < |\hat{\sigma}|$ exists, the assumption continues to hold.
    Otherwise, the current $\hat{S}$ no longer satisfies the assumption due to $S[\![\tau]\!] \in G$.
    In such a case, since $S[\![\tau]\!]$ is also added to $O$, by selecting $S[\![\tau]\!]$ as the new $\hat{S}$, we can ensure that the assumption holds in the next iteration.

    When $\hat{S}$ is removed from $O$ due to line~\ref{alg:search:remove-dominated}, then $\hat{S} \preceq_a S[\![\tau]\!]$ and $g(S) \times w_\tau(S) \leq g(\hat{S})$.
    By Definition~\ref{def:dominance}, for any $\hat{S}$-solution $\hat{\sigma}$, there exists an $S[\![\tau]\!]$-solution $\sigma^\tau$ such that $\mathsf{solution\_cost}(\sigma^\tau, S[\![\tau]\!]) \leq \mathsf{solution\_cost}(\hat{\sigma}, \hat{S})$ and $|\sigma^\tau| \leq |\hat{\sigma}|$.
    By isotonicity, we have $\mathsf{solution\_cost}(\langle \sigma(S); w_\tau(S); \sigma^\tau \rangle) = g(S) \times w_\tau(S) \times \mathsf{solution\_cost}(\sigma^\tau, S[\![\tau]\!]) \leq g(\hat{S}) \times \mathsf{solution\_cost}(\hat{\sigma}, \hat{S}) = \mathsf{solution\_cost}(\langle \sigma(\hat{S}); \hat{\sigma} \rangle) \leq \hat{\gamma}$.
    By Assumption~\ref{asm:open-not-empty}, $\mathsf{solution\_cost}(\langle \sigma(S); w_\tau(S); \sigma^\tau \rangle) \leq \mathsf{solution\_cost}(\langle \sigma(\hat{S}); \hat{\sigma} \rangle) \leq \hat{\gamma}$ and $|\sigma^\tau| \leq |\hat{\sigma}| \leq |\sigma'|$ for each $S'$-solution $\sigma'$ with $\mathsf{solution\_cost}(\langle \sigma(S'); \sigma' \rangle) \leq \hat{\gamma}$ for each $S' \in G$.
    Since $S[\![\tau]\!]$ is added to $O$, by selecting $S[\![\tau]\!]$ as the new $\hat{S}$, the assumption will hold in the next iteration.
\end{proof}

By this sublemma, if the removed state $S$ is not $\hat{S}$, Assumption~\ref{asm:open-not-empty} will hold in the next iteration.
Thus, we only consider the case where $S = \hat{S}$ in what follows.
We show that one of the successor states of $S$ that can be considered the new $\hat{S}$ is added to $O$.

Let $\hat{\sigma} = \langle \hat{\sigma}_1, ..., \hat{\sigma}_m \rangle$.
Then, $\hat{S}[\![\hat{\sigma}_1]\!]$ is a successor state of $\hat{S}$ with an $\hat{S}[\![\hat{\sigma}_1]\!]$-solution $\langle \hat{\sigma}_2, ..., \hat{\sigma}_m \rangle$ satisfying $\mathsf{solution\_cost}(\langle \sigma(\hat{S}); \hat{\sigma}_1; \langle \hat{\sigma}_2, ..., \hat{\sigma}_m \rangle \rangle) = \mathsf{solution\_cost}(\sigma(\hat{S}); \hat{\sigma}) \leq \hat{\gamma}$ and $|\langle \hat{\sigma}_2, ..., \hat{\sigma}_m \rangle| < |\hat{\sigma}|$.
Since $|\hat{\sigma}| \leq |\sigma'|$ for each $S'$-solution $\sigma'$ with $\mathsf{solution\_cost}(\langle \sigma(S'); \sigma' \rangle) \leq \hat{\gamma}$ for each $S' \in G$ by Assumption~\ref{asm:open-not-empty}, $|\langle \hat{\sigma}_2, ..., \hat{\sigma}_m \rangle| = |\hat{\sigma}| - 1 < |\sigma'|$ holds.
Thus, if $\hat{S}[\![\hat{\sigma}_1]\!]$ is added to $O$ in line~\ref{alg:search:insert}, the assumption will hold.

If there exists a forced transition $\tau^f \neq \hat{\sigma}_1$, by Definition~\ref{def:forced}, then there exists an $\hat{S}[\![\tau^f]\!]$-solution $\sigma^f$ such that $\mathsf{solution\_cost}(\langle \sigma(\hat{S}); \tau^f; \sigma^f \rangle) \leq \mathsf{solution\_cost}(\langle \sigma(\hat{S}); \hat{\sigma} \rangle) \leq \hat{\gamma}$ and $|\sigma^f| \leq |\hat{\sigma}| - 1$ .
Thus, we can continue the discussion by considering $\hat{S}[\![\tau^f]\!]$ instead of $\hat{S}[\![\hat{\sigma}_1]\!]$.

We examine lines~\ref{alg:search:dominating} and \ref{alg:search:check-bound} that might prevent reaching line~\ref{alg:search:insert} to insert $\hat{S}[\![\hat{\sigma}_1]\!]$ (or $\hat{S}[\![\tau^f]\!]$) into $O$.

\subparagraph{State dominance check in line~\ref{alg:search:dominating}}
For a state $S' \in G$ to satisfy $S' \preceq_a \hat{S}[\![\hat{\sigma}_1]\!]$, there must exist an $S'$-solution $\sigma'$ that is not worse and not shorter than the $S[\![\hat{\sigma}_1]\!]$-solution $\langle \hat{\sigma}_2, ..., \hat{\sigma}_m \rangle$.
In other words, $\mathsf{solution\_cost}a(\langle \sigma(S'); \sigma' \rangle) \leq \mathsf{solution\_cost}(\langle \sigma(\hat{S}); \hat{\sigma} \rangle) \leq \hat{\gamma}$ and $|\sigma'| \leq |\hat{\sigma}| - 1 < |\hat{\sigma}|$.
However, such an $S'$-solution does not exist by Assumption~\ref{asm:open-not-empty}.

\subparagraph{Bound check in line~\ref{alg:search:check-bound}}
Since $\eta(\hat{S}[\![\hat{\sigma}_1]\!]) \leq \mathsf{solution\_cost}(\langle \hat{\sigma}_2, ..., \hat{\sigma}_m \rangle, \hat{S}[\![\hat{\sigma}_1]\!])$, we have $g(\hat{S}) \times w_{\hat{\sigma}_1}(\hat{S}) \times \eta(\hat{S}[\![\hat{\sigma}_1]\!]) \leq g(\hat{S}) \times w_{\hat{\sigma}_1}(\hat{S}) \times \mathsf{solution\_cost}(\langle \hat{\sigma}_2, ..., \hat{\sigma}_m \rangle, \hat{S}[\![\hat{\sigma}_1]\!]) = \mathsf{solution\_cost}(\langle \sigma(\hat{S}); \hat{\sigma}\rangle) \leq \hat{\gamma} < \overline{\gamma}$ by isotonicity.
Therefore, line~\ref{alg:search:insert} is reached.

\section{Additional Problem and Model Definitions} \label{sec:cp-models}

We present problem definitions and DyPDL models that are not covered in Section~\ref{sec:models} and new CP models used in the experimental evaluation.

\subsection{Multi-Dimensional Knapsack Problem (MDKP)} \label{sec:mdkp}

The multi-dimensional knapsack problem (MDKP) \cite{Martello1990,Kellerer2004} is a generalization of the 0-1 knapsack problem.
In this problem, each item $i \in \{ 0, ..., n-1 \}$ has the integer profit $p_i \geq 0$ and $m$-dimensional nonnegative weights $(w_{i,0}, ..., w_{i,m-1})$, and the knapsack has the $m$-dimensional capacities $(q_{0}, ..., q_{m-1})$.
In each dimension, the total weight of items included in the knapsack must not exceed the capacity.
The objective is to maximize the total profit.
MDKP is strongly NP-hard \cite{Garey1990,Cacchiani2022}

\subsubsection{DyPDL Model for MDKP} \label{sec:mdkp-dp}

In our DyPDL model, we decide whether to include an item one by one.
An element variable $i$ represents the index of the item considered currently, and a numeric variable $r_j$ represents the remaining space in the $j$-th dimension.
We can use the total profit of the remaining items as a dual bound function.
In addition, we consider an upper bound similar to that of OPTW by ignoring dimensions other than $j$.
Let $e_{kj} = p_k / w_{kj}$ be the efficiency of item $k$ in dimension $j$.
Then, $\lfloor r_j \max_{k=i,...,n-1} e_{kj} \rfloor$ is an upper bound on the cost of a $(i, r_j)$-solution.
If $w_{kj} = 0$, we define $e_{kj} = \sum_{k = i, ..., n-1} p_k$, i.e., the maximum additional profit achieved from $(i, r_j)$.
In such a case, $\max\{ r_j, 1 \} \cdot \max_{k=1,...,n-1} e_{ij}$ is still a valid upper bound.
\begin{align}
     & \text{compute } V(0, q_0, ..., q_{m-1})                                                                                                                                                 \\
     & V(i, r_0, ..., r_{m-1}) = \begin{cases}
                                     0                                                         & \text{if } i = n          \\
                                     \max\left\{ \begin{array}{l}
                        p_i + V(i + 1, r_0 - w_{i0}, ..., r_{m-1} - w_{i,m-1}) \\
                        V(i + 1, r_0, ..., r_{m-1})
                    \end{array} \right. & \text{else if } \forall j \in M, w_{ij} \leq r_{ij} \\
                                     V(i + 1, r_0, ..., r_{m-1})                               & \text{else}
                                 \end{cases}                                                           \\
     & V(i, r_0, ..., r_{m-1}) \leq \min\left\{ \sum_{k= i,...,n-1} p_k, \min\limits_{j \in M} \left\lfloor \max\{ r_j, 1 \} \cdot \max\limits_{k = i, ..., n-1} e_{kj} \right\rfloor \right\}
\end{align}
where $M = \{ 0, ..., m-1 \}$.
Similar to OPTW, this model is monoidal and the isotonicity is satisfied, but it is not cost-algebraic, so the first solution found by CAASDy may not be optimal (Theorem~\ref{thm:a*}).

\subsubsection{CP Model for MDKP} \label{sec:mdkp-cp}

We use the $\mathsf{Pack}$ global constraint \cite{Shaw2004} and consider packing all items into two bins;
one represents the knapsack, and the other represents items not selected.
We introduce a binary variable $x_i$ representing the bin where item $i$ is packed ($x_i = 0$ represents the item is in the knapsack).
We define an integer variable $y_{j,0}$ representing the total weight of the items in the knapsack in dimension $j$ of the knapsack and $y_{j,1}$ representing the total weight of the items not selected.
\begin{align}
    \max         & \sum_{i \in N} p_i (1 - x_i)                                                                                \\
    \text{s.t. } & \mathsf{Pack}(\{ y_{j,0}, y_{j,1} \}, \{ x_i \mid i \in N \}, \{ w_{ij} \mid i \in N \}) & j = 0, ..., m-1  \\
                 & y_{j,0} \leq q_j                                                                         & j = 0, ..., m-1  \\
                 & y_{j,0}, y_{j,1} \in \mathbb{Z}_0^+                                                      & j = 0, ..., m-1  \\
                 & x_i \in \{ 0, 1 \}                                                                       & \forall i \in N.
\end{align}

\subsection{Single Machine Total Weighted Tardiness ($1||\sum w_i T_i$)} \label{sec:wt}

In single machine scheduling to minimize the total weighted tardiness ($1||\sum w_i T_i$) \cite{Emmons1969}, a set of jobs $N$ is given, and each job $i \in N$ has a processing time $p_i$,  a deadline $d_i$, and an weight $w_i$, all of which are nonnegative.
The objective is to schedule all jobs on a machine while minimizing the sum of the weighted tardiness, $\sum_{i \in N} w_i \max\{0, C_i - d_i \}$ where $C_i$ is the completion time of job $i$.
This problem is strongly NP-hard \cite{Lenstra1977}.

\subsubsection{DyPDL Model for $1||\sum w_i T_i$} \label{sec:wt-dp}

We formulate a DyPDL model based on an existing DP model \cite{Held1962,AbdulRazaq1990}, where one job is scheduled at each step.
Let $F$ be a set variable representing the set of scheduled jobs.
A numeric expression $T(i, F) = \max\{0, \sum_{j \in F} p_j + p_i - d_i \}$ represents the tardiness of $i$ when it is scheduled after $F$.
We introduce a set $P_i$, representing the set of jobs that can be scheduled before $i$ without losing optimality.
While $P_i$ is redundant information not defined in the problem, it can be extracted in preprocessing using precedence theorems \cite{Emmons1969}.
\begin{align}
     & \text{compute } V(\emptyset)                                                                                  \\
     & V(F) = \begin{cases}
                  0                                                                                   & \text{if } F = N \\
                  \min\limits_{i \in N \setminus F : P_i \subseteq F} w_i T(i, F) + V(F \cup \{ i \}) & \text{else}
              \end{cases} \\
     & V(F) \geq 0.
\end{align}
This model is cost-algebraic with a cost algebra $\langle \mathbb{Q}_0^+, +, 0 \rangle$ since $w_i T(i, F) \geq 0$.
Since the base cost is always zero, the first solution found by CAASDy is optimal.

\subsubsection{CP Model for $1||\sum w_i T_i$} \label{sec:wt-cp}

We use an interval variable $x_i$ with the duration $p_i$ and the time within $[0, \sum_{j \in N} p_j]$, representing the interval of time when job $i$ is processed.
\begin{align}
    \min         & \sum_{i \in N} w_i \max\{\mathsf{EndOf}(x_i) - d_i, 0 \}                       &                                    \\
    \text{s.t. } & \mathsf{NoOverlap}(\pi)                                                                                             \\
                 & \mathsf{Before}(\pi, x_i, x_j)                                                 & \forall j \in N, \forall i \in P_j \\
                 & x_i : \mathsf{intervalVar}\left(p_i, \left[0, \sum_{j \in N} p_j\right]\right) & \forall i \in N                    \\
                 & \pi : \mathsf{sequenceVar}(\{ x_0, ..., x_{n-1} \}).
\end{align}

\subsection{DyPDL Model for Talent Scheduling}

In talent scheduling, we are given a set of scenes $N = \{ 0, ..., n-1 \}$, a set of actors $A = \{ 0, ..., m -1 \}$, the set of actors $A_s \subseteq A$ playing in scene $s$, and the set of scenes $N_a \subseteq N$ where actor $a$ plays.
In addition, $d_s$ is the duration of a scene $s$, and $c_a$ is the cost of an actor $a$ per day.

\subsubsection{DyPDL Model for Talent Scheduling} \label{sec:talent-dp}

We use a set variable $Q$ representing the set of scenes that are not yet shot and a set expression $L(Q) = \bigcup_{s \in Q} A_s \cap \bigcup_{s \in N \setminus Q} A_s$ to represent the set of actors on location after shooting $N \setminus Q$.
If $A_s = L(Q)$, then $s$ should be immediately shot because all actors are already on location: a forced transition.
For the dual bound function, we underestimate the cost to shoot $s$ by $b_s = \sum_{a \in A_s} c_a$.

If there exist two scenes $s_1$ and $s_2$ in $Q$ such that $A_{s_1} \subseteq A_{s_2}$ and $A_{s_2} \subseteq \bigcup_{s \in N \setminus Q} A_s \cup A_{s_1}$, it is known that scheduling $s_2$ before $s_1$ is always better, denoted by $s_2 \preceq s_1$.
Since two scenes with the same set of actors are merged into a single scene in preprocessing without losing optimality, we can assume that all $A_s$ are different.
With this assumption, the relationship is a partial order:
it is reflexive because $A_{s_1} \subseteq A_{s_1}$ and $A_{s_1} \subseteq \bigcup_{s \in N \setminus Q} A_s \cup A_{s_1}$;
it is antisymmetric because if $s_1 \preceq s_2$ and $s_2 \preceq s_1$, then $A_{s_1} \subseteq A_{s_2}$ and $A_{s_2} \subseteq A_{s_1}$, which imply $s_1 = s_2$;
it is transitive because if $s_2 \preceq s_1$ and $s_3 \preceq s_2$, then $A_{s_1} \subseteq A_{s_2} \subseteq A_{s_3}$ and $A_{s_3} \subseteq \bigcup_{s \in N \setminus Q} A_s \cup A_{s_2} \subseteq \bigcup_{s \in N \setminus Q} A_s \cup A_{s_1}$, which imply $s_3 \preceq s_1$.
Therefore, the set of candidate scenes to shoot next, $R(Q) = \{ s_1 \in Q \mid \not \exists s_2 \in Q \setminus \{ s_1 \}, s_2 \preceq s_1 \}$, is not empty.

\begin{align}
     & V(Q) = \begin{cases}
                  0                                                                                    & \text{if } Q = \emptyset                    \\
                  b_s + V(Q \setminus \{ s \})                                                         & \text{else if } \exists s \in Q, A_s = L(Q) \\
                  \min\limits_{s \in R(Q)} d_s \sum_{a \in A_s \cup L(Q)} c_a + V(Q \setminus \{ s \}) & \text{else if } Q \neq \emptyset
              \end{cases} \\
     & V(Q) \geq \sum_{s \in Q} b_s.
\end{align}

\subsubsection{CP Model for Talent Scheduling} \label{sec:talent-cp}

We extend the model used by \citeauthor{Chu2015}~\cite{Chu2015}, which was originally implemented with MiniZinc \cite{Nethercote2007}, with the $\mathsf{AllDifferent}$ global constraint.
While $\mathsf{AllDifferent}$ is redundant in the model, it slightly improved the performance in our preliminary experiment.
Let $x_i$ be a variable representing the $i$-th scene in the schedule, $b_{si}$ be a variable representing if scene $s$ is shot before the $i$-th scene, $o_{ai}$ be a variable representing if any scene in $N_a$ is shot by the $i$-th scene, and $f_{ai}$ be a variable representing if all scenes in $N_a$ finish before the $i$-th scene.
The CP model is
\begin{align}
    \min         & \sum_{i \in N} d_{x_i} \sum_{a \in A} c_a o_{ai} (1 - f_{ai})                                                                         \\
    \text{s.t. } & \mathsf{AllDifferent}(\{x_i \mid i \in N \})                                                                                          \\
                 & b_{s0} = 0                                                                       & \forall s \in N                                    \\
                 & b_{si} = b_{s,i-1} + \mathbbm{1}(x_{i-1} = s)                                    & \forall i \in N \setminus \{ 0 \}, \forall s \in N \\
                 & b_{si} = 1 \rightarrow x_i \neq s                                                & \forall i \in N \setminus \{ 0 \}, \forall s \in N \\
                 & o_{a0} = \mathbbm{1}\left(\bigvee_{s \in N_a} x_0 = s \right)                    & \forall a \in A                                    \\
                 & o_{ai} = \mathbbm{1}\left(o_{a,i-1} = 1 \lor \bigvee_{s \in N_a} x_i = s \right) & \forall i \in N \setminus \{ 0 \}, \forall a \in A \\
                 & f_{ai} = \prod_{s \in N_a} b_{si}                                                & \forall i \in N, a \in A                           \\
                 & x_i \in N                                                                        & \forall i \in N                                    \\
                 & b_{si}, f_{si} \in \{ 0, 1 \}                                                    & \forall s, i \in N.
\end{align}

\subsection{DyPDL Model for Minimization of Open Stacks Problem (MOSP)} \label{sec:mosp-dp}

In the minimization of open stacks problem (MOSP) \cite{Yuen1995}, customers $C$ and products $P$ are given, and each customer $c$ orders products $P_c \subseteq P$.
A solution is a sequence in which products are produced.
When producing product $i$, a stack for customer $c$ with $i \in P_c$ is opened, and it is closed when all of $P_c$ are produced.
The objective is to minimize the maximum number of open stacks at a time.
MOSP is strongly NP-hard \cite{Linhares2002}.

For MOSP, customer search is a state-of-the-art exact method \cite{Chu2009}.
It searches for an order of customers to close stacks, from which the order of products is determined;
for each customer $c$, all products ordered by $c$ and not yet produced are consecutively produced in an arbitrary order.
We formulate customer search as a DyPDL model.
A set variable $R$ represents customers whose stacks are not closed, and $O$ represents customers whose stacks have been opened.
Let $N_c = \{ c' \in C \mid P_c \cap P_{c'} \neq \emptyset \}$ be the set of customers that order at least one of the same products as customer $c$.
When producing items for customer $c$, we need to open stacks for customers in $N_c \setminus O$, and stacks for customers in $O \cap R$ remain open.
\begin{align}
     & \text{compute } V(C, \emptyset)                                                                                                                         \\
     & V(R, O) = \begin{cases}
                     0                                                                                                                  & \text{if } R = \emptyset \\
                     \min\limits_{c \in R} \max\left\{ |(O \cap R) \cup (N_c \setminus O)|, V(R \setminus \{ c \}, O \cup N_c) \right\} & \text{else}
                 \end{cases} \\
     & V(R, O) \geq 0.
\end{align}
Similar to the DyPDL model for graph-clear in Section~\ref{sec:graph-clear}, this model is cost-algebraic, and the base cost is zero, so the first solution found by CAASDy is optimal.

\subsection{CP Model for Orienteering Problem with Time Windows (OPTW)} \label{sec:optw-cp}

In the orienteering problem with time windows (OPTW), we are given a set of customers $N = \{ 0, ..., n-1 \}$, the travel time $c_{ij}$ from customer $i$ to $j$, and the profit $p_i$ of customer $i$.
We define an optional interval variable $x_i$ that represents visiting customer $i$ in $[a_i, b_i]$.
We also introduce an interval variable $x_n$ that represents returning to the depot ($0$) and define $c_{in} = c_{i0}$ and $c_{ni} = c_{0i}$ for each $i \in N$.
We use a sequence variable $\pi$ to sequence the interval variables.
\begin{align}
    \max         & \sum_{i \in N \setminus \{ 0 \}} p_i \mathsf{Pres}(x_i)                                                                                                          & \label{eqn:optw-cp:obj}           \\
    \text{s.t. } & \mathsf{NoOverlap}(\pi, \{ c_{ij} \mid (i, j) \in (N \cup \{ n \}) \times (N \cup \{ n \}) \})                                    \label{eqn:optw-cp:no-overlap}                                     \\
                 & \mathsf{First}(\pi, x_0)                                                                                                           \label{eqn:optw-cp:first}                                         \\
                 & \mathsf{Last}(\pi, x_n)                                                                                                           \label{eqn:optw-cp:last}                                           \\
                 & x_i : \mathsf{optIntervalVar}(0, [a_i, b_i] )                                                                                                                    & \forall i \in N \setminus \{ 0 \} \\
                 & x_0 : \mathsf{intervalVar}(0, [0, 0])                                                                                                                                                                \\
                 & x_n : \mathsf{intervalVar}(0, [0, b_0])                                                                                                                                                              \\
                 & \pi : \mathsf{sequenceVar}(\{ x_0, ..., x_n \}).
\end{align}
Objective~\eqref{eqn:optw-cp:obj} maximizes the total profit, where $\mathsf{Pres}(x_i) = 1$ if the optional interval variable $x_i$ is present.
Constraint~\eqref{eqn:optw-cp:no-overlap} ensures that if $x_j$ is present in $\pi$ after $x_i$, the distance between them is at least $c_{ij}$.
Constraints~\eqref{eqn:optw-cp:first} and \eqref{eqn:optw-cp:last} ensure that the tour starts from and returns to the depot.

\subsection{CP Model for Bin Packing} \label{sec:bpp-cp}

In the bin packing problem, we are given a set of items $N = \{ 0, ..., n-1 \}$, the weight $w_i$ of each item $i \in N$, and the capacity of a bin $q$.
We compute the upper bound $\bar{m}$ on the number of bins using the first-fit decreasing heuristic and use $M = \{ 0, ..., \hat{m}-1 \}$.
We use $\mathsf{Pack}$ and ensure that item $i$ is packed in the $i$-th or an earlier bin.
\begin{align}
    \min         & \max_{i \in N} x_i + 1                                                                                   \\
    \text{s.t. } & \mathsf{Pack}(\{ y_j \mid j \in M \}, \{ x_i \mid i \in N \}, \{ w_i \mid i \in N \})                    \\
                 & 0 \leq y_j \leq q                                                                     & \forall j \in M  \\
                 & 0 \leq x_i \leq i                                                                     & \forall i \in N  \\
                 & y_j \in \mathbb{Z}                                                                    & \forall j \in M  \\
                 & x_i \in \mathbb{Z}                                                                    & \forall i \in N.
\end{align}

\subsection{CP Model for Simple Assembly Line Balancing Problem (SALBP-1)} \label{sec:salbp-1-cp}

In SALBP-1, in addition to tasks and the capacity of a station, we are given $P_i$, the set of the predecessors of task $i$.
We implement the CP model proposed by \citeauthor{Bukchin2018}~\cite{Bukchin2018} with the addition of $\mathsf{Pack}$.
For an upper bound on the number of stations, instead of using a heuristic to compute it, we use $\bar{m} = \min \{ n, 2 \lceil \sum_{i \in N} w_i / q \rceil \}$ following the MIP model \cite{Ritt2018}.
Let $m$ be the number of stations, $x_i$ be the index of the station of task $i$, and $y_j$ be the sum of the weights of tasks scheduled in station $j$.
The set of all direct and indirect predecessors of task $i$ is $\tilde{P}_i = \{ j \in N \mid j \in P_i \lor \exists k \in \tilde{P}_j, j \in \tilde{P}_k \}$.
The set of all direct and indirect successors of task $i$ is $\tilde{S}_i = \{ j \in N \mid i \in P_j \lor \exists k \in \tilde{S}_i, j \in \tilde{S}_k \}$.
Thus, $e_i = \left\lceil \frac{w_i + \sum_{k \in \tilde{P}_i} t_k}{q} \right\rceil$ is a lower bound on the number of stations required to schedule task $i$, $l_i = \left\lfloor \frac{w_i - 1 + \sum_{k \in \tilde{S}_i} t_k}{q} \right\rfloor$ is a lower bound on the number of stations between the station of task $i$ and the last station, and $d_{ij} = \left\lfloor \frac{w_i + t_j - 1 + \sum_{k \in \tilde{S}_i \cap \tilde{P}_j} t_k}{q} \right\rfloor$ is a lower bound on the number of stations between the stations of tasks $i$ and $j$.
\begin{align}
    \min\        & m                                                                                                                                                                                                                                                                              \\
    \text{s.t. } & \mathsf{Pack}(\{ y_j \mid j \in M \}, \{ x_i \mid i \in N \}, \{ w_i \mid i \in N \}) \label{eqn:salbp-1-cp:pack}                                                                                                                                                              \\
                 & 0 \leq y_j \leq q                                                                                                 & \forall j \in M  \label{eqn:salbp-1-cp:capacity}                                                                                                           \\
                 & e_i - 1 \leq x_i \leq m - 1 - l_i                                                                                 & \forall i \in N \label{eqn:salbp-1-cp:bounds}                                                                                                              \\
                 & x_i + d_{ij} \leq x_j                                                                                             & \forall j \in N, \forall i \in \tilde{P}_j, \not\exists k \in \tilde{S}_i \cap \tilde{P}_j : d_{ij} \leq d_{ik} + d_{kj} \label{eqn:salbp-1-cp:precedence} \\
                 & m \in \mathbb{Z}                                                                                                  &                                                                                                                                                            \\
                 & y_j \in \mathbb{Z}                                                                                                & \forall j \in M                                                                                                                                            \\
                 & x_i \in \mathbb{Z}                                                                                                & \forall i \in N.
\end{align}
Constraint~\eqref{eqn:salbp-1-cp:bounds} states the lower and upper bounds on the index of the station of $i$.
Constraint~\eqref{eqn:salbp-1-cp:precedence} is an enhanced version of the precedence constraint using $d_{ij}$.

\section{Modeling in AI Planning} \label{sec:state-based-comparison}

In Section~\ref{sec:state-based-main}, we compare DIDP with AI planning approaches:
numeric planning using planning domain definition languages (PDDL) \cite{McDermott2000,FoxL03} and Picat, a logic programming language that has an AI planning module.
We explain how we formulate models for these approaches.

\subsection{PDDL}

We model TSPTW, CVRP, m-PDTSP, bin packing, and SALBP-1 as linear numeric planning tasks \cite{Hoffmann03}, where preconditions and effects of actions are represented by linear formulas of numeric state variables.
In these models, the objective is to minimize the sum of nonnegative action costs, which is a standard in optimal numeric planning \cite{Scala2017,Piacentini2018,Piacentini2018b,ScalaHTR20,Leofante2020,KuroiwaSPCB22,KuroiwaSB2022,Shleyfman2023,Kuroiwa2023Numeric}.
We use PDDL 2.1 \cite{FoxL03} to formulate the models and NLM-CutPlan Orbit \cite{NLMCutPlan}, the winner of the optimal numeric track of the International Planning Competition (IPC) 2023,\footnote{\url{https://ipc2023-numeric.github.io/}} to solve the models.
The PDDL models are adaptations of the DyPDL models presented in Section~\ref{sec:models}, where a numeric or element variable in DyPDL becomes a numeric variable in PDDL (except for the element variable representing the current location in the routing problems as explained in the next paragraph), a set variable is represented by a predicate and a set of objects, the target state becomes the initial state, base cases become the goal conditions, and a transition becomes an action.
However, we are unable to model dominance between states and dual bound functions in PDDL.
In addition, we cannot differentiate forced transitions and other transitions.
We explain other differences between the PDDL models and the DyPDL models in the following paragraphs.

In the DyPDL models of the routing problems (TSPTW, CVRP, and m-PDTSP), each transition visits one customer and increases the cost by the travel time from the current location to the customer.
Since a goal state in PDDL is not associated with a cost unlike a base case in DyPDL, we also define an action to return to the depot, which increases the cost by the travel time to the depot.
While the travel time depends on the current location, for NLM-CutPlan Orbit, the cost of an action must be a nonnegative constant independent of a state, which is a standard in admissible heuristic functions for numeric planning \cite{Scala2017,Piacentini2018b,ScalaHTR20,KuroiwaSPCB22,KuroiwaSB2022,Kuroiwa2023Numeric}.
Thus, in the PDDL models, we define one action with two parameters, the current location (\textsf{?from}) and the destination (\textsf{?to}), so that the cost of each grounded action becomes a state-independent constant \textsf{(c ?from ?to)} corresponding to the travel time.
We define a predicate \textsf{(visited ?customer)} representing if a customer is visited and \textsf{(location ?customer)} representing the current location.
In each action, we use \textsf{(location ?from)} and \textsf{(not (visited ?to))} as preconditions and \textsf{(not (location ?from))}, \textsf{(location ?to)}, and \textsf{(visited ?to)} as effects.

In the DyPDL models of TSPTW, each transition updates the current time $t$ to $\max\{ t + c_{ij}, a_j \}$, where $c_{ij}$ is the travel time and $a_j$ is the beginning of the time window at the destination.
While this effect can be written as \textsf{(assign (t) (max (+ (t) (c ?from ?to)) (a ?to)))} in PDDL, to represent effects as linear formulas, we introduce two actions:
one with a precondition \textsf{($>=$ (+ (t) (c ?from ?to)) (a ?to))} and an effect \textsf{(increase (t) (c ?from ?to))}, corresponding to $t \leftarrow t + c_{ij}$ if $t + c_{ij} \geq a_j$, and another with a precondition \textsf{($<$ (+ (t) (c ?from ?to)) (a ?to))} and an effect \textsf{(assign (t) (a ?to))}, corresponding to $t \leftarrow a_j$ if $t + c_{ij} < a_j$.

In the DyPDL models of TSPTW and CVRP, we have redundant state constraints.
While a state constraint could be modeled by introducing it as a precondition of each action, we do not use the state constraints in the PDDL models of TSPTW and CVRP because efficiently modeling them is non-trivial:
straightforward approaches result in an exponential number of actions.
For TSPTW, the state constraint checks if all unvisited customers can be visited by the deadline, represented as $\forall j \in U, t + c^*_{ij} \leq b_j$, where $U$ is the set of the unvisited customers, $c^*_{ij}$ is the shortest travel time from the current location to customer $j$, and $b_j$ is the deadline.
One possible way to model this constraint is to define a disjunctive precondition \textsf{(or (visited ?j) ($<$= (+ (t) (cstar ?from ?j)) (b ?j)))} for each customer \textsf{?j}, where \textsf{(t)} is a numeric variable corresponding to $t$, \textsf{(cstar ?from ?j)} is a numeric constant corresponding to $c^*_{ij}$, and \textsf{(b ?j)} is a numeric constant corresponding to $b_j$.
However, the heuristic function used by NLM-CutPlan Orbit does not support disjunctive preconditions, and NLM-CutPlan Orbit compiles an action with disjunctive preconditions into a set of actions with different combinations of the preconditions.\footnote{This approach is inherited from Fast Downward \cite{Helmert06}, the standard classical planning framework on which NLM-CutPlan Orbit is based.}
In our case, each action has one of the two preconditions, \textsf{(visited ?j)} or \textsf{($<=$ (+ (t) (cstar ?from ?j)) (b ?j))}, resulting in $2^n$ actions in total, where $n$ is the number of customers.
In CVRP, the state constraint takes the sum of demands over all unvisited customers.
To model this computation independently of a state, we need to define an action for each possible set of unvisited customers, resulting in $2^n$ actions in total.

In the DyPDL models of bin packing and SALBP-1, each transition packs an item in the current bin (schedules a task in the current station for SALBP-1) or opens a new bin.
When opening a new bin, the transition checks if no item can be packed in the current bin as a precondition, which is unnecessary but useful to exclude suboptimal solutions.
However, for similar reasons to the state constraint in TSPTW, we do not model this precondition in the PDDL models.
We could model this condition by defining \textsf{(or (packed ?j) ($>$ (w ?j) (r)))} for each item \textsf{?j}, where \textsf{(packed ?j)} represents if \textsf{?j} is already packed, \textsf{(w ?j)} represents the weight of \textsf{?j}, and \textsf{(r)} is the remaining capacity.
However, as discussed above, NLM-CutPlan Orbit would generate an exponential number of actions with this condition.

In addition to the above problem classes, we also use MOSP:
it was used as a benchmark domain in the classical planning tracks of the International Planning Competitions from 2006 to 2014.
This PDDL formulation is different from our DyPDL model.
To solve the model, we use Ragnarok \cite{ragnarok}, the winner of the optimal classical track of IPC 2023.\footnote{\url{https://ipc2023-classical.github.io/}}

We do not use other problem classes since their DyPDL models do not minimize the sum of the state-independent and nonnegative action costs.
In $1||\sum w_iT_i$  and talent scheduling, since the cost of each transition depends on a set variable, we need an exponential number of actions to make it state-independent.
In OPTW and MDKP, the objective is to maximize the sum of the nonnegative profits.
In graph-clear, the objective is to minimize the maximum value of state-dependent weights associated with transitions.

\subsection{Picat}

Picat is a logic programming language, in which DP can be used with tabling, a feature to store and reuse the evaluation results of predicates, without implementing a DP algorithm.
Picat provides an AI planning module based on tabling, where a state, goal conditions, and actions can be programmatically described by expressions in Picat.
While the cost of a plan is still restricted to the sum of the nonnegative action costs, each action cost can be state-dependent.
In addition, an admissible heuristic function can be defined and used by a solving algorithm.
Thus, we can define a dual bound function as an admissible heuristic function in the AI planning module.
However, we cannot model dominance between states.
Using the AI planning module, we formulate models for TSPTW, CVRP, m-PDTSP, bin packing, SALBP-1, $1||\sum w_i T_i$, and talent scheduling, which are the same as the DyPDL models except that they do not define dominance.
To solve the formulated models, we use the \textsf{best_plan_bb} predicate, which performs a branch-and-bound algorithm using the heuristic function.
For OPTW, MDKP, MOSP, and graph-clear, we do not use the AI planning module due to the objective structure of their DyPDL models.
We define DP models for these problem classes, which are the same as the DyPDL models except that they do not define dominance and dual bound functions, using tabling without the AI planning module.



\section*{Acknowledgement}

\noindent
This work was supported by the Natural Sciences and Engineering Research Council of Canada.
This reseearch was enabled in part by support provided by Compute Ontario and the Digital Research Alliance of Canada (\url{alliancecan.ca}).

\bibliographystyle{elsarticle-num-names}
\bibliography{references}

@article{Uchoa2017,
  author    = {Eduardo Uchoa and Diego Pecin and Artur Pessoa and Marcus Poggi and Thibaut Vidal and Anand Subramanian},
  doi       = {10.1016/j.ejor.2016.08.012},
  number    = {3},
  journal   = {Eur. J. Oper. Res.},
  pages     = {845--858},
  publisher = {Elsevier B.V.},
  title     = {New Benchmark Instances for the Capacitated Vehicle Routing Problem},
  volume    = {257},
  year      = {2017}
}

@article{Gadegaard2021,
  author    = {S. L. Gadegaard and J. Lysgaard},
  doi       = {10.1016/j.dam.2020.02.012},
  journal   = {Discrete Appl. Math.},
  pages     = {179--192},
  publisher = {Elsevier B.V.},
  title     = {A Symmetry-Free Polynomial Formulation of the Capacitated Vehicle Routing Problem},
  volume    = {296},
  year      = {2021}
}

@inproceedings{Saadaoui2019,
  author    = {Rabbouch, Bochra
               and Sa{\^a}daoui, Foued
               and Mraihi, Rafaa},
  title     = {Constraint Programming Based Algorithm for Solving Large-Scale Vehicle Routing Problems},
  booktitle = {Hybrid Artificial Intelligent Systems},
  year      = {2019},
  publisher = {Springer International Publishing},
  address   = {Cham},
  pages     = {526--539},
  doi       = {10.1007/978-3-030-29859-3_45}
}

@article{Hungerlander2018,
  author    = {Philipp Hungerl\"{a}nder and Christian Truden},
  journal   = {Transp. Res. Proc.},
  pages     = {157--166},
  publisher = {Elsevier B.V.},
  title     = {Efficient and Easy-to-Implement Mixed-Integer Linear Programs for the Traveling Salesperson Problem with Time Windows},
  volume    = {30},
  doi       = {10.1016/j.trpro.2018.09.018},
  year      = {2018}
}

@techreport{Gavish1978,
  author      = {Bezalel Gavish and Stephen C. Graves},
  title       = {The Travelling Salesman Problem and Related Problems},
  institution = {Operations Research Center, Massachusetts Institute of Technology},
  note        = {Working Paper OR 078-78},
  year        = {1978}
}

@article{Dumas1995,
  author  = {Yvan Dumas and Jacques Desrosiers and Eric Gelinas and Marius M Solomon},
  number  = {2},
  journal = {Oper. Res.},
  title   = {An Optimal Algorithm for the Traveling Salesman Problem with Time Windows},
  volume  = {43},
  doi     = {10.1287/opre.43.2.367},
  pages   = {367--371},
  year    = {1995}
}

@article{Booth2016,
  author    = {Kyle E.C. Booth and Tony T. Tran and Goldie Nejat and J. Christopher Beck},
  doi       = {10.1109/LRA.2016.2522096},
  number    = {1},
  journal   = {IEEE Robot. Autom. Lett.},
  pages     = {500--507},
  publisher = {Institute of Electrical and Electronics Engineers Inc.},
  title     = {Mixed-Integer and Constraint Programming Techniques for Mobile Robot Task Planning},
  volume    = {1},
  year      = {2016}
}

@article{Morrison2014,
  author    = {David R. Morrison and Edward C. Sewell and Sheldon H. Jacobson},
  doi       = {10.1016/j.ejor.2013.11.033},
  number    = {2},
  journal   = {Eur. J. Oper. Res.},
  pages     = {403--409},
  publisher = {Elsevier B.V.},
  title     = {An Application of the Branch, Bound, and Remember Algorithm to a New Simple Assembly Line Balancing Dataset},
  volume    = {236},
  year      = {2014}
}

@article{Ritt2018,
  author  = {Marcus Ritt and Alysson M. Costa},
  doi     = {10.1111/itor.12206},
  number  = {4},
  journal = {Int. Trans. Oper. Res.},
  pages   = {1345--1359},
  title   = {Improved Integer Programming Models for Simple Assembly Line Balancing and Related Problems},
  volume  = {25},
  year    = {2018}
}

@article{Bukchin2018,
  author    = {Yossi Bukchin and Tal Raviv},
  doi       = {10.1016/j.omega.2017.06.008},
  journal   = {Omega},
  pages     = {57--68},
  publisher = {Elsevier Ltd},
  title     = {Constraint Programming for Solving Various Assembly Line Balancing Problems},
  volume    = {78},
  year      = {2018}
}

@article{Sewell2012,
  author  = {E. C. Sewell and S. H. Jacobson},
  doi     = {10.1287/ijoc.1110.0462},
  number  = {3},
  journal = {INFORMS J. Comput.},
  pages   = {433--442},
  title   = {A Branch, Bound, and Remember Algorithm for the Simple Assembly Line Balancing Problem},
  volume  = {24},
  year    = {2012}
}

@article{Delorme2018,
  author    = {Maxence Delorme and Manuel Iori and Silvano Martello},
  doi       = {10.1007/s11590-017-1192-z},
  number    = {2},
  journal   = {Optim. Lett.},
  pages     = {235--250},
  publisher = {Springer Verlag},
  title     = {{BPPLIB}: A Library for Bin Packing and Cutting Stock Problems},
  volume    = {12},
  year      = {2018}
}

@article{Delorme2016,
  author    = {Maxence Delorme and Manuel Iori and Silvano Martello},
  doi       = {10.1016/j.ejor.2016.04.030},
  number    = {1},
  journal   = {Eur. J. Oper. Res.},
  pages     = {1--20},
  publisher = {Elsevier},
  title     = {Bin Packing and Cutting Stock Problems: Mathematical Models and Exact Algorithms},
  volume    = {255},
  year      = {2016}
}

@inproceedings{Shaw2004,
  author    = {Shaw, Paul},
  title     = {A Constraint for Bin Packing},
  booktitle = {Principles and Practice of Constraint Programming -- CP 2004},
  year      = {2004},
  publisher = {Springer},
  address   = {Berlin, Heidelberg},
  pages     = {648--662},
  doi       = {10.1007/978-3-540-30201-8_47}
}

@article{Martin2021,
  author  = {Mateus Martin and Horacio Hideki Yanasse and Maria Jos\'{e} Pinto},
  doi     = {10.1111/itor.13053},
  volume  = {29},
  number  = {5},
  pages   = {2944--2967},
  journal = {Int. Trans. Oper. Res.},
  title   = {Mathematical Models for the Minimization of Open Stacks Problem},
  year    = {2021}
}

@article{DeLaBanda2007,
  author    = {{Garcia de la Banda}, Maria and Stuckey, Peter J.},
  doi       = {10.1287/ijoc.1060.0205},
  number    = {4},
  journal   = {INFORMS J. Comput.},
  pages     = {607--617},
  publisher = {INFORMS Inst.for Operations Res.and the Management Sciences},
  title     = {Dynamic Programming to Minimize the Maximum Number of Open Stacks},
  volume    = {19},
  year      = {2007}
}

@inproceedings{Chu2009,
  author    = {Geoffrey Chu and Peter J. Stuckey},
  doi       = {10.1007/978-3-642-04244-7_21},
  booktitle = {Principles and Practice of Constraint Programming -- CP 2009},
  pages     = {242--257},
  publisher = {Springer},
  address   = {Berlin, Heidelberg},
  title     = {Minimizing the Maximum Number of Open Stacks by Customer Search},
  year      = {2009}
}

@inproceedings{Kolling2007,
  author    = {Andreas Kolling and Stefano Carpin},
  doi       = {10.1109/IROS.2007.4399368},
  booktitle = {Proceedings of IEEE International Conference on Intelligent Robots and Systems (IROS)},
  pages     = {1003--1008},
  title     = {The GRAPH-CLEAR Problem: Definition, Theoretical Properties and its Connections to Multirobot Aided Surveillance},
  year      = {2007}
}

@article{Fusy2009,
  author  = {Fusy, \'{E}ric},
  doi     = {10.1002/rsa.20275},
  number  = {4},
  journal = {Random Struct. Algor.},
  pages   = {464--522},
  title   = {Uniform Random Sampling of Planar Graphs in Linear Time},
  volume  = {35},
  year    = {2009}
}

@article{Morin2018,
  author    = {Michael Morin and Margarita P. Castro and Kyle E.C. Booth and Tony T. Tran and Chang Liu and J. Christopher Beck},
  doi       = {10.1007/s10601-018-9288-3},
  number    = {3},
  journal   = {Constraints},
  pages     = {335--354},
  publisher = {Constraints},
  title     = {Intruder Alert! {O}ptimization Models for Solving the Mobile Robot Graph-Clear Problem},
  volume    = {23},
  year      = {2018}
}

@article{Dijkstra1959,
  title     = {A Note on Two Problems in Connexion with Graphs},
  author    = {Dijkstra, Edsger W},
  journal   = {Numer. Math.},
  doi       = {10.1007/BF01386390},
  number    = {1},
  pages     = {269--271},
  year      = {1959},
  publisher = {Springer}
}

@article{GNUParallel,
  author  = {Ole Tange},
  title   = {{GNU} Parallel - The Command-Line Power Tool},
  journal = {;login: USENIX Mag.},
  volume  = {36},
  pages   = {42--47},
  year    = {2011}
}

@article{Gendreau1998,
  author    = {Michel Gendreau and Alain Hertz and Gilbert Laporte and Mihnea Stan},
  doi       = {10.1287/opre.46.3.330},
  number    = {3},
  journal   = {Oper. Res.},
  pages     = {330--346},
  publisher = {INFORMS Inst.for Operations Res.and the Management Sciences},
  title     = {A Generalized Insertion Heuristic for the Traveling Salesman Problem with Time Windows},
  volume    = {46},
  year      = {1998}
}

@article{Ohlmann2007,
  author    = {Jeffrey W. Ohlmann and Barrett W. Thomas},
  doi       = {10.1287/ijoc.1050.0145},
  number    = {1},
  journal   = {INFORMS J. Comput.},
  pages     = {80--90},
  publisher = {INFORMS Inst.for Operations Res.and the Management Sciences},
  title     = {A Compressed-Annealing Heuristic for the Traveling Salesman Problem with Time Windows},
  volume    = {19},
  year      = {2007}
}

@phdthesis{Ascheuer1995,
  author = {Norbert Ascheuer},
  title  = {Hamiltonian Path Problems in the On-Line Optimization of Flexible Manufacturing Systems},
  school = {Technische Universit\"{a}t Berlin},
  year   = 1995
}

@book{Martello1990,
  author    = {Martello, Silvano and Toth, Paolo},
  title     = {Knapsack Problems: Algorithms and Computer Implementations},
  year      = {1990},
  publisher = {John Wiley \& Sons, Inc.},
  address   = {New York, NY, USA}
}

@misc{ConstraintModellingChallenge,
  title        = {Constraint Modelling Challenge Report 2005},
  author       = {Smith, Barbara and Gent, Ian},
  year         = {2005},
  howpublished = {\url{https://ipg.host.cs.st-andrews.ac.uk/challenge/}}
}

@article{Faggioli1998,
  author   = {Enrico Faggioli and Carlo Alberto Bentivoglio},
  journal  = {Eur. J. Oper. Res.},
  pages    = {564--575},
  title    = {Heuristic and Exact Methods for the Cutting Sequencing Problem},
  volume   = {110},
  number   = {3},
  doi      = {10.1016/S0377-2217(97)00268-3},
  year     = {1998}
}

@article{Hart1968,
  author  = {Peter E. Hart and
             Nils J. Nilsson and
             Bertram Raphael},
  title   = {A Formal Basis for the Heuristic Determination of Minimum Cost Paths},
  journal = {{IEEE} Trans. Syst. Sci. Cybern.},
  volume  = {4},
  number  = {2},
  pages   = {100--107},
  doi     = {10.1109/TSSC.1968.300136},
  year    = {1968}
}

@inproceedings{Edelkamp2005,
  author    = {Stefan Edelkamp and Shahid Jabbar and Alberto Lluch Lafuente},
  booktitle = {Proceedings of the 20th National Conference on Artificial Intelligence (AAAI)},
  pages     = {1362--1367},
  title     = {Cost-Algebraic Heuristic Search},
  publisher = {{AAAI} Press},
  year      = {2005}
}

@article{Gromicho2012,
  author    = {J. Gromicho and J. J. Van Hoorn and A. L. Kok and J. M.J. Schutten},
  doi       = {10.1016/j.cor.2011.07.002},
  number    = {5},
  journal   = {Comput. Oper. Res.},
  pages     = {902--909},
  publisher = {Elsevier},
  title     = {Restricted Dynamic Programming: A Flexible Framework for Solving Realistic {VRP}s},
  volume    = {39},
  year      = {2012}
}

@article{Scholl1997,
  author  = {Armin Scholl and Robert Klein},
  number  = {4},
  journal = {INFORMS J. Comput.},
  pages   = {319--335},
  title   = {{SALOME}: A Bidirectional Branch-and-Bound Procedure for Assembly Line Balancing},
  volume  = {9},
  doi     = {10.1287/ijoc.9.4.319},
  year    = {1997}
}

@article{Bergman2016,
  author    = {Bergman, David and Cire, Andre A. and {van Hoeve}, Willem-Jan and Hooker, J. N.},
  doi       = {10.1287/ijoc.2015.0648},
  number    = {1},
  journal   = {INFORMS J. Comput.},
  pages     = {47--66},
  publisher = {INFORMS Inst.for Operations Res.and the Management Sciences},
  title     = {Discrete Optimization with Decision Diagrams},
  volume    = {28},
  year      = {2016}
}

@article{Yuen1995,
  author   = {Boon J Yuen and Ken V Richardson},
  journal  = {Eur. J. Oper. Res.},
  pages    = {590--598},
  title    = {Establishing the Optimality of Sequencing Heuristics for Cutting Stock Problems},
  volume   = {84},
  number   = {3},
  doi      = {10.1016/0377-2217(95)00025-L},
  year     = {1995}
}

@inproceedings{Thayer2011,
  author    = {Jordan T. Thayer and Wheeler Ruml},
  doi       = {10.5591/978-1-57735-516-8/IJCAI11-119},
  booktitle = {Proceedings of the 22nd International Joint Conference on Artificial Intelligence, {IJCAI-11}},
  publisher = {{AAAI} Press/International Joint Conferences on Artificial Intelligence Orginization},
  address   = {Menlo Park, California},
  pages     = {674--679},
  title     = {Bounded Suboptimal Search: A Direct Approach Using Inadmissible Estimates},
  year      = {2011}
}

@techreport{Ghallab1998,
  author      = {Malik Ghallab and Adele Howe and Craig Knoblock and Drew McDermott and Ashwin Ram and Manuela Veloso and Daniel Weld and David Wilkins},
  title       = {{PDDL} - The Planning Domain Definition Language},
  institution = {Yale Center for Computational Vison and Control},
  note        = {CVC TR-98-003/DCS TR-1165},
  year        = {1998}
}

@article{DeLaBanda2011,
  author   = {Maria {Garcia de la Banda} and Peter J. Stuckey and Geoffrey Chu},
  doi      = {10.1287/ijoc.1090.0378},
  number   = {1},
  journal  = {INFORMS J. Comput.},
  pages    = {120--137},
  title    = {Solving Talent Scheduling with Dynamic Programming},
  volume   = {23},
  year     = {2011}
}

@article{Boysen2021,
  author    = {Nils Boysen and Philipp Schulze and Armin Scholl},
  doi       = {10.1016/j.ejor.2021.11.043},
  number    = {3},
  journal   = {Eur. J. Oper. Res.},
  publisher = {Elsevier B.V.},
  title     = {Assembly Line Balancing: What Happened in the Last Fifteen Years?},
  volume    = {301},
  year      = {2021},
  pages     = {797--814}
}

@book{VRP2014,
  author    = {Toth, Paolo and Vigo, Daniele},
  title     = {Vehicle Routing: Problems, Methods, and Applications},
  year      = {2014},
  edition   = {Second},
  publisher = {Society for Industrial and Applied Mathematics},
  doi       = {10.1137/1.9781611973594}
}

@book{Lew2006,
  author    = {Art Lew and Holger Mauch},
  publisher = {Springer},
  address   = {Berlin, Heidelberg},
  title     = {Dynamic Programming: A Computational Tool},
  year      = {2006},
  doi       = {10.1007/978-3-540-37014-7}
}

@inproceedings{Giegerich2002,
  author    = {Robert Giegerich and Carsten Meyer},
  booktitle = {Proceedings of the Ninth Algebraic Methodology and Software Technology},
  pages     = {349--364},
  title     = {Algebraic Dynamic Programming},
  publisher = {Springer},
  address   = {Berlin, Heidelberg},
  doi       = {10.1007/3-540-45719-4_24},
  year      = {2002}
}

@article{zuSiederdissen2015,
  author    = {Christian H\"{o}ner {zu Siederdissen} and Sonja J. Prohaska and Peter F. Stadler},
  doi       = {10.1186/1471-2105-16-S19-S2},
  number    = {19},
  journal   = {BMC Bioinformatics},
  pmid      = {26695390},
  publisher = {BioMed Central Ltd.},
  title     = {Algebraic Dynamic Programming over General Data Structures},
  volume    = {16},
  year      = {2015},
  numpages  = {13}
}

@article{KuroiwaSPCB22,
  author  = {Ryo Kuroiwa and
             Alexander Shleyfman and
             Chiara Piacentini and
             Margarita P. Castro and
             J. Christopher Beck},
  title   = {The {LM}-Cut Heuristic Family for Optimal Numeric Planning with Simple Conditions},
  journal = {J. Artif. Intell. Res.},
  volume  = {75},
  pages   = {1477--1548},
  year    = {2022},
  doi     = {10.1613/jair.1.14034}
}

@inproceedings{KuroiwaSB2022,
  title        = {{LM}-Cut Heuristics for Optimal Linear Numeric Planning},
  doi          = {10.1609/icaps.v32i1.19803},
  abstractnote = {While numeric variables play an important, sometimes central, role in many planning problems arising from real world scenarios, most of the currently available heuristic search planners either do not support such variables or impose heavy restrictions on them. In particular, most admissible heuristics are restricted to domains where actions can only change numeric variables by predetermined constants. In this work, we consider the setting of optimal numeric planning with linear effects, where actions can have numeric effects that assign the result of the evaluation of a linear formula. We extend a recent formulation of Numeric LM-cut for simple effects by adding conditional effects and second-order simple effects, allowing the heuristic to produce admissible estimates for tasks with linear numeric effects. Empirical comparison shows that the proposed LM-cut heuristics favorably compete with the currently available state-of-the-art heuristics and achieve significant improvement in coverage in the domains with second-order simple effects.},
  booktitle    = {Proceedings of the 32nd International Conference on Automated Planning and Scheduling (ICAPS)},
  author       = {Kuroiwa, Ryo and Shleyfman, Alexander and Beck, J. Christopher},
  publisher    = {{AAAI} Press},
  address      = {Palo Alto, California USA},
  year         = {2022},
  pages        = {203--212}
}

@book{Picat,
  author    = {Neng-Fa Zhou and H\r{a}kan Kjellerstrand and Jonathan Fruhman},
  title     = {Constraint Solving and Planning with Picat},
  year      = {2015},
  publisher = {Springer},
  address   = {Cham},
  doi       = {10.1007/978-3-319-25883-6}
}

@article{Freuder1997,
  author   = {Eugene Freuder},
  doi      = {10.1023/A:1009749006768},
  number   = {1},
  journal  = {Constraints},
  pages    = {57--61},
  title    = {In Pursuit of the Holy Grail},
  volume   = {2},
  year     = {1997}
}

@inproceedings{Gillard2020,
  title     = {Ddo, a Generic and Efficient Framework for {MDD}-Based Optimization},
  author    = {Gillard, Xavier and Schaus, Pierre and Coppé, Vianney},
  booktitle = {Proceedings of the 29th International Joint Conference on
               Artificial Intelligence, {IJCAI-20}},
  publisher = {International Joint Conferences on Artificial Intelligence Organization},
  pages     = {5243--5245},
  year      = {2020},
  note      = {Demos},
  doi       = {10.24963/ijcai.2020/757}
}

@inproceedings{Gillard2021,
  author    = {Gillard, Xavier
               and Copp{\'e}, Vianney
               and Schaus, Pierre
               and Cire, Andr{\'e} Augusto},
  title     = {Improving the Filtering of Branch-and-Bound {MDD} Solver},
  booktitle = {Integration of Constraint Programming, Artificial Intelligence, and Operations Research -- 18th International Conference, {CPAIOR} 2021},
  year      = {2021},
  pages     = {231--247},
  publisher = {Springer International Publishing},
  address   = {Cham},
  doi       = {10.1007/978-3-030-78230-6_15}
}

@inproceedings{Kuroiwa2023CAASDy,
  title     = {Domain-Independent Dynamic Programming: Generic State Space Search for Combinatorial Optimization},
  author    = {Ryo Kuroiwa and J. Christopher Beck},
  year      = {2023},
  booktitle = {Proceedings of the 33rd International Conference on Automated Planning and Scheduling (ICAPS)},
  publisher = {{AAAI} Press},
  address   = {Palo Alto, California USA},
  doi       = {10.1609/icaps.v33i1.27200},
  pages     = {236--244}
}

@inproceedings{Kuroiwa2023Anytime,
  title     = {Solving Domain-Independent Dynamic Programming Problems with Anytime Heuristic Search},
  author    = {Ryo Kuroiwa and J. Christopher Beck},
  year      = {2023},
  booktitle = {Proceedings of the 33rd International Conference on Automated Planning and Scheduling (ICAPS)},
  publisher = {{AAAI} Press},
  address   = {Palo Alto, California USA},
  doi       = {10.1609/icaps.v33i1.27201},
  pages     = {245--253}
}

@inproceedings{Zhang1998,
  author    = {Weixiong Zhang},
  booktitle = {Proceedings of the 15th National Conference on Artificial Intelligence (AAAI)},
  publisher = {{AAAI} Press},
  pages     = {425--430},
  title     = {Complete Anytime Beam Search},
  year      = {1998}
}

@article{Laborie2018,
  author    = {Laborie, Philippe and Rogerie, Jérôme and Shaw, Paul and Vilím, Petr},
  doi       = {10.1007/s10601-018-9281-x},
  journal   = {Constraints},
  number    = {2},
  pages     = {210--250},
  publisher = {Constraints},
  title     = {{IBM} {ILOG} {CP} Optimizer for Scheduling},
  volume    = {23},
  year      = {2018}
}

@article{Berthold2013,
  author   = {Timo Berthold},
  doi      = {10.1016/j.orl.2013.08.007},
  number   = {6},
  journal  = {Oper. Res. Lett.},
  pages    = {611--614},
  title    = {Measuring the Impact of Primal Heuristics},
  volume   = {41},
  year     = {2013}
}

@article{Hernandez-Perez2009,
  author   = {Hipólito Hernández-Pérez and Juan José Salazar-González},
  doi      = {10.1016/j.ejor.2008.05.009},
  number   = {3},
  journal  = {Eur. J. Oper. Res.},
  pages    = {987--995},
  title    = {The Multi-Commodity One-to-One Pickup-and-Delivery Traveling Salesman Problem},
  volume   = {196},
  year     = {2009}
}

@inproceedings{Hooker2013,
  author    = {Hooker, John N.},
  title     = {Decision Diagrams and Dynamic Programming},
  booktitle = {Integration of AI and OR Techniques in Constraint Programming for Combinatorial Optimization Problems -- 10th International Conference, {CPAIOR} 2013},
  year      = {2013},
  pages     = {94--110},
  publisher = {Springer},
  address   = {Berlin, Heidelberg},
  doi       = {10.1007/978-3-642-38171-3_7}
}

@inproceedings{Gentzel2020,
  author    = {Gentzel, Rebecca
               and Michel, Laurent
               and {van Hoeve}, W.-J.},
  title     = {HADDOCK: A Language and Architecture for Decision Diagram Compilation},
  booktitle = {Principles and Practice of Constraint Programming -- CP 2020},
  year      = {2020},
  publisher = {Springer International Publishing},
  address   = {Cham},
  pages     = {531--547},
  doi       = {10.1007/978-3-030-58475-7_31}
}

@article{FikesN1971,
  title    = {STRIPS: A New Approach to the Application of Theorem Proving to Problem Solving},
  journal  = {Artif. Intell.},
  volume   = {2},
  number   = {3},
  pages    = {189--208},
  year     = {1971},
  doi      = {10.1016/0004-3702(71)90010-5},
  author   = {Richard E. Fikes and Nils J. Nilsson}
}

@article{BackstromN95,
  author  = {Christer B{\"{a}}ckstr{\"{o}}m and
             Bernhard Nebel},
  title   = {Complexity Results for {SAS+} Planning},
  journal = {Comput. Intell.},
  volume  = {11},
  pages   = {625--656},
  year    = {1995},
  moth    = {11},
  doi     = {10.1111/j.1467-8640.1995.tb00052.x}
}

@inproceedings{Helmert02,
  author    = {Malte Helmert},
  title     = {Decidability and Undecidability Results for Planning with Numerical State Variables},
  booktitle = {Proceedings of the Sixth International Conference on Artificial Intelligence Planning Systems (AIPS)},
  pages     = {44--53},
  publisher = {{AAAI} Press},
  year      = {2002}
}

@inproceedings{ShleyfmanGJ2023,
  title        = {Structurally Restricted Fragments of Numeric Planning -- a Complexity Analysis},
  doi          = {10.1609/aaai.v37i10.26428},
  abstractnote = {Numeric planning is known to be undecidable even under severe restrictions. Prior work has investigated the decidability boundaries by restricting the expressiveness of the planning formalism in terms of the numeric functions allowed in conditions and effects. We study a well-known restricted form of Hoffmann’s simple numeric planning, which is undecidable. We analyze the complexity by imposing restrictions on the causal structure, exploiting a novel method for bounding variable domain sizes. First, we show that plan existence for tasks where all numeric variables are root nodes in the causal graph is in PSPACE.
                  Second, we show that for tasks with only numeric leaf variables the problem is decidable, and that it is in PSPACE if the propositional state space has a fixed size. Our work lays a strong foundation for future investigations of structurally more complex tasks. From a practical perspective, our method allows to employ heuristics and methods that are geared towards finite variable domains (such as pattern database heuristics or decoupled search) to solve non-trivial families of numeric planning problems.},
  booktitle    = {Proceedings of the 36th AAAI Conference on Artificial Intelligence (AAAI)},
  publisher    = {{AAAI} Press},
  address      = {Washington, DC, USA},
  author       = {Shleyfman, Alexander and Gnad, Daniel and Jonsson, Peter},
  year         = {2023},
  pages        = {12112--12119}
}

@inproceedings{GnadHJS2023,
  title        = {Planning over Integers: Compilations and Undecidability},
  doi          = {10.1609/icaps.v33i1.27189},
  abstractnote = {Restricted Tasks (RT) are a special case of numeric planning characterized by numeric conditions that involve one numeric variable per formula and numeric effects that allow only the addition of constants. Despite this, RTs form an expressive class whose planning problem is undecidable. The restricted nature of RTs often makes problem modeling awkward and unnecessarily complicated. We show that this can be alleviated by compiling mathematical operations that are not natively supported into RTs using macro-like action sequences. With that, we can encode many features found in general numeric planning such as constant multiplication, addition of linear formulas, and integer division and residue. We demonstrate how our compilations can be used to capture challenging mathematical problems such as the (in)famous Collatz conjecture. Our approach additionally gives a simple undecidability proof for RTs, and the proof shows that the number of variables needed to construct an undecidable class of RTs is
                  surprisingly low: two numeric and one propositional variable.},
  booktitle    = {Proceedings of the 23rd International Conference on Automated Planning and Scheduling (ICAPS)},
  publisher    = {{AAAI} Press},
  address      = {Palo Alto, California USA},
  author       = {Gnad, Daniel and Helmert, Malte and Jonsson, Peter and Shleyfman, Alexander},
  year         = {2023},
  pages        = {148--152}
}

@inproceedings{GiganteS2023,
  title     = {On the Compilability of Bounded Numeric Planning},
  author    = {Gigante, Nicola and Scala, Enrico},
  booktitle = {Proceedings of the 32nd International Joint Conference on Artificial Intelligence, {IJCAI-23}},
  publisher = {International Joint Conferences on Artificial Intelligence Organization},
  pages     = {5341--5349},
  year      = {2023},
  note      = {Main track},
  doi       = {10.24963/ijcai.2023/593}
}

@article{FoxL03,
  author  = {Maria Fox and Derek Long},
  title   = {{PDDL}2.1: An Extension to {PDDL} for Expressing Temporal Planning Domains},
  journal = {J. Artif. Intell. Res.},
  volume  = {20},
  pages   = {61--124},
  year    = {2003},
  doi     = {10.1613/jair.1129}
}

@article{BonetG2001,
  title    = {Planning as Heuristic Search},
  journal  = {Artificial Intelligence},
  volume   = {129},
  number   = {1},
  pages    = {5--33},
  year     = {2001},
  doi      = {10.1016/S0004-3702(01)00108-4},
  author   = {Blai Bonet and Héctor Geffner}
}

@article{HoffmannN01,
  author  = {J{\"{o}}rg Hoffmann and
             Bernhard Nebel},
  title   = {The {FF} Planning System: Fast Plan Generation Through Heuristic Search},
  journal = {Journal of Artificial Intelligence Research},
  volume  = {14},
  pages   = {253--302},
  year    = {2001},
  doi     = {10.1613/jair.855}
}

@article{Helmert06,
  author  = {Malte Helmert},
  title   = {The {Fast Downward} Planning System},
  journal = {J. Artif. Intell. Res.},
  volume  = {26},
  pages   = {191--246},
  year    = {2006},
  doi     = {10.1613/jair.1705}
}

@article{ScalaHTR20,
  author  = {Enrico Scala and
             Patrik Haslum and
             Sylvie Thi{\'{e}}baux and
             Miquel Ram{\'{\i}}rez},
  title   = {Subgoaling Techniques for Satisficing and Optimal Numeric Planning},
  journal = {J. Artif. Intell. Res.},
  volume  = {68},
  pages   = {691--752},
  year    = {2020},
  doi     = {10.1613/jair.1.11875}
}

@article{Hoffmann03,
  author  = {J{\"{o}}rg Hoffmann},
  title   = {The {Metric-FF} Planning System: Translating ''Ignoring Delete Lists'' to Numeric State Variables},
  journal = {J. Artif. Intell. Res.},
  volume  = {20},
  pages   = {291--341},
  year    = {2003},
  doi     = {10.1613/jair.1144}
}

@inproceedings{Torralba015,
  author    = {{\'{A}}lvaro Torralba and
               J{\"{o}}rg Hoffmann},
  title     = {Simulation-Based Admissible Dominance Pruning},
  address   = {Palo Alto, California USA},
  booktitle = {Proceedings of the 24th International Joint Conference on Artificial Intelligence, {IJCAI}-15},
  pages     = {1689--1695},
  publisher = {{AAAI} Press/International Joint Conferences on Artificial Intelligence Organization},
  year      = {2015}
}

@inproceedings{Libralesso2020,
  author    = {Luc Libralesso and Abdel-Malik Bouhassoun and Hadrien Cambazard and Vincent Jost},
  booktitle = {{ECAI} 2020 -- 24th European Conference on Artificial Intelligence},
  series    = {Frontiers in Artificial Intelligence and Applications},
  volume    = {325},
  pages     = {459--465},
  title     = {Tree Search for the Sequential Ordering Problem},
  publisher = {{IOS} Press},
  doi       = {10.3233/FAIA200126},
  year      = {2020}
}

@inproceedings{Vadlamudi2012,
  author    = {Satya Gautam Vadlamudi and Piyush Gaurav and Sandip Aine and Partha Pratim Chakrabarti},
  booktitle = {AI 2012: Advances in Artificial Intelligence},
  publisher = {Springer},
  address   = {Berlin, Heidelberg},
  pages     = {254--265},
  doi       = {10.1007/978-3-642-35101-3_22},
  title     = {Anytime Column Search},
  year      = {2012}
}

@article{Vadlamudi2016,
  author  = {Satya Gautam Vadlamudi and Sandip Aine and Partha Pratim Chakrabarti},
  number  = {3},
  journal = {Nat. Comput.},
  pages   = {395--414},
  doi     = {10.1007/978-3-642-45062-4_88},
  title   = {Anytime Pack Search},
  volume  = {15},
  year    = {2016}
}

@inproceedings{Harvey1995,
  author    = {William D. Harvey and Matthew L. Ginsberg},
  booktitle = {Proceedings of the 14th International Joint Conference on Artificial Intelligence, {IJCAI-95}},
  pages     = {607--613},
  publisher = {Morgan Kaufmann Publishers Inc.},
  address   = {San Francisco, CA, USA},
  title     = {Limited Discrepancy Search},
  year      = {1995}
}

@inproceedings{Beck2000,
  author    = {J. Christopher Beck and Laurent Perron},
  title     = {Discrepancy-Bounded Depth First Search},
  booktitle = {Second International Workshop on Integration of AI and OR Technologies for Combinatorial Optimization Problems, {CPAIOR} 2000},
  numpages  = {10},
  year      = {2000}
}

@article{Kao2009,
  author  = {Gio K. Kao and Edward C. Sewell and Sheldon H. Jacobson},
  number  = {2},
  journal = {J. Sched.},
  pages   = {163--175},
  title   = {A Branch, Bound, and Remember Algorithm for the {$1|r_t| \sum t_i$} Scheduling Problem},
  volume  = {12},
  doi     = {10.1007/s10951-008-0087-3},
  year    = {2009}
}

@article{Libralesso2022,
  author  = {Luc Libralesso and Pablo Andres Focke and Aurélien Secardin and Vincent Jost},
  number  = {1},
  journal = {Eur. J. Oper. Res.},
  pages   = {217--234},
  title   = {Iterative Beam Search Algorithms for the Permutation Flowshop},
  volume  = {301},
  doi     = {10.1016/j.ejor.2021.10.015},
  year    = {2022}
}

@book{Bellman1957,
  author    = {Richard Bellman },
  title     = {Dynamic Programming},
  publisher = {Princeton University Press},
  year      = {1957}
}

@article{Svelsbergh1985,
  title   = {Local Search in Routing Problems with Time Windows},
  author  = {Savelsbergh, M. W. P.},
  journal = {Ann. Oper. Res.},
  volume  = {4},
  number  = {1},
  pages   = {285--305},
  year    = {1985},
  doi     = {10.1007/BF02022044}
}

@article{Ibaraki1978,
  title    = {Branch-and-Bound Procedure and State-Space Representation of Combinatorial Optimization Problems},
  journal  = {Inf. Control},
  volume   = {36},
  number   = {1},
  pages    = {1--27},
  year     = {1978},
  doi      = {10.1016/S0019-9958(78)90197-3},
  author   = {Toshihide Ibaraki}
}

@article{Nau1984,
  title    = {General Branch and Bound, and its Relation to {A*} and {AO*}},
  journal  = {Artif. Intell.},
  volume   = {23},
  number   = {1},
  pages    = {29--58},
  year     = {1984},
  doi      = {10.1016/0004-3702(84)90004-3},
  author   = {Dana S. Nau and Vipin Kumar and Laveen Kanal}
}

@article{Zhou2006,
  title    = {Breadth-First Heuristic Search},
  journal  = {Artif. Intell.},
  volume   = {170},
  number   = {4},
  pages    = {385--408},
  year     = {2006},
  doi      = {10.1016/j.artint.2005.12.002},
  author   = {Rong Zhou and Eric A. Hansen}
}

@inproceedings{Martelli1975,
  author    = {Martelli, A. and Montanari, U.},
  title     = {From Dynamic Programming to Search Algorithms with Functional Costs},
  year      = {1975},
  publisher = {Morgan Kaufmann Publishers Inc.},
  address   = {San Francisco, CA, USA},
  booktitle = {Proceedings of the Fourth International Joint Conference on Artificial Intelligence, {IJCAI-75}},
  pages     = {345--350},
  numpages  = {6}
}

@article{Karp1967,
  title     = {Finite-State Processes and Dynamic Programming},
  author    = {Karp, Richard M. and Held, Michael},
  journal   = {SIAM J. Appl. Math.},
  volume    = {15},
  number    = {3},
  pages     = {693--718},
  year      = {1967},
  publisher = {Society for Industrial and Applied Mathematics},
  doi       = {10.1137/0115060}
}

@article{Castro2020,
  author    = {Margarita P. Castro and Andre A. Cire and J. Christopher Beck},
  number    = {2},
  journal   = {INFORMS J. Comput.},
  pages     = {263--278},
  publisher = {INFORMS Inst.for Operations Res.and the Management Sciences},
  title     = {An {MDD}-Based Lagrangian Approach to the Multicommodity Pickup-and-Delivery {TSP}},
  volume    = {32},
  year      = {2020},
  doi       = {10.1287/ijoc.2018.0881}
}

@article{AbdulRazaq1990,
  author  = {T. S. Abdul-Razaq and C. N. Potts and L. N. Van Wassenhove},
  journal = {Discrete Appl. Math.},
  pages   = {235--253},
  title   = {A Survey of Algorithms for the Single Machine Total Weighted Tardiness Scheduling Problem},
  volume  = {26},
  number  = {2},
  year    = {1990},
  doi     = {10.1016/0166-218X(90)90103-J}
}

@article{Emmons1969,
  author  = {Hamilton Emmons},
  number  = {4},
  journal = {Oper. Res.},
  pages   = {701--715},
  title   = {One-Machine Sequencing to Minimize Certain Functions of Job Tardiness},
  volume  = {17},
  year    = {1969},
  doi     = {10.1287/opre.17.4.701}
}

@article{Beasley1990,
  doi       = {10.2307/2582903},
  author    = {J. E. Beasley},
  journal   = {J. Oper. Res. Soc.},
  number    = {11},
  pages     = {1069--1072},
  publisher = {Palgrave Macmillan Journals},
  title     = {{OR-Library}: Distributing Test Problems by Electronic Mail},
  volume    = {41},
  year      = {1990}
}

@article{Keha2009,
  title    = {Mixed Integer Programming Formulations for Single Machine Scheduling Problems},
  journal  = {Comput. Ind. Eng.},
  volume   = {56},
  number   = {1},
  pages    = {357--367},
  year     = {2009},
  doi      = {10.1016/j.cie.2008.06.008},
  author   = {Ahmet B. Keha and Ketan Khowala and John W. Fowler}
}

@article{Gouveia2015,
  title    = {Load-Dependent and Precedence-Based Models for Pickup and Delivery Problems},
  journal  = {Comput. Oper. Res.},
  volume   = {63},
  pages    = {56--71},
  year     = {2015},
  doi      = {10.1016/j.cor.2015.04.008},
  author   = {Luis Gouveia and Mario Ruthmair}
}

@article{Kanet2007,
  doi       = {10.1287/moor.1070.0255},
  author    = {J. J. Kanet},
  journal   = {Math. Oper. Res.},
  number    = {3},
  pages     = {579--588},
  publisher = {INFORMS},
  title     = {New Precedence Theorems for One-Machine Weighted Tardiness},
  volume    = {32},
  year      = {2007}
}

@article{Righini2009,
  title    = {Decremental State Space Relaxation Strategies and Initialization Heuristics for Solving the Orienteering Problem with Time Windows with Dynamic Programming},
  journal  = {Comput. Oper. Res.},
  volume   = {36},
  number   = {4},
  pages    = {1191--1203},
  year     = {2009},
  doi      = {10.1016/j.cor.2008.01.003},
  author   = {Giovanni Righini and Matteo Salani}
}

@article{Kantor1992,
  doi       = {10.1057/jors.1992.88},
  author    = {Marisa G. Kantor and Moshe B. Rosenwein},
  journal   = {J. Oper. Res. Soc.},
  number    = {6},
  pages     = {629--635},
  publisher = {Palgrave Macmillan Journals},
  title     = {The Orienteering Problem with Time Windows},
  volume    = {43},
  year      = {1992}
}

@article{Dantzig1957,
  doi       = {10.1287/opre.5.2.266},
  author    = {George B. Dantzig},
  journal   = {Oper. Res.},
  number    = {2},
  pages     = {266--277},
  publisher = {INFORMS},
  title     = {Discrete-Variable Extremum Problems},
  volume    = {5},
  year      = {1957}
}

@techreport{Righini2006,
  author      = {Righini, Giovanni and Salani, Matteo},
  institution = {Dipartimento di Tecnologie dell'Informazione, Universita degli Studi Milano},
  address     = {Crema, Italy},
  title       = {Dynamic Programming for the Orienteering Problem with Time Windows},
  year        = {2006},
  number      = {91}
}

@article{Righini2008,
  author   = {Righini, Giovanni and Salani, Matteo},
  title    = {New Dynamic Programming Algorithms for the Resource Constrained Elementary Shortest Path Problem},
  journal  = {Networks},
  volume   = {51},
  number   = {3},
  pages    = {155--170},
  doi      = {10.1002/net.20212},
  year     = {2008}
}

@article{Motemanni2009,
  author  = {Montemanni, Roberto and Gambardella, Luca Maria},
  journal = {Found. Comput. Decis. Sci.},
  title   = {Ant Colony System for Team Orienteering Problems with Time Windows},
  volume  = {34},
  year    = {2009},
  pages   = {287--306}
}

@article{Vansteenwegen2009,
  title    = {Iterated Local Search for the Team Orienteering Problem with Time Windows},
  journal  = {Comput. Oper. Res.},
  volume   = {36},
  number   = {12},
  pages    = {3281--3290},
  year     = {2009},
  note     = {new developments on hub location},
  doi      = {10.1016/j.cor.2009.03.008},
  author   = {Pieter Vansteenwegen and Wouter Souffriau and Greet {Vanden Berghe} and Dirk {Van Oudheusden}}
}

@article{Vansteenwegen2011,
  title    = {The Orienteering Problem: A Survey},
  journal  = {Eur. J. Oper. Res.},
  volume   = {209},
  number   = {1},
  pages    = {1--10},
  year     = {2011},
  doi      = {10.1016/j.ejor.2010.03.045},
  author   = {Pieter Vansteenwegen and Wouter Souffriau and Dirk Van Oudheusden}
}

@article{Cacchiani2022,
  title    = {Knapsack Problems — An Overview of Recent Advances. Part {II}: Multiple, Multidimensional, and Quadratic Knapsack Problems},
  journal  = {Comput. Oper. Res.},
  volume   = {143},
  pages    = {105693},
  year     = {2022},
  doi      = {10.1016/j.cor.2021.105693},
  author   = {Valentina Cacchiani and Manuel Iori and Alberto Locatelli and Silvano Martello}
}

@article{Falkenauer1996,
  author   = {Falkenauer, Emanuel},
  title    = {A Hybrid Grouping Genetic Algorithm for Bin Packing},
  journal  = {J. Heuristics},
  year     = {1996},
  volume   = {2},
  number   = {1},
  pages    = {5--30},
  doi      = {10.1007/BF00226291}
}

@article{SchollKJ1997,
  title    = {Bison: A Fast Hybrid Procedure for Exactly Solving the One-Dimensional Bin Packing Problem},
  journal  = {Comput. Oper. Res.},
  volume   = {24},
  number   = {7},
  pages    = {627--645},
  year     = {1997},
  doi      = {10.1016/S0305-0548(96)00082-2},
  author   = {Armin Scholl and Robert Klein and Christian Jürgens}
}

@article{Wascher1996,
  author   = {Wäscher, Gerhard and Gau, Thomas},
  title    = {Heuristics for the Integer One-Dimensional Cutting Stock Problem: A Computational Study},
  journal  = {Oper.-Res.-Spektrum},
  year     = {1996},
  volume   = {18},
  number   = {3},
  pages    = {131--144},
  doi      = {10.1007/BF01539705}
}

@article{Schwerin1997,
  title    = {The Bin-Packing Problem: A Problem Generator and Some Numerical Experiments with {FFD} Packing and {MTP}},
  journal  = {Int. Trans. Oper. Res.},
  volume   = {4},
  number   = {5},
  pages    = {377--389},
  year     = {1997},
  doi      = {10.1016/S0969-6016(97)00025-7},
  author   = {Petra Schwerin and Gerhard Wäscher}
}

@techreport{Schoenfield2002,
  author      = {Schoenfield, J. E.},
  institution = {US Army Space and Missile Defense Command},
  address     = {Hutsville, Alabama, USA},
  title       = {Fast, Exact Solution of Open Bin Packing Problems Without Linear Programming},
  year        = {2002}
}

@article{Qin2016,
  title    = {An Enhanced Branch-and-Bound Algorithm for the Talent Scheduling Problem},
  journal  = {Eur. J. Oper. Res.},
  volume   = {250},
  number   = {2},
  pages    = {412--426},
  year     = {2016},
  doi      = {10.1016/j.ejor.2015.10.002},
  author   = {Hu Qin and Zizhen Zhang and Andrew Lim and Xiaocong Liang}
}

@inproceedings{Chu2015,
  author    = {Chu, Geoffrey
               and Stuckey, Peter J.},
  title     = {Learning Value Heuristics for Constraint Programming},
  booktitle = {Integration of AI and OR Techniques in Constraint Programming -- 12th International Conference, {CPAIOR} 2015},
  year      = {2015},
  publisher = {Springer International Publishing},
  address   = {Cham},
  pages     = {108--123},
  doi       = {10.1007/978-3-319-18008-3_8}
}

@article{Letchford2016,
  title    = {Stronger Multi-Commodity Flow Formulations of the (Capacitated) Sequential Ordering Problem},
  journal  = {Eur. J. Oper. Res.},
  volume   = {251},
  number   = {1},
  pages    = {74--84},
  year     = {2016},
  doi      = {10.1016/j.ejor.2015.11.001},
  author   = {Adam N. Letchford and Juan-José Salazar-González}
}

@inproceedings{Nethercote2007,
  author    = {Nethercote, Nicholas
               and Stuckey, Peter J.
               and Becket, Ralph
               and Brand, Sebastian
               and Duck, Gregory J.
               and Tack, Guido},
  title     = {{MiniZinc}: Towards a Standard {CP} Modelling Language},
  booktitle = {Principles and Practice of Constraint Programming -- CP 2007},
  year      = {2007},
  publisher = {Springer},
  address   = {Berlin, Heidelberg},
  pages     = {529--543}
}

@inproceedings{Cappart2021,
  title        = {Combining Reinforcement Learning and Constraint Programming for Combinatorial Optimization},
  doi          = {10.1609/aaai.v35i5.16484},
  abstractnote = {Combinatorial optimization has found applications in numerous fields, from aerospace to transportation planning and economics. The goal is to find an optimal solution among a finite set of possibilities. The well-known challenge one faces with combinatorial optimization is the state-space explosion problem: the number of possibilities grows exponentially with the problem size, which makes solving intractable for large problems. In the last years, deep reinforcement learning (DRL) has shown its promise for designing good heuristics dedicated to solve NP-hard combinatorial optimization problems. However, current approaches have an important shortcoming: they only provide an approximate solution with no systematic ways to improve it or to prove optimality. In another context, constraint programming (CP) is a generic tool to solve combinatorial optimization problems. Based on a complete search procedure, it will always find the optimal solution if we allow an execution time large enough. A critical design choice, that makes CP non-trivial to use in practice, is the branching decision, directing how the search space is explored. In this work, we propose a general and hybrid approach, based on DRL and CP, for solving combinatorial optimization problems. The core of our approach is based on a dynamic programming formulation, that acts as a bridge between both techniques. We experimentally show that our solver is efficient to solve three challenging problems: the traveling salesman problem with time windows, the 4-moments portfolio optimization problem, and the 0-1 knapsack problem. Results obtained show that the framework introduced outperforms the stand-alone RL and CP solutions, while being competitive with industrial solvers.},
  booktitle    = {Proceedings of the 35th AAAI Conference on Artificial Intelligence (AAAI)},
  publisher    = {{AAAI} Press},
  address      = {Palo Alto, California USA},
  author       = {Cappart, Quentin and Moisan, Thierry and Rousseau, Louis-Martin and Prémont-Schwarz, Isabeau and Cire, Andre A.},
  year         = {2021},
  pages        = {3677--3687}
}

@inproceedings{Felix2021,
  author    = {Chalumeau, F{\'e}lix
               and Coulon, Ilan
               and Cappart, Quentin
               and Rousseau, Louis-Martin},
  title     = {SeaPearl: A Constraint Programming Solver Guided by Reinforcement Learning},
  booktitle = {Integration of Constraint Programming, Artificial Intelligence, and Operations Research -- 18th International Conference, {CPAIOR} 2021},
  year      = {2021},
  publisher = {Springer International Publishing},
  address   = {Cham},
  pages     = {392--409},
  doi       = {10.1007/978-3-030-78230-6_25}
}

@inproceedings{Hernadvolgyi2000,
  author    = {Hern{\'a}dv{\"o}lgyi, Istv{\'a}n T.  and Holte, Robert C. and Walsh, Toby},
  title     = {Experiments with Automatically Created Memory-Based Heuristics},
  booktitle = {Abstraction, Reformulation, and Approximation. SARA 2000},
  year      = {2000},
  publisher = {Springer},
  address   = {Berlin, Heidelberg},
  pages     = {281--290},
  doi       = {10.1007/3-540-44914-0_18}
}

@inproceedings{Holte2015,
  author    = {Holte, Robert and Fan, Gaojin},
  booktitle = {Planning, Search, and Optimization: Papers from the 2015 AAAI Workshop},
  title     = {State Space Abstraction in Artificial Intelligence and Operations Research},
  pages     = {55--60},
  year      = {2015}
}

@book{Ghalib2004,
  title     = {Automated Planning},
  booktitle = {Automated Planning},
  publisher = {Morgan Kaufmann},
  address   = {Burlington},
  year      = {2004},
  series    = {The Morgan Kaufmann Series in Artificial Intelligence},
  doi       = {10.1016/B978-1-55860-856-6.X5000-5},
  author    = {Malik Ghallab and Dana Nau and Paolo Traverso}
}

@article{Aine2016,
  author  = {Sandip Aine and
             Siddharth Swaminathan and
             Venkatraman Narayanan and
             Victor Hwang and
             Maxim Likhachev},
  title   = {Multi-Heuristic {A*}},
  journal = {Int. J. Robot. Res.},
  volume  = {35},
  number  = {1--3},
  pages   = {224--243},
  year    = {2016},
  doi     = {10.1177/0278364915594029}
}

@inproceedings{Fickert2022,
  title        = {New Results in Bounded-Suboptimal Search},
  doi          = {10.1609/aaai.v36i9.21256},
  abstractnote = {In bounded-suboptimal heuristic search, one attempts to find a solution that costs no more than a prespecified factor of optimal as quickly as possible. This is an important setting, as it admits faster-than-optimal solving while retaining some control over solution cost. In this paper, we investigate several new algorithms for bounded-suboptimal search, including novel variants of EES and DPS, the two most prominent previous proposals, and methods inspired by recent work in bounded-cost search that leverages uncertainty estimates of the heuristic. We perform what is, to our knowledge, the most comprehensive empirical comparison of bounded-suboptimal search algorithms to date, including both search and planning benchmarks, and we find that one of the new algorithms, a simple alternating queue scheme, significantly outperforms previous work.},
  booktitle    = {Proceedings of the 36th AAAI Conference on Artificial Intelligence (AAAI)},
  publisher    = {{AAAI} Press},
  address      = {Palo Alto, California USA},
  author       = {Fickert, Maximilian and Gu, Tianyi and Ruml, Wheeler},
  year         = {2022},
  pages        = {10166--10173}
}

@article{Pearl1982,
  author  = {Pearl, Judea and Kim, Jin H.},
  journal = {IEEE Trans. Pattern Anal. Machh. Intell.},
  title   = {Studies in Semi-Admissible Heuristics},
  year    = {1982},
  volume  = {PAMI-4},
  number  = {4},
  pages   = {392--399},
  doi     = {10.1109/TPAMI.1982.4767270}
}

@article{Chakrabarti1989,
  title    = {Increasing Search Efficiency Using Multiple Heuristics},
  journal  = {Inf. Process. Lett.},
  volume   = {30},
  number   = {1},
  pages    = {33--36},
  year     = {1989},
  doi      = {10.1016/0020-0190(89)90171-3},
  author   = {P.P. Chakrabarti and S. Ghose and A. Pandey and S.C. {De Sarkar}}
}

@article{Baier2009,
  title    = {A Heuristic Search Approach to Planning with Temporally Extended Preferences},
  journal  = {Artif. Intell.},
  volume   = {173},
  number   = {5},
  pages    = {593--618},
  year     = {2009},
  note     = {Advances in Automated Plan Generation},
  doi      = {10.1016/j.artint.2008.11.011},
  author   = {Jorge A. Baier and Fahiem Bacchus and Sheila A. McIlraith}
}

@article{Dantzig1959,
  doi       = {10.1287/mnsc.6.1.80},
  author    = {G. B. Dantzig and J. H. Ramser},
  journal   = {Manag. Sci.},
  number    = {1},
  pages     = {80--91},
  publisher = {INFORMS},
  title     = {The Truck Dispatching Problem},
  volume    = {6},
  year      = {1959}
}

@book{Kellerer2004,
  author    = {Kellerer, Hans
               and Pferschy, Ulrich
               and Pisinger, David},
  title     = {Knapsack Problems},
  year      = {2004},
  publisher = {Springer},
  address   = {Berlin, Heidelberg},
  doi       = {10.1007/978-3-540-24777-7}
}

@article{Jackson1956,
  doi       = {10.1287/mnsc.2.3.261},
  author    = {James R. Jackson},
  journal   = {Manag. Sci.},
  number    = {3},
  pages     = {261--271},
  publisher = {INFORMS},
  title     = {A Computing Procedure for a Line Balancing Problem},
  volume    = {2},
  year      = {1956}
}

@article{Salveson1955,
  author   = {Salveson, M. E.},
  title    = {The Assembly-Line Balancing Problem},
  journal  = {J. Ind. Eng.},
  volume   = {6},
  number   = {3},
  pages    = {18--25},
  year     = {1955},
  doi      = {10.1115/1.4014559}
}

@article{Baybars1986,
  doi       = {10.1287/mnsc.32.8.909},
  author    = {İlker Baybars},
  journal   = {Manag. Sci.},
  number    = {8},
  pages     = {909--932},
  publisher = {INFORMS},
  title     = {A Survey of Exact Algorithms for the Simple Assembly Line Balancing Problem},
  volume    = {32},
  year      = {1986}
}

@article{Lawler1964,
  doi       = {10.1287/mnsc.11.2.280},
  author    = {Eugene L. Lawler},
  journal   = {Manag. Sci.},
  number    = {2},
  pages     = {280--288},
  publisher = {INFORMS},
  title     = {On Scheduling Problems with Deferral Costs},
  volume    = {11},
  year      = {1964}
}

@article{Cheng1993,
  author  = {Cheng, T. C. E. and Diamond, J. E. and Lin, B. M. T.},
  title   = {Optimal Scheduling in Film Production to Minimize Talent Hold Cost},
  journal = {J. Optim. Theory Appl.},
  volume  = {79},
  number  = {3},
  pages   = {479--492},
  year    = {1993},
  doi     = {10.1007/BF00940554}
}

@misc{gurobi,
  author  = {Gurobi Optimization, LLC},
  title   = {Gurobi Optimizer Reference Manual},
  year    = 2023,
  url     = {https://www.gurobi.com},
  urldate = {2024-05-29},
  note    = {Accessed on 2024-05-31}
}

@article{Jonhson1988,
  doi       = {10.1287/mnsc.34.2.240},
  author    = {Roger V. Johnson},
  journal   = {Manag. Sci.},
  number    = {2},
  pages     = {240--253},
  publisher = {INFORMS},
  title     = {Optimally Balancing Large Assembly Lines with {`Fable'}},
  volume    = {34},
  year      = {1988}
}

@book{Russel2020book,
  author    = {Stuart Russell and
               Peter Norvig},
  title     = {Artificial Intelligence: {A} Modern Approach},
  publisher = {Pearson},
  year      = {2020},
  edition   = {Fourth}
}

@incollection{Russel2020,
  author    = {Stuart Russell and
               Peter Norvig},
  title     = {Solving Problems by Searching},
  chapter   = {3},
  pages     = {63--109},
  booktitle = {Artificial Intelligence: {A} Modern Approach},
  publisher = {Pearson},
  year      = {2020},
  edition   = {Fourth}
}

@book{Pearl1984,
  author    = {Pearl, Judea},
  title     = {Heuristics: Intelligent Search Strategies for Computer Problem Solving},
  year      = {1984},
  publisher = {Addison-Wesley Longman Publishing Co., Inc.},
  address   = {USA},
  doi       = {10.5555/525}
}

@book{Edelkamp2012,
  author    = {Edelkamp, Stefan and Schrödl, Stefan},
  title     = {Heuristic Search: Theory and Applications},
  year      = {2012},
  publisher = {Morgan Kaufmann},
  address   = {San Francisco},
  doi       = {10.1016/C2009-0-16511-X}
}

@article{McDermott2000,
  title        = {The 1998 {AI} Planning Systems Competition},
  volume       = {21},
  doi          = {10.1609/aimag.v21i2.1506},
  abstractnote = {The 1998 Planning Competition at the AI Planning Systems Conference was the first of its kind. Its goal was to create planning domains that a wide variety of planning researchers could agree on to make comparison among planners more meaningful, measure overall progress in the field, and set up a framework for long-term creation of a repository of problems in a standard notation. A rules committee for the competition was created in 1997 and had long discussions on how the contest should go. One result of these discussions was the pddl notation for planning domains. This notation was used to set up a set of planning problems and get a modest problem repository started. As a result, five planning systems were able to compete when the contest took place in June 1998. All these systems solved problems in the strips framework, with some slight extensions. The attempt to find domains for other forms of planning foundered because of technical and organizational problems. In spite of this problem, the competition achieved its goals partially in that it con-firmed that substantial progress had occurred in some subfields of planning, and it allowed qualitative comparison among different planning algorithms. It is urged that the competition continue to take place and to evolve.},
  number       = {2},
  journal      = {AI Mag.},
  author       = {McDermott, Drew M.},
  year         = {2000},
  pages        = {35--55}
}

@book{Korte2018,
  author    = {Korte, Bernhard and Vygen, Jens},
  title     = {Combinatorial Optimization: Theory and Algorithms},
  year      = {2018},
  publisher = {Springer},
  address   = {Berlin, Heidelberg},
  edition   = {Sixth},
  doi       = {10.1007/978-3-662-56039-6}
}

@article{Fox2006,
  author     = {Fox, Maria and Long, Derek},
  title      = {Modelling Mixed Discrete-Continuous Domains for Planning},
  year       = {2006},
  issue_date = {September 2006},
  publisher  = {AI Access Foundation},
  address    = {El Segundo, CA, USA},
  volume     = {27},
  journal    = {J. Artif. Intell. Res.},
  doi        = {10.1613/jair.2044},
  pages      = {235--297},
  numpages   = {63}
}

@misc{Sanner2010,
  author = {Scott Sanner},
  title  = {Relational Dynamic Influence Diagram Language (RDDL): Language Description},
  url    = {http://users.cecs.anu.edu.au/~ssanner/IPPC_2011/RDDL.pdf},
  year   = 2010,
  note   = {Accessed on 2024-05-31}
}

@book{Pinedo2009,
  title     = {Planning and Scheduling in Manufacturing and Services},
  publisher = {Springer},
  address   = {New York, NY},
  edition   = {Second},
  author    = {Michael L. Pinedo},
  doi       = {10.1007/978-1-4419-0910-7},
  year      = {2009}
}

@article{Ibaraki1972,
  title    = {Representation Theorems for Equivalent Optimization Problems},
  journal  = {Inf. Control},
  volume   = {21},
  number   = {5},
  pages    = {397--435},
  year     = {1972},
  doi      = {10.1016/S0019-9958(72)90125-8},
  author   = {Toshihide Ibaraki}
}

@article{Ibaraki1973a,
  author   = {Ibaraki, Toshihide},
  title    = {Finite State Representations of Discrete Optimization Problems},
  journal  = {SIAM J. Comput.},
  volume   = {2},
  number   = {3},
  pages    = {193--210},
  year     = {1973},
  doi      = {10.1137/0202016}
}

@article{Ibaraki1973b,
  title    = {Solvable Classes of Discrete Dynamic Programming},
  journal  = {J. Math. Anal. Appl.},
  volume   = {43},
  number   = {3},
  pages    = {642--693},
  year     = {1973},
  doi      = {10.1016/0022-247X(73)90283-7},
  author   = {Toshihide Ibaraki}
}

@article{Ibaraki1974,
  title    = {Classes of Discrete Optimization Problems and Their Decision Problems},
  journal  = {J. Comput. Syst. Sci.},
  volume   = {8},
  number   = {1},
  pages    = {84--116},
  year     = {1974},
  doi      = {10.1016/S0022-0000(74)80024-3},
  author   = {Toshihide Ibaraki}
}

@incollection{Martelli1975b,
  author    = {Martelli, A.
               and Montanari, U.},
  editor    = {Rinaldi, Sergio},
  title     = {On the Foundations of Dynamic Programming},
  booktitle = {Topics in Combinatorial Optimization},
  year      = {1975},
  publisher = {Springer},
  address   = {Vienna},
  pages     = {145--163},
  doi       = {10.1007/978-3-7091-3291-3_9}
}

@inproceedings{Alfonso1979,
  author    = {Catalano, Alfonso
               and Gnesi, Stefania
               and Montanari, Ugo},
  title     = {Shortest Path Problems and Tree Grammars: An Algebraic Framework},
  booktitle = {Graph-Grammars and Their Application to Computer Science and Biology},
  year      = {1979},
  publisher = {Springer},
  address   = {Berlin, Heidelberg},
  doi       = {10.1007/BFb0025719},
  pages     = {167--179}
}

@article{Stefania1981,
  author    = {Gnesi, Stefania and Montanari, Ugo and Martelli, Alberto},
  title     = {Dynamic Programming as Graph Searching: An Algebraic Approach},
  year      = {1981},
  publisher = {Association for Computing Machinery},
  address   = {New York, NY, USA},
  volume    = {28},
  number    = {4},
  doi       = {10.1145/322276.322285},
  journal   = {J. ACM},
  pages     = {737--751},
  numpages  = {15}
}

@inproceedings{Eisner2005,
  title     = {Compiling Comp Ling: Weighted Dynamic Programming and the {D}yna Language},
  author    = {Eisner, Jason  and
               Goldlust, Eric  and
               Smith, Noah A.},
  booktitle = {Proceedings of Human Language Technology Conference and Conference on Empirical Methods in Natural Language Processing (HLT/EMNLP)},
  year      = {2005},
  publisher = {Association for Computational Linguistics},
  address   = {USA},
  doi       = {10.3115/1220575.1220611},
  pages     = {281--290}
}

@inproceedings{Vieira2017,
  doi       = {10.1145/3088525.3088562},
  author    = {Tim Vieira and Matthew Francis-Landau and Nathaniel
               Wesley Filardo and Farzad Khorasani and Jason Eisner},
  title     = {Dyna: Toward a Self-Optimizing Declarative Language
               for Machine Learning Applications},
  booktitle = {Proceedings of the First ACM SIGPLAN Workshop on
               Machine Learning and Programming Languages (MAPL)},
  pages     = {8--17},
  year      = {2017},
  publisher = {ACM}
}

@inproceedings{Puchinger2008,
  author    = {Puchinger, Jakob and Stuckey, Peter J.},
  title     = {Automating Branch-and-Bound for Dynamic Programs},
  year      = {2008},
  publisher = {Association for Computing Machinery},
  address   = {New York, NY, USA},
  doi       = {10.1145/1328408.1328421},
  booktitle = {Proceedings of the 2008 ACM SIGPLAN Symposium on Partial Evaluation and Semantics-Based Program Manipulation},
  pages     = {81--89},
  numpages  = {9},
  series    = {PEPM '08}
}

@inproceedings{Tamaki1986,
  author    = {Tamaki, Hisao
               and Sato, Taisuke},
  title     = {OLD Resolution with Tabulation},
  booktitle = {Third International Conference on Logic Programming (ICLP)},
  year      = {1986},
  publisher = {Springer},
  address   = {Berlin, Heidelberg},
  pages     = {84--98}
}

@inproceedings{Sundstorm2009,
  author    = {Sundstrom, Olle and Guzzella, Lino},
  booktitle = {2009 IEEE Control Applications, (CCA) \& Intelligent Control, (ISIC)},
  title     = {A Generic Dynamic Programming Matlab Function},
  year      = {2009},
  pages     = {1625--1630},
  doi       = {10.1109/CCA.2009.5281131}
}

@article{Miretti2021,
  title    = {DynaProg: Deterministic Dynamic Programming Solver for Finite Horizon Multi-Stage Decision Problems},
  journal  = {SoftwareX},
  volume   = {14},
  pages    = {100690},
  year     = {2021},
  doi      = {10.1016/j.softx.2021.100690},
  author   = {Federico Miretti and Daniela Misul and Ezio Spessa}
}

@article{Bird1980,
  author    = {Bird, R. S.},
  title     = {Tabulation Techniques for Recursive Programs},
  year      = {1980},
  publisher = {Association for Computing Machinery},
  address   = {New York, NY, USA},
  volume    = {12},
  number    = {4},
  doi       = {10.1145/356827.356831},
  journal   = {ACM Comput. Surveys},
  pages     = {403--417},
  numpages  = {15}
}

@article{Michie1968,
  author     = {Michie, Donald},
  title      = {“Memo” Functions and Machine Learning},
  journal    = {Nature},
  year       = {1968},
  volume     = {218},
  number     = {5136},
  pages      = {19--22},
  doi        = {10.1038/218019a0}
}

@phdthesis{Helman1982,
  author = {Helman, Paul},
  school = {University of Michigan, Ann Arbor},
  title  = {A New Theory of Dynamic Programming},
  year   = {1982}
}

@incollection{Kumar1988,
  author    = {Kumar, Vipin
               and Kanal, Laveen N.},
  editor    = {Kanal, Laveen
               and Kumar, Vipin},
  title     = {The CDP: A Unifying Formulation for Heuristic Search, Dynamic Programming, and Branch-and-Bound},
  booktitle = {Search in Artificial Intelligence},
  year      = {1988},
  publisher = {Springer},
  address   = {New York, NY},
  pages     = {1--27},
  doi       = {10.1007/978-1-4613-8788-6_1}
}

@article{Held1962,
  doi       = {10.1137/0110015},
  author    = {Michael Held and Richard M. Karp},
  journal   = {J. Soc. Ind. Appl. Math.},
  number    = {1},
  pages     = {196--210},
  publisher = {Society for Industrial and Applied Mathematics},
  title     = {A Dynamic Programming Approach to Sequencing Problems},
  volume    = {10},
  year      = {1962}
}

@inproceedings{Kuroiwa2024Parallel,
  author    = {Ryo Kuroiwa and J. Christopher Beck},
  booktitle = {Proceedings of the 38th AAAI Conference on Artificial Intelligence (AAAI)},
  publisher = {{AAAI} Press},
  address   = {Washington, DC, USA},
  title     = {Parallel Beam Search Algorithms for Domain-Independent Dynamic Programming},
  doi       = {10.1609/aaai.v38i18.30062},
  year      = {2024},
  pages     = {20743--20750}
}

@article{Beck1998,
  title    = {A Generic Framework for Constraint-Directed Search and Scheduling},
  volume   = {19},
  doi      = {10.1609/aimag.v19i4.1426},
  number   = {4},
  journal  = {AI Mag.},
  author   = {Beck, J. Christopher and Fox, Mark S.},
  year     = {1998},
  pages    = {103}
}

@article{Bengio2021,
  title    = {Machine Learning for Combinatorial Optimization: A Methodological Tour d’Horizon},
  journal  = {Eur. J. Oper. Res.},
  volume   = {290},
  number   = {2},
  pages    = {405--421},
  year     = {2021},
  doi      = {10.1016/j.ejor.2020.07.063},
  author   = {Yoshua Bengio and Andrea Lodi and Antoine Prouvost}
}

@article{Toth2002,
  title    = {Models, Relaxations and Exact Approaches for the Capacitated Vehicle Routing Problem},
  journal  = {Discrete Applied Mathematics},
  volume   = {123},
  number   = {1},
  pages    = {487--512},
  year     = {2002},
  doi      = {10.1016/S0166-218X(01)00351-1},
  author   = {Paolo Toth and Daniele Vigo}
}

@article{Golden1987,
  author   = {Golden, Bruce L. and Levy, Larry and Vohra, Rakesh},
  title    = {The Orienteering Problem},
  journal  = {Naval Research Logistics (NRL)},
  volume   = {34},
  number   = {3},
  pages    = {307--318},
  doi      = {10.1002/1520-6750(198706)34:3<307::AID-NAV3220340302>3.0.CO;2-D},
  year     = {1987}
}

@book{Garey1990,
  author    = {Garey, Michael R. and Johnson, David S.},
  title     = {Computers and Intractability; A Guide to the Theory of NP-Completeness},
  year      = {1979},
  publisher = {W. H. Freeman and Company},
  address   = {New York}
}

@article{Lenstra1977,
  title    = {Complexity of Machine Scheduling Problems},
  series   = {Annals of Discrete Mathematics},
  volume   = {1},
  pages    = {343--362},
  year     = {1977},
  doi      = {10.1016/S0167-5060(08)70743-X},
  author   = {J.K. Lenstra and A.H.G. {Rinnooy Kan} and P. Brucker}
}

@article{Linhares2002,
  title    = {Connections Between Cutting-Pattern Sequencing, VLSI Design, and Flexible Machines},
  journal  = {Computers \& Operations Research},
  volume   = {29},
  number   = {12},
  pages    = {1759--1772},
  year     = {2002},
  doi      = {10.1016/S0305-0548(01)00054-5},
  author   = {Alexandre Linhares and Horacio Hideki Yanasse}
}

@article{Alvalez-Miranda2019,
  title    = {On the Complexity of Assembly Line Balancing Problems},
  journal  = {Computers \& Operations Research},
  volume   = {108},
  pages    = {182--186},
  year     = {2019},
  doi      = {10.1016/j.cor.2019.04.005},
  author   = {Eduardo Álvarez-Miranda and Jordi Pereira}
}

@article{Coppe2024,
  author   = {Copp\'{e}, Vianney and Gillard, Xavier and Schaus, Pierre},
  title    = {Decision Diagram-Based Branch-and-Bound with Caching for Dominance and Suboptimality Detection},
  journal  = {INFORMS J. Comput.},
  doi      = {10.1287/ijoc.2022.0340},
  year     = {2024}
}

@inproceedings{Coppe2023,
  author    = {Copp\'{e}, Vianney and Gillard, Xavier and Schaus, Pierre},
  title     = {Boosting Decision Diagram-Based Branch-And-Bound by Pre-Solving with Aggregate Dynamic Programming},
  booktitle = {29th International Conference on Principles and Practice of Constraint Programming (CP 2023)},
  pages     = {13:1--13:17},
  series    = {Leibniz International Proceedings in Informatics (LIPIcs)},
  year      = {2023},
  publisher = {Schloss Dagstuhl -- Leibniz-Zentrum f{\"u}r Informatik},
  address   = {Dagstuhl, Germany},
  doi       = {10.4230/LIPIcs.CP.2023.13}
}

@inproceedings{Coppe2024b,
  author    = {Copp\'{e}, Vianney and Gillard, Xavier and Schaus, Pierre},
  booktitle = {Integration of Constraint Programming, Artificial Intelligence, and Operations Research -- 21st International Conference, {CPAIOR} 2024},
  title     = {Modeling and Exploiting Dominance Rules for Discrete Optimization with Decision Diagrams},
  year      = {2024}
}

@incollection{Sutton2018,
  author    = {Sutton, Richard S. and Barto, Andrew G.},
  booktitle = {Reinforcement Learning: An Introduction},
  title     = {Dynamic Programming},
  chapter   = {4},
  year      = {2018},
  edition   = {Second Edition},
  publisher = {A Bradford Book},
  address   = {Cambridge, MA, USA}
}

@incollection{Howard1960,
  author    = {Howard, Ronald A.},
  title     = {Dynamic Programming and Markov Process},
  publisher = { John Wiley \& Sons, Inc.},
  address   = {New York},
  year      = {1960}
}

@article{Lauriere1978,
  title   = {A Language and a Program for Stating and Solving Combinatorial Problems},
  journal = {Artif. Intell.},
  volume  = {10},
  number  = {1},
  pages   = {29--127},
  year    = {1978},
  doi     = {10.1016/0004-3702(78)90029-2},
  author  = {Jena-Lonis Lauriere}
}

@misc{ragnarok,
  author  = {Dominik Drexler and Daniel Gnad and Paul H\"{o}ft and Jendrik Seipp and David Speck and Simon St\r{a}hlberg},
  title   = {Ragnarok},
  year    = {2023},
  urldate = {2024-05-22},
  note    = {Accessed on 2024-05-31},
  url     = {https://ipc2023-classical.github.io/abstracts/planner17_ragnarok.pdf}
}

@inproceedings{Leofante2020,
  title     = {Optimal Planning Modulo Theories},
  author    = {Leofante, Francesco and Giunchiglia, Enrico and Ábráham, Erika and Tacchella, Armando},
  booktitle = {Proceedings of the 29th International Joint Conference on Artificial Intelligence, {IJCAI-20}},
  publisher = {International Joint Conferences on Artificial Intelligence Organization},
  pages     = {4128--4134},
  year      = {2020},
  note      = {Main track},
  doi       = {10.24963/ijcai.2020/571}
}

@inproceedings{Piacentini2018,
  title        = {Compiling Optimal Numeric Planning to Mixed Integer Linear Programming},
  doi          = {10.1609/icaps.v28i1.13919},
  abstractnote = { &lt;p&gt; Compilation techniques in planning reformulate a problem into an alternative encoding for which efficient, off-the-shelf solvers are available. In this work, we present a novel mixed-integer linear programming (MILP) compilation for cost-optimal numeric planning with instantaneous actions. While recent works on the problem are restricted to actions that modify variables present in simple numeric conditions, our MILP formulation, in addition, handles linear conditions and linear action effects on numeric state variables. Such problems are particularly challenging due to the state-dependency of the action effects. Experiments show that our approach, in addition to being the state of the art for the more general problem class, is competitive with heuristic search-based planners on domains with only simple numeric conditions. &lt;/p&gt; },
  booktitle    = {Proceedings of the 28th International Conference on Automated Planning and Scheduling (ICAPS)},
  author       = {Piacentini, Chiara and Castro, Margarita and Cire, Andre and Beck, J. Christopher},
  year         = {2018},
  publisher    = {{AAAI} Press},
  pages        = {383--387}
}

@inproceedings{Scala2017,
  author    = {Enrico Scala and Patrik Haslum and Daniele Magazzeni and Sylvie Thiébaux},
  title     = {Landmarks for Numeric Planning Problems},
  booktitle = {Proceedings of the 26th International Joint Conference on Artificial Intelligence, {IJCAI-17}},
  publisher = {International Joint Conferences on Artificial Intelligence Organization},
  pages     = {4384--4390},
  year      = {2017},
  note      = {Main track},
  doi       = {10.24963/ijcai.2017/612}
}

@inproceedings{Piacentini2018b,
  author    = {Piacentini, Chiara and Castro, Margarita P. and Cire, Andre A. and Beck, J. Christopher},
  title     = {Linear and Integer Programming-Based Heuristics for Cost-Optimal Numeric Planning},
  year      = {2018},
  publisher = {{AAAI} Press},
  address   = {Palo Alto, California USA},
  doi       = {10.1609/aaai.v32i1.12082},
  pages     = {6254--6261},
  booktitle = {Proceedings of the 32nd AAAI Conference on Artificial Intelligence (AAAI)}
}

@inproceedings{Shleyfman2023,
  title        = {Symmetry Detection and Breaking in Linear Cost-Optimal Numeric Planning},
  doi          = {10.1609/icaps.v33i1.27218},
  abstractnote = {One of the main challenges of domain-independent numeric planning is the complexity of the search problem. The exploitation of structural symmetries in a search problem can constitute an effective method of pruning search branches that may lead to exponential improvements in performance. For over a decade, symmetry breaking techniques have been successfully used within both optimal and satisficing classical planning. In this work, we show that symmetry detection methods applied in classical planning with some effort can be modified to detect symmetries in linear numeric planning. The detected symmetry group, thereafter, can be used almost directly in the A*-based symmetry breaking algorithms such as DKS and Orbit Space Search. We empirically validate that symmetry pruning can yield a substantial reduction in the search effort, even if algorithms are equipped with a strong heuristic, such as LM-cut.},
  booktitle    = {Proceedings of the 33rd International Conference on Automated Planning and Scheduling (ICAPS)},
  author       = {Shleyfman, Alexander and Kuroiwa, Ryo and Beck, J. Christopher},
  year         = {2023},
  publisher    = {{AAAI} Press},
  address      = {Palo Alto, California USA},
  pages        = {393--401}
}

@inproceedings{Kuroiwa2023Numeric,
  author    = {Ryo Kuroiwa and
               Alexander Shleyfman and
               {J. Christopher} Beck},
  title     = {Extracting and Exploiting Bounds of Numeric Variables for Optimal Linear Numeric Planning},
  booktitle = {{ECAI} 2023 -- 26th European Conference on Artificial Intelligence},
  series    = {Frontiers in Artificial Intelligence and Applications},
  volume    = {372},
  pages     = {1332--1339},
  publisher = {{IOS} Press},
  year      = {2023},
  doi       = {10.3233/FAIA230409}
}

@misc{NLMCutPlan,
  author  = {Kuroiwa, Ryo and Shleyfman, Alexander and Beck, J. Christopher},
  title   = {{NLM-CutPlan}},
  url     = {https://ipc2023-numeric.github.io/abstracts/NLM_CutPlan_Abstract.pdf},
  urldate = {2024-05-22},
  note    = {Accessed on 2024-05-31},
  year    = {2023}
}

@book{Baptiste2001,
  author    = { Philippe Baptiste and  Claude Pape and Wim Nuijten },
  title     = { Constraint-Based Scheduling: Applying Constraint Programming to Scheduling Problems},
  doi       = {10.1007/978-1-4615-1479-4},
  publisher = {Springer},
  address   = {New York, NY},
  year      = {2001},
  series    = {International Series in Operations Research \& Management Science}
}

@inproceedings{Michel2024,
  author    = {Michel, Laurent and van Hoeve, Willem-Jan},
  booktitle = {{ECAI} 2024 -- 27th European Conference on Artificial Intelligence},
  series    = {Frontiers in Artificial Intelligence and Applications},
  volume    = {392},
  pages     = {4240--4247},
  title     = {{CODD}: A Decision Diagram-Based Solver for Combinatorial Optimization},
  publisher = {{IOS} Press},
  doi       = {10.3233/FAIA240997},
  year      = {2024}
}

@book{Bertsekas1995,
  author    = {Bertsekas, Dimitri P.},
  title     = {Dynamic Programming and Optimal Control},
  year      = {1995},
  publisher = {Athena Scientific}
}

\end{document}